# YOLO advances to its genesis: a decadal and comprehensive review of the You Only Look Once (YOLO) series

**Ranjan Sapkota**[1] · **Marco Flores-Calero**[2] · **Rizwan Qureshi**[3] · **Chetan Badgujar**[4] · **Upesh Nepal**[5] · **Alwin Poulose**[6] · **Peter Zeno**[7] · **Uday Bhanu Prakash Vaddevolu**[8] · **Sheheryar Khan**[9] · **Maged Shoman**[10] · **Hong Yan**[11,12] · **Manoj Karkee**[1,13]

Accepted: 2 May 2025

**Abstract**
This review systematically examines the progression of the You Only Look Once (YOLO) object detection algorithms from YOLOv1 to the recently unveiled YOLOv12. Employing a reverse chronological analysis, this study examines the advancements introduced by YOLO algorithms, beginning with YOLOv12 and progressing through YOLO11 (or YOLOv11), YOLOv10, YOLOv9, YOLOv8, and subsequent versions to explore each version's contributions to enhancing speed, detection accuracy, and computational efficiency in real-time object detection. Additionally, this study reviews the alternative versions derived from YOLO architectural advancements of YOLO-NAS, YOLO-X, YOLO-R, DAMO-YOLO, and Gold-YOLO. Moreover, the study highlights the transformative impact of YOLO models across five critical application areas: autonomous vehicles and traffic safety, healthcare and medical imaging, industrial manufacturing, surveillance and security, and agriculture. By detailing the incremental technological advancements in subsequent YOLO versions, this review chronicles the evolution of YOLO, and discusses the challenges and limitations in each of the earlier versions. The evolution signifies a path towards integrating YOLO with multimodal, context-aware, and Artificial General Intelligence (AGI) systems for the next YOLO decade, promising significant implications for future developments in AI-driven applications.

## 1 Introduction

Object detection is a critical component of computer vision systems, which enables automated systems to identify and locate objects of interest within images or video frames (Liu et al. 2020; Badgujar et al. 2024; Ahmad and Rahimi 2022; Gheorghe et al. 2024; Arkin et al.

Extended author information available on the last page of the article

Published online: 11 June 2025
Springer

2023). Real-time object detection has become integral to numerous applications requiring real- and near-real-time analysis, monitoring and interaction with dynamic environments such as agriculture, transportation, education, and health-care (Fernandez et al. 2016; Wang et al. 2018; Ren et al. 2015; Tang et al. 2024; Chen and Guan 2022; Ragab et al. 2024). For instance, real-time object detection is the foundational technology for the success of autonomous vehicles and robotic systems (Flippo et al. 2023; Malligere Shivanna and Guo 2024; Flores-Calero et al. 2024), allowing the system to quickly recognize and track different objects of interest such as vehicles, pedestrians, bicycles, and other obstacles, enhancing navigational safety and efficiency (Guerrero-Ibáñez et al. 2018; Shoman et al. 2024a). The utility of object recognition extends beyond vehicular applications, and is also pivotal in action recognition within video sequences, useful in digital surveillance, monitoring, sports analysis, cityscapes (Hnewa and Radha 2023) and human–machine interaction (Hussain and Zeadally 2018; Fernandez et al. 2016; Shoman et al. 2024b). These areas benefit from the capability to analyze and respond to situational dynamics in real time, illustrating its broad applicability, acceptance, and impact. However, the problem of object detection involves several challenges:

- Complexity of real-world environments: Real-world environments/scenes are highly variable and unpredictable. Objects can appear in various orientations, scales, distances and lighting conditions, making it difficult for a detection algorithm to generalize and maintain accuracy in real time (Kaushal et al. 2018).
- Illumination factors: Illumination plays a crucial role in object detection, as factors like lighting intensity, direction, shadows, and glare can significantly affect performance (Xiang et al. 2014; Xiao et al. 2020). Non-uniform or low light, color temperature changes, and dynamic lighting variations can obscure object features or cause false detections. Solutions include controlled lighting setups, preprocessing techniques like normalization and color correction, and training models with diverse, augmented datasets to enhance robustness (Seoni et al. 2024).
- Occlusions and clutter: Objects may be partially or fully obscured by other objects, creating cluttered scenes that result in incomplete information, which requires careful interpretation for accurate analysis (Khan and Shah 2008; Mostafa et al. 2022).
- Speed and efficiency: Many applications necessitate rapid processing of visual data to enable timely decision-making. This requires detection algorithms to achieve a balance between high accuracy and low latency, ensuring that the systems can deliver efficient and reliable results in real- or near-real-time scenarios, such as autonomous vehicles and traffic safety, healthcare and medical imaging, industrial manufacturing, security and surveillance and agricultural automation (Gupta et al. 2021).

Addressing these challenges, as discussed below, required innovative techniques, which conventionally relied on hand-crafted features and classical machine learning methods. Later, the focus shifted towards automated feature learning and end-to-end deep learning methods.

### 1.1 Traditional object detection approaches

Before the advent of deep learning, object detection relied on a combination of hand-crafted features and machine learning classifiers (Zou et al. 2023). Some of the notable traditional methods include:

- Correlation filters: Used to detect objects by correlating a filter with the image, such as matching templates (Park et al. 2019). These approaches struggle with variations in the appearance of objects and lighting conditions (Liu et al. 2021).
- Sliding window approach: This method involves moving a fixed-size window across the image and applying a classifier to each window to determine whether it contains an object (Teutsch and Kruger 2015). However, it struggles with varying object sizes and aspect ratios, which can lead to inaccurate detections and a high computational cost due to the exhaustive search involved.
- Viola–Jones detector: The Viola-Jones detector, introduced in 2001, uses Haar-like features (Lienhart and Maydt 2002) and a cascade of AdaBoost trained classifiers (Jun-Feng and Yu-Pin 2009) to detect objects in images efficiently (Li et al. 2012).

Supporting these methods are various hand-crafted feature extraction techniques, including:

- Gabor features: Extracted texture features using Gabor filters, which are effective for texture representation but can be computationally intensive (Hu et al. 2020).
- Histogram of oriented gradients (HOG): Captures edge or gradient structures that characterize the shape of objects, typically combined with Support Vector Machines (SVM) for classification (Surasak et al. 2018).
- Local binary patterns (LBP): Utilizes pixel intensity comparisons to form a binary pattern, used in texture classification and face recognition (Karis et al. 2016).
- Haar-like features: These features consider adjacent rectangular regions in a detection window, sum up the pixel intensities in each region, and calculate the difference between the sums. This difference is then used to categorize sub-regions in images (Mita et al. 2005).
- DPM (Deformable part models): DPM (Yan et al. 2014) *represents objects as a collection of deformable parts arranged in a spatial structure. It proved particularly effective for detecting objects under occlusions, pose variations, and cluttered backgrounds.*
- SIFT: Scale-Invariant Feature Transform (SIFT) is a robust method for detecting and describing local features in images (Piccinini et al. 2012). Beyond feature extraction, it has been effectively used for object detection by matching keypoints between input images and reference templates, leveraging its invariance to scale, rotation, and illumination changes.
- SURF: speeded-up robust features A faster alternative to SIFT, SURF detects and describes features using an efficient Hessian matrix approximation, making it suitable for real-time object detection tasks (Li and Zhang 2013).

### 1.1.1 Classification techniques used in traditional object detection methods

Some of the most commonly employed classification methods for these traditional object detectors include Support Vector Machine (SVM), statistical classifiers (e.g., Bayesian Classifier) and ensemble methods (e.g. Adaboost, Random Forest) and Multilayer Perceptrons (MLP) Neural Networks (Chiu et al. 2020; Mienye and Sun 2022).

These traditional methods in early computer vision systems, reliant on hand-crafted features and classical classifiers, offered moderate success under controlled conditions but struggled with robustness and generalization in diverse real-world scenarios, lacking the accuracy achieved by modern deep learning techniques (Xiang et al. 2014). Figure 1 shows

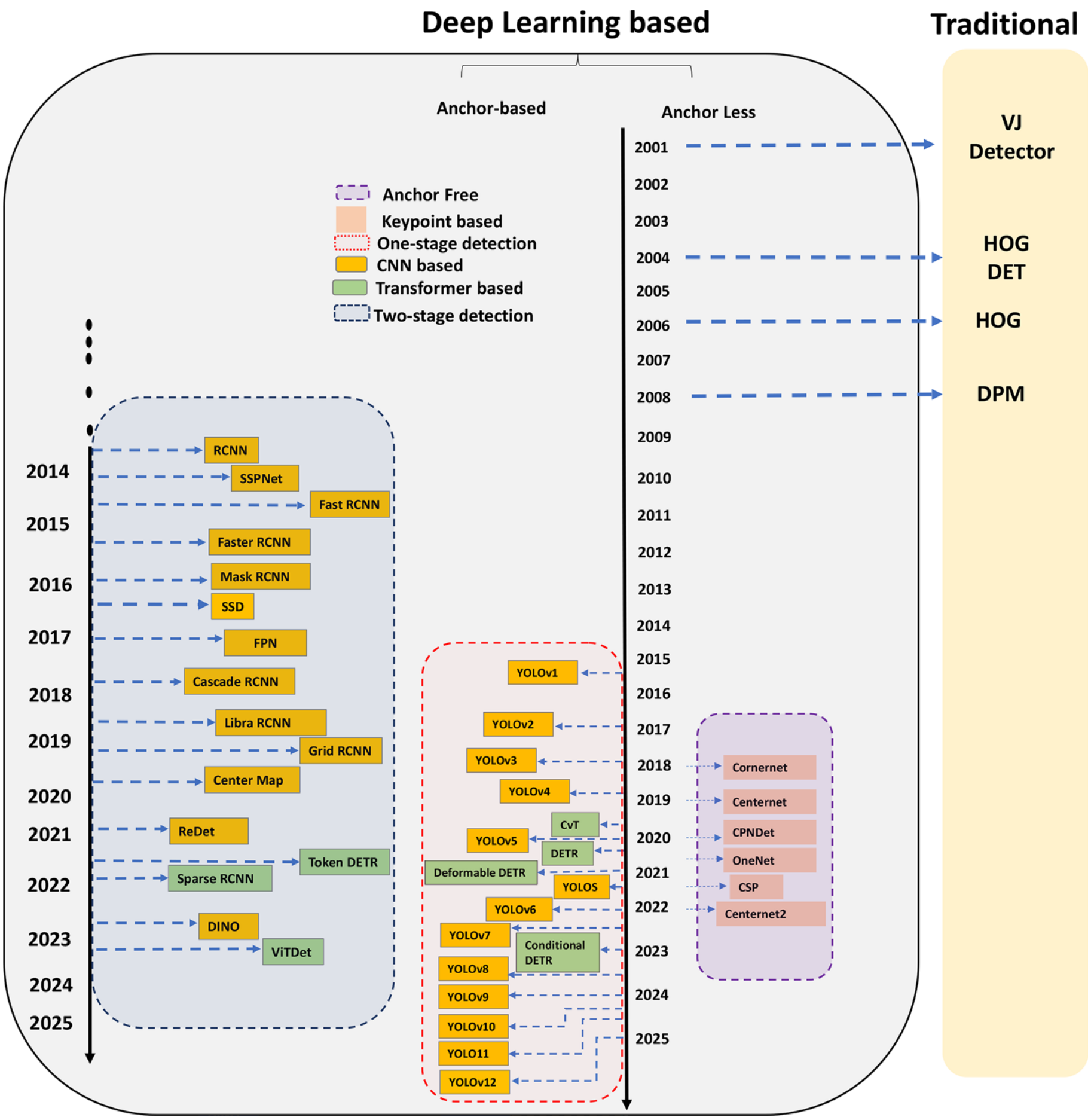

**Fig. 1** Timeline of object detection paradigms and evolution of YOLO models. The figure shows the progression from traditional methods like VJ Detector and HOG to deep learning-based approaches, including R-CNN, Fast R-CNN, Faster R-CNN, and YOLO series. Recent advancements also highlight transformer-based models, such as DETR (Detection Transformer) (Carion et al. 2020) and ViTDet (Vision Transformer for Detection) (Li et al. 2022), which have demonstrated significant progress in object detection tasks

the historical development of computer vision systems emphasizing how object detection algorithms evolved.

### 1.2 Emergence of convolutional neural networks

After 2010, the performance of handcrafted features plateaued, leading to saturation in object detection research. However, in 2012, the world witnessed the emergence of convolutional neural networks (CNNs), marking a significant turning point in the field (Krizhevsky et al. 2012; Xie et al. 2021; Tang and Yuan 2015; Zhiqiang and Jun 2017). As a deep convolutional network, CNNs are able to learn robust and high-level feature representations of an image, and are particularly effective because:

- Hierarchical feature learning: CNNs learn to extract low-level features (e.g., edges, textures) in early layers and high-level features (e.g., object parts, shapes) in deeper layers, facilitating robust object representation (Li et al. 2020).
- Spatial invariance: Convolutional layers enable CNNs to recognize objects regardless of their position within the image, enhancing detection robustness (Crawford and Pineau 2019).
- Scalability and generalizability: CNNs can be scaled to handle larger datasets and more complex models, improving performance and robustness on a wide range of tasks and application environments (Tan et al. 2020).

However, CNNs can not be directly applied to the object detection task, due to varying numbers of objects, varying sizes, aspect ratios, and orientation. CNNs were primarily designed for image classification, meaning they output a single label for the entire image (Arkin et al. 2023). Whereas object detection tasks require not only classifying the object but also localizing it in the image, i.e., identifying the position of the object through bounding boxes.

### 1.3 Timeline of object detection paradigms and the evolution of YOLO models

One of the earliest deep learning-based object detectors was R-CNN, introduced in 2014 by Girshick et al. (2014). It marked a pivotal milestone in the development of detection models, breaking the stagnation in object detection by introducing Regions with CNN features (R-CNN). This groundbreaking approach revolutionized the field, sparking rapid advancements and accelerating the evolution of object detection at an unprecedented pace.

The idea behind R-CNN is simple, it uses the selective search algorithm to generate about 2000 region proposals, which are then processed by a CNN to extract features (Girshick et al. 2014). Finally, linear SVM classifiers are utilized to detect objects within each region and identify their respective categories. While R-CNN achieved significant progress, it has notable drawbacks: the redundant feature computations across a large number of overlapping proposals (over 2,000 boxes per image) result in extremely slow detection speeds, taking 14 s per image even with GPU acceleration (Xie et al. 2021).

After that, Fast R-CNN, introduced in 2015, improved object detection by addressing the redundant feature computations across numerous overlapping proposals (Bhat et al. 2023). It integrated region proposal feature extraction and classification into a single pass, significantly enhancing efficiency and speed compared to previous methods like R-CNN (Gir-shick 2015). Building on this, Faster R-CNN advanced the approach further by introducing Region Proposal Networks (RPNs), enabling end-to-end training. This innovation eliminated the reliance on selective search, reducing computational complexity and streamlining the pipeline, thus allowing for faster and more accurate object detection without the need for external proposal generation (Ren et al. 2015; Mostafa et al. 2022).

Later, the Single Shot Multibox Detector (SSD) (Liu et al. 2016), introduced at the beginning of 2016, discretizes bounding boxes into predefined default boxes of various scales and aspect ratios. It predicts object scores and adjusts box shapes accordingly. By leveraging multi-scale feature maps, SSD handles objects of different sizes effectively. Unlike earlier methods, it eliminates the need for a separate proposal generation step, simplifying the training process and improving performance, particularly in detecting small objects.

These breakthroughs laid the foundation for numerous subsequent advancements in the field. Over the years, several other detectors have been developed, and by 2024, the YOLO series has become a dominant force in the domain, continuously evolving to improve speed, accuracy, and efficiency in real-time object detection tasks.

Fig. 1 presents a chronological overview of deep learning-based detectors, categorized into YOLO and others. Traditional detectors, shown on the right side, as well as Transformer-based detectors, are included for visual comparison.

### 1.4 You Only Look Once approach

The "You Only Look Once" (YOLO) object detection algorithm was first introduced by Redmon et al. (2016) in 2015, revolutionized real-time object detection by combining region proposal and classification into a single neural network, significantly reducing computation time. YOLO's unified architecture divides the image into a grid, predicting bounding boxes and class probabilities directly for each cell, enabling end-to-end learning (Redmon et al. 2016). YOLOv1 utilized a simplified CNN backbone, setting the stage with basic bounding box predictions. Subsequent versions like YOLOv2 incorporated the DarkNet-19 backbone and refined anchor boxes through K-means clustering. YOLOv3 expanded this further with a DarkNet-53 architecture, integrating multi-scale detection and residual connections. The series continued to innovate with YOLOv4 implementing CSPDarkNet-53 and PANet alongside mosaic data augmentation. YOLOv5 and YOLOv6 introduced CSPNet with dynamic anchor refinement and further enhancements in PANet, respectively. YOLOv7 featured an EfficientRep backbone with dynamic label assignment, while YOLOv8 introduced a Path Aggregation Network with Dynamic Kernel Attention. YOLOv9 developed multi-level auxiliary feature extraction, and YOLOv10 optimized the system with a lightweight classification head and distinct spatial and channel transformations. YOLOv11 introduced the C3k2 block in its backbone and utilized C2PSA for improved spatial attention. The latest, YOLOv12, marks a significant shift towards an attention-centric design, introducing the Area Attention ($A^2$) module for efficient large receptive field processing, Residual Efficient Layer Aggregation Networks (R-ELAN) for enhanced feature aggregation, and architectural optimizations including FlashAttention and adjusted MLP ratios. This attention-based approach allows YOLOv12 to achieve state-of-the-art performance in both accuracy and efficiency, surpassing previous CNN-based models while maintaining real-time detection capabilities (Tian et al. 2025). autonomous vehicles and traffic safety (Gheorghe et al. 2024; Wang et al. 2024; Shoman et al. 2024a, c), healthcare (Patel et al. 2023; Luo et al. 2021;

Salinas-Medina and Neme 2023), industrial applications (Pham et al. 2023; Klarák et al. 2024; Wang et al. 2024), surveillance and security (Arroyo et al. 2019; Bordoloi et al. 2020) and agriculture (Badgujar et al. 2024; Li et al. 2022; Fu et al. 2021; Zhong et al. 2018; Wang et al. 2022; Jiang et al. 2022; Chen et al. 2023; Yu et al. 2024; Jia et al. 2023; Umar et al. 2024; Sapkota et al. 2024a), where accuracy and speed are crucial.

In autonomous vehicles and traffic safety, YOLO has been widely adopted to improve safety in both road transport and aviation, allowing real-time object detection for collision avoidance, traffic monitoring, pedestrian detection, traffic sign detection, and aviation hazard analysis. This application aids in minimizing accidents and improving operational efficiency (Gheorghe et al. 2024; Bakirci and Bayraktar 2024; Wang et al. 2024).

In healthcare, YOLO has been instrumental in assisting and improving diagnostic processes and treatment outcomes. The applications include, but are not limited to, cancer detection (Prinzi et al. 2024; Aly et al. 2021), skin segmentation (Ünver and Ayan 2019), and pill identification (Tan et al. 2021; Suksawatchon et al. 2022) which showcase the model's ability to adapt to different needs, and essential tasks.

In industrial applications, YOLO aids in surface inspection processes to detect defects and anomalies (Pham et al. 2023; Klarák et al. 2024), structural health monitoring (Pratibha et al. 2023), reliability and safety (Fahim and Hasan 2024) ensuring quality control in manufacturing and production (Pham et al. 2023).

Surveillance and security systems also leverage YOLO for real-time monitoring and rapid identification of suspicious activities (Arroyo et al. 2019; Bordoloi et al. 2020). By integrating these models into surveillance systems, security personnel can more effectively monitor and respond to potential threats, enhancing public safety (Gorave et al. 2020). Similarly, in the context of public health measures like social distancing and face mask detection during pandemics (Kolpe et al. 2022; Bashir et al. 2023), YOLO models provided essential support in enforcing health regulations.

In agriculture, YOLO models have been applied to detect and classify crops (Ajayi et al. 2023), pests, and diseases (Morbekar et al. 2020; Li et al. 2022; Cheeti et al. 2021), facilitating precision agriculture techniques and automating farming operations to increase productivity and optimizing inputs. Additionally, in remote sensing, YOLO contributes to object recognition in satellite (Pham et al. 2020; Cheng et al. 2021) and aerial imagery (Chen et al. 2023; Luo et al. 2022), which supports urban planning, land use mapping, and environmental monitoring. These capabilities demonstrate YOLO's contribution to critical global challenges such as urban development and environmental conservation.

### 1.5 Motivation and organization of the study

Since "You Only Look Once" has been widely adopted in the field of computer vision, a search for this keyword in Google Scholar yields approximately 5,550,000 results as of June 9, 2024. The acronym "YOLO" further emphasizes its popularity, generating around 210,000 search results at the same time instant. Thousands of researchers have cited YOLO papers, highlighting its significant influence. This study aims to review and critically summarize the YOLO's decadal progress and its advancements over time, as visually summarized in the mind-map, shown in Fig. 2.

This comprehensive analysis starts with Sect. 2: *YOLO trajectory*, tracing the evolution from YOLOv1 to YOLOv12. In Sect. 3: *Context and distinctions of prior YOLO literature*

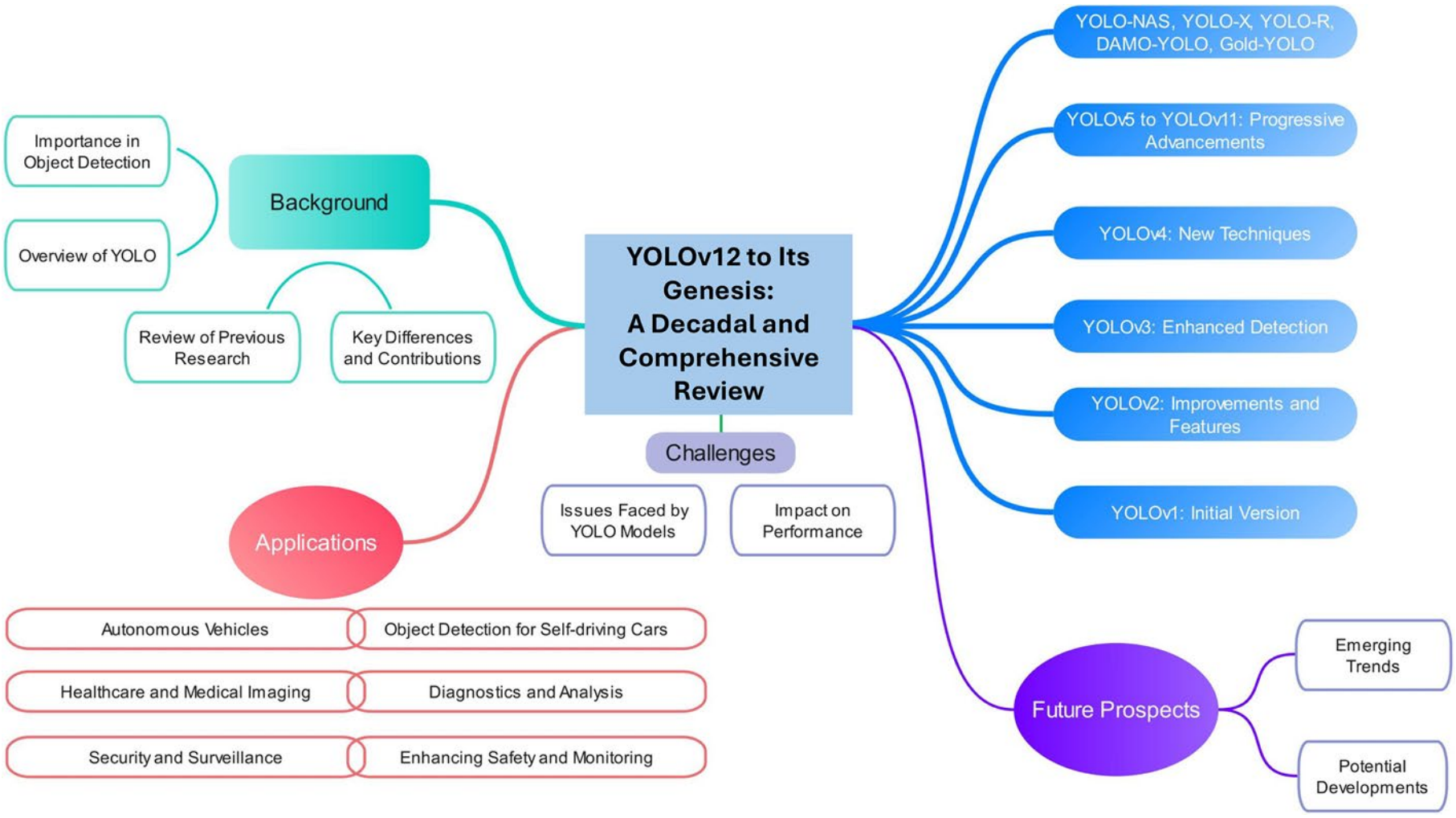

**Fig. 2** A schematic of the manuscript. We discuss all versions of YOLO, with comparative analysis. Applications in key areas; such as, autonomous vehicles, and security and surveillance are also presented. Challenges in each YOLO version, with performance enhancements, are also highlighted, we also provide visionary thoughts on the future impact of YOLO on industry and society

offers insights into the background and unique aspects of earlier studies. Section 4: *Review of YOLO versions* details the key features and improvements of each version. In Sect. 5: *Applications* various use cases across different domains are highlighted, demonstrating the versatility of YOLO models. Following this, Sect. 6 *Challenges, limitations and future directions* addresses current issues and potential advancements. Finally, the *Conclusion* section summarizes the findings of this comprehensive review. Each section is further divided into various sub-subsections to present and discuss specific topic areas relevant to the corresponding sections.

## 2 The evolution of YOLO: trajectory and variants

Fig. 1 illustrates the development timeline of the YOLO models, beginning with the release of YOLOv1 and progressing to the latest version, YOLO12. This timeline highlights the key advancements and iterations in the YOLO series.

YOLOv1 (Redmon et al. 2016) was introduced in 2016 as a novel approach to object detection, offering good accuracy and computational speed by processing images using a single stage network architecture. The first YOLO version laid the foundation for real-time applications of machine vision systems, setting a new standard for subsequent developments.

YOLOv2, or YOLO9000 (Li and Yang 2018; Nakahara et al. 2018), expanded on the foundation of YOLOv1 by improving the resolution at which the model operated and by expanding the capability to detect over 9000 object categories, thus enhancing its versatility and accuracy. YOLOv2 introduced two primary variants: a smaller version optimized for speed and a larger version focused on higher accuracy.

YOLOv3 further advanced these capabilities by implementing multi-scale predictions and a deeper network architecture, which allowed better detection of smaller objects (Kim et al. 2018). YOLOv3 introduced three primary variants, each designed to balance model size and performance: YOLOv3-spp (Small), Standard, and YOLOv3-tiny(Tiny), catering to different trade-offs between speed and accuracy.

The series continued to evolve with YOLOv4 and YOLOv5, each introducing more refined techniques and optimizations to improve detection performance (i.e., accuracy and speed) even further (Nepal and Eslamiat 2022; Sozzi et al. 2022; Mohod et al. 2022). YOLOv4 introduced four main variants: the standard version, YOLOv4-CSP, which incorporates Cross-Stage Partial (CSP) networks to enhance performance and reduce computational cost; YOLOv4x-mish, which utilizes the Mish activation function to improve accuracy while maintaining efficiency; and YOLOv4-tiny, a lightweight version optimized for real-time applications and edge devices, sacrificing some accuracy for speed. YOLOv5, developed by Ultralytics (2020) [1], brought significant improvements in terms of ease of use and performance, establishing itself as a popular choice in the computer vision community. YOLOv5 introduced five primary variants to meet various performance needs: YOLOv5s (small), optimized for speed and efficiency in resource-constrained environments; YOLOv5m (medium), offering a balanced trade-off between speed and accuracy; YOLOv5l (large), designed for higher accuracy at the expense of resources; YOLOv5x (extra-large), focused on top-tier accuracy for powerful hardware; and YOLOv5n (nano), a lightweight version tailored for rapid inference and low computational demands, ideal for real-time applications and edge devices.

Subsequent versions, YOLOv6 through YOLO11, have continued to build on this success, focusing on enhancing model scalability, reducing computational demands, and improving real-time performance metrics.

Li et al. (2022) introduced YOLOv6 in 2022. Developed by a team from Meituan, a Chinese e-commerce platform, YOLOv6 features a novel backbone and neck architecture. It also incorporates advanced training techniques such as Anchor-Aided Training (AAT) and Self-Distillation to enhance performance and efficiency. YOLOv6 introduces three main variants: the standard version, balancing accuracy and speed for general detection tasks; YOLOv6-Nano, optimized for real-time applications with a focus on speed and performance on edge devices; and YOLOv6-Tiny, designed for even faster inference on low-resource hardware, trading off some accuracy.

YOLOv7 (Wang et al. 2022, 2023) introduces advanced techniques like trainable bag-of-freebies (optimizations that improve accuracy without increasing inference cost) and dynamic label assignment. It introduces three variants: the standard version, balancing speed and accuracy; YOLOv7-X, a more powerful variant optimized for performance but requiring more computational resources; and YOLOv7-Tiny, a lightweight version designed for real-time applications on edge devices, prioritizing speed over accuracy.

YOLOv8 was released in 2023 by Ultralytics (Jocher et al. 2023). It features a more efficient architecture, enhanced training techniques, and support for larger datasets. Its user-friendly implementation in PyTorch makes it accessible for both research and production. YOLOv8 introduces four variants: YOLOv8-S, optimized for fast inference on edge devices with some accuracy trade-offs; YOLOv8-M, balancing accuracy and speed for general tasks;

[1] Ultralytics specializes in AI and deep learning, known for YOLOv5, YOLOv8, and YOLO11 models used in object detection, segmentation, and classification for computer vision tasks.

YOLOv8-L, prioritizing accuracy at the cost of computational demand; and YOLOv8-Tiny, a lightweight version for real-time applications.

YOLOv9 (Wang et al. 2024) proposed the concept of programmable gradient information (PGI) to cope with the various changes required by deep networks to achieve multiple objectives. PGI can provide complete input information for the target task to calculate objective function, so that reliable gradient information can be obtained to update network weights. In addition, a new lightweight network architecture – Generalized Efficient Layer Aggregation Network (GELAN), based on gradient path planning is designed. GELAN's architecture confirms that PGI has gained superior results on lightweight models. We verified the proposed GELAN and PGI on MS COCO dataset based object detection. Its variants are YOLOv9t, YOLOv9s, YOLOv9m, YOLOv9c, YOLOv9e (Ultralytics 2023a).

YOLOv10 (Ultralytics 2023b), developed by researchers at Tsinghua University, introduces a novel approach to real-time object detection, addressing the limitations of both post-processing and model architecture in previous YOLO versions. By eliminating non-maximum suppression (NMS) and optimizing key components of the model, YOLOv10 offers significant improvements in efficiency and performance. This version introduces six distinct variants as YOLOv10-N, YOLOv10-S, YOLOv10-M, YOLOv10-B, YOLOv10-L, and YOLOv10-X (Wang et al. 2024).

Notably, YOLOv10-N and YOLOv10-S exhibit the lowest latencies at 1.84 ms and 2.49 ms, respectively, making them highly suitable for applications requiring low latency. These models outperform their predecessors, with YOLOv10-X achieving the highest mAP of 54.4% and a latency of 10.70 ms, reflecting a well-balanced enhancement in both accuracy and inference speed.

YOLOv11 represents a significant advancement in object detection, featuring a sophisticated backbone and neck architecture for enhanced feature extraction. It optimizes speed and efficiency while maintaining high accuracy. YOLOv11 balances precision and computational efficiency, suitable for various applications from embedded systems to large-scale deployments. The model comes in five variants: YOLOv11n, YOLOv11s, YOLOv11m, YOLOv11L, and YOLOv11x, based on network depth.

YOLOv12 (Tian et al. 2025) further revolutionizes the field with an attention-centric approach. It introduces the Area Attention ($A^2$) module and Residual Efficient Layer Aggregation Networks (R-ELAN) for improved feature processing. Utilizing these architectural changes, YOLOv12 achieves state-of-the-art performance while maintaining real-time detection capabilities. For instance, YOLOv12-N was reported to achieve 40.6% mAP with a 1.64 ms inference latency on a T4 GPU, outperforming YOLOv10-N and YOLOv11-N by 2.1 mAP with a comparable speed. YOLOv12 also demonstrated superior object contour definition and foreground activation compared to its predecessors. It supports various tasks, including object detection, segmentation, classification, pose estimation, and oriented object detection, making it a versatile tool for diverse computer vision applications. YOLOv12 is available in five variants: YOLOv12-N, YOLOv12-S, YOLOv12-M, YOLOv12-L, and YOLOv12-X, each offering different trade-offs between performance and computational requirements (Tian et al. 2025).

To summarize, each iteration of the YOLO series has set new benchmarks for object detection capabilities and significantly impacted various application areas, from autonomous vehicles and traffic safety

## 2.1 Significance of latency and mAP scores in YOLO

Inference Time ($T_{inf}$) and mean Average Precision (mAP) are critical metrics to assess the performance of object detection models such as YOLO (Tang and Yuan 2015; Mao et al. 2019). Inference Time specifically measures the duration required for the model to process an image and generate predictions, focusing solely on the computational phase and is typically measured in milliseconds (ms) (Mao et al. 2019). This metric excludes any delays from image pre-processing or post-processing, providing a clear measure of the model's computational efficiency. Lower inference times are crucial for real-time applications such as autonomous driving, surveillance, and robotics, where rapid and accurate detections are essential (Chen et al. 2020). High inference times can lead to delays that compromise safety and effectiveness in these dynamic settings (Pestana et al. 2021).

Frames Per Second (FPS) is another essential metric that indicates how many images the model can evaluate each second, complementing inference time by illustrating the model's ability to handle streaming video or rapid image sequences. Both inference time and FPS provide a detailed view of the real-time operational performance of a model.

It is also important to note that these performance metrics are highly dependent on the hardware platform used for testing. Differences in computational power can significantly influence results, which makes it essential to standardize hardware during benchmark tests to ensure fair comparisons. Likewise, mAP is a comprehensive metric used to evaluate the accuracy of object detection models (Zhou et al. 2018). It considers both precision and recall (Table 1), and it is calculated by taking the average precision (AP) across all classes and then averaging these AP scores (Hall et al. 2020; Zhou et al. 2018). It provides a balanced view of how well the model performs across different object categories and varying conditions within the dataset. Other metrics used for comprehensive evaluation of YOLO models (Goutte and Gaussier 2005; Liang et al. 2021) are detailed in Table 1.

**Table 1** Summary of performance metrics used in model evaluation

| No. | Performance metric | Symbol | Equation | Description |
|---|---|---|---|---|
| 1 | Precision | $P$ | $P = \frac{TP}{TP+FP}$ | Ratio of true positive detections to the total predicted positives |
| 2 | Recall | $R$ | $R = \frac{TP}{TP+FN}$ | Ratio of true positive detections to the total actual positives |
| 3 | F1 score | $F1$ | $F1 = 2 \cdot \frac{P \cdot R}{P+R}$ | Harmonic mean of precision and recall, balancing both metrics to provide a single performance measure for the model |
| 4 | Intersection over union | IoU | $\mathrm{IoU} = \frac{\text{Area of overlap}}{\text{Area of union}}$ | Measures the overlap between the predicted and actual bounding boxes |
| 5 | Frames per second | FPS | $\mathrm{FPS} = \frac{1}{L}$ | Number of images the model processes per second, inversely related to latency |
| 6 | Non-maximum suppression | NMS | – | NMS is a post-processing step in YOLO to remove redundant bounding-boxes |

Here, True Positives (TP), True Negatives (TN), False Positives (FP), and False Negatives (FN) are the key performance evaluators. TP is instance where the model correctly identifies an object as present. TN occurs when the model correctly predicts the absence of an object. FP arises when the model incorrectly identifies an object as present, and FN happens when the model fails to detect an object that is actually present. These metrics are crucial for assessing the accuracy and reliability of the YOLO object detection (Hall et al. 2020; Zhou et al. 2018; Liang et al. 2021).

## 2.2 Single-stage detection with YOLO

The Single Shot MultiBox Detector (SSD) (Liu et al. 2016) (Fig. 3) introduced in 2016 revolutionized object detection by streamlining the process through a single-stage approach, significantly inspiring subsequent developments in YOLO models (Liu et al. 2016; Fu et al. 2017; Zhang et al. 2018). Unlike two-stage models like R-CNN, which rely on a region proposal step before actual object detection, SSD and by extension, YOLO variants, perform detection and classification in a single sweep across the image. This paradigm shift enhances the detection process by eliminating intermediate steps, thus facilitating faster and more efficient object detection suitable for real-time applications. The architecture of SSD, which YOLO models have adapted, utilizes multiple feature maps at different resolutions to detect objects of various sizes, employing a diverse array of anchor boxes at each feature map location to improve localization accuracy (Cui et al. 2018; Lin et al. 2017).

Figure 3 illustrates a YOLO model that incorporates SSD's architectural principles to enhance real-time detection capabilities through improved feature extraction using Multi-Headed Attention layers. This adoption from SSD methodology significantly boosts the processing speed and detection accuracy of models such as YOLOv8, YOLOv9, and YOLOv10,

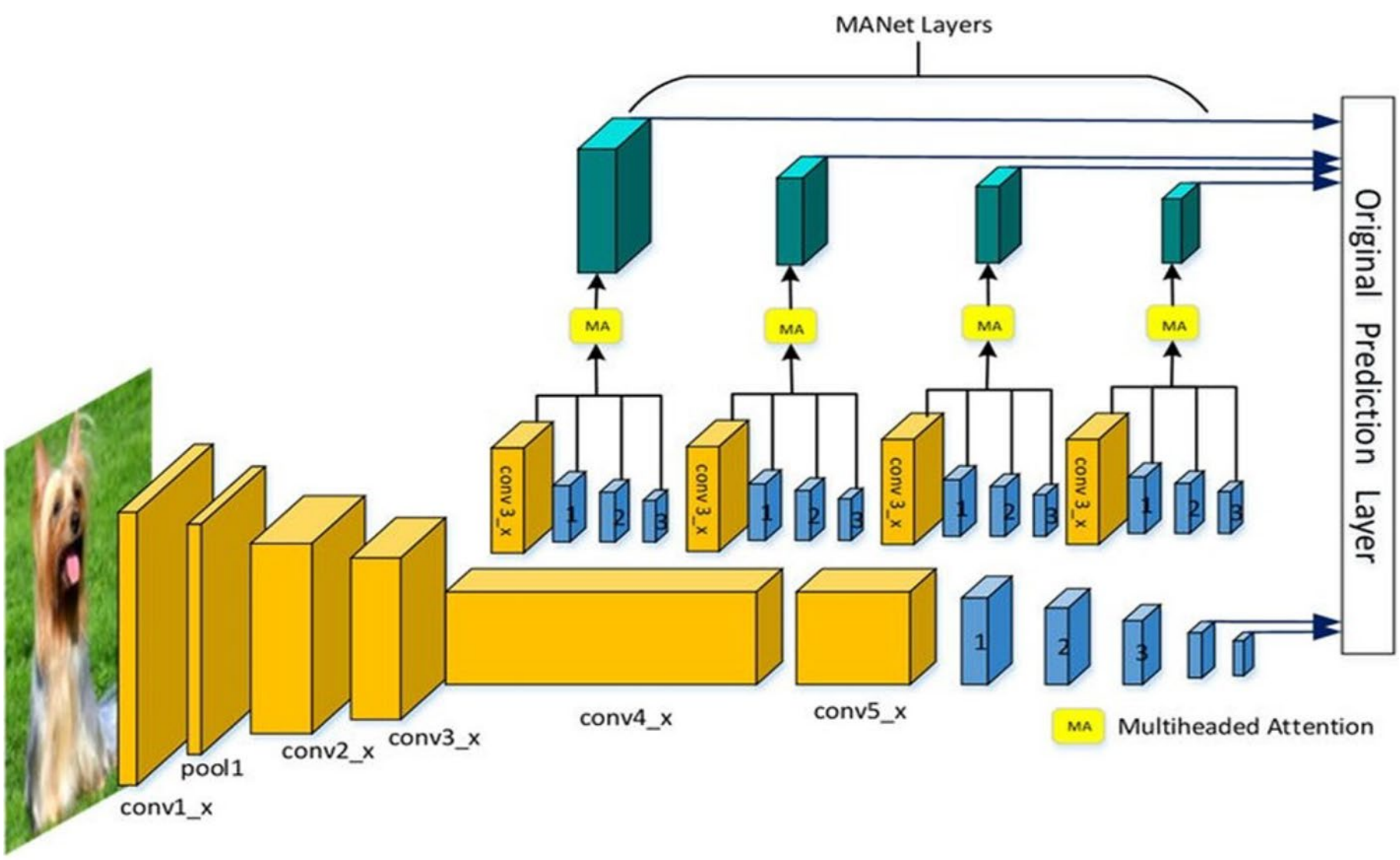

**Fig. 3** Enhanced YOLO model architecture incorporating SSD's single-stage detection approach with Multi-Headed Attention (MA) layers for superior real-time object detection performance (Jiang et al. 2019)

making them ideal for rapid and reliable object detection in resource-constrained environments (Tang et al. 2018; Li et al. 2017). The efficient single-shot mechanism, which directly classifies and localizes objects, highlights the ongoing evolution of the YOLO series to meet the accuracy and speed requirements of diverse real-world scenarios (Zhang et al. 2018).

## 3 Prior YOLO literature: context and distinctions

We collected the existing published literature on YOLO to document and critically analyze the past knowledge, including major highlights and limitations, which are briefly summarized and discussed here:

- *A review of YOLO algorithm developments* by Jiang et al. (2022) provided an insightful overview of YOLO algorithm development and its evolution through its versions. The authors analyze the fundamental aspects of YOLO's to object detection, comparing its various iterations to traditional CNNs. They emphasize ongoing improvements in YOLO, particularly in enhancing target recognition and feature extraction capabilities. It also discusses the application of YOLO in specific fields, such as finance, highlighting its practical implications in feature extraction for image-based news analysis (Jiang et al. 2022).
- *A comprehensive systematic review of YOLO for medical object detection (2018 to 2023)* by Ragab et al. (2024) presented a systematic review of YOLO's application in the medical field, that analyzes how different variants, particularly YOLOv7 and YOLOv8, have been employed for various medical detection tasks. They highlight the algorithm's significant performance in lesion detection, skin lesion classification, and other critical areas, demonstrating YOLO's superiority over traditional methods in accuracy and computational efficiency. Despite its successes, the review identifies challenges, such as the need for well-annotated datasets and addresses the high computational demands of YOLO implementations. The paper suggested directions for future research to optimize YOLO's application in medical object detection (Ragab et al. 2024).
- *A comprehensive review of YOLO architectures in computer vision: from YOLOv1 to YOLOv8 and YOLO-NAS* by Terven et al. (2023) provides an extensive analysis of the evolutionary trajectory of the YOLO algorithm, detailing how each iteration has contributed to advances in real-time object detection. Their review covers the significant architectural and training enhancements from YOLOv1 through YOLOv8 and introduces YOLO-NAS and YOLO with Transformers. This study serves as a valuable resource for understanding the progression in network architecture, which has progressively improved YOLO's efficacy in diverse applications such as robotics and autonomous driving.
- *YOLOv1 to v8: unveiling each variant-a comprehensive review of YOLO* by Hussain (2024), provided in-depth analyses of the internal components and architectural innovations of each YOLO variant. It provided a deep dive into the structural details and incremental improvements that have marked the evolution of YOLO, presenting a well-structured analysis complete with performance benchmarks. This methodological approach not only highlights the capabilities of each variant but also discusses their practical impact across different domains, suggesting the potential for future enhancements like federated learning to improve privacy and model generalization (Hussain 2024).

- *YOLO-v1 to YOLO-v8, the rise of YOLO and its complementary nature toward digital manufacturing and industrial defect detection* by Hussain (2023) reviewed and showed rapid progression of the YOLO variants, focusing on their critical role in industrial applications, specifically for defect detection in manufacturing. Starting with YOLOv1 and extending through YOLOv8, the paper illustrates how each version has been optimized to meet the demanding needs of real-time, high-accuracy defect detection on constrained devices. Hussain's work not only examines the technical advancements within each YOLO iteration but also validates their practical efficacy through deployment scenarios in the manufacturing sector, emphasizing YOLO's alignment with industrial needs (Hussain 2023).
- *YOLOv1 to YOLOv10: The fastest and most accurate real-time object detection systems* by Wang et al. (2024) provide a comprehensive literature survey on the YOLO series, from YOLOv1 to YOLOv10. This review uniquely revisits the characteristics of YOLO through a contemporary technical lens, highlighting its ongoing influence in advancing real-time computer vision research and subsequent technological developments. The authors explore the evolution of YOLO's methodologies over the past decade and its diverse applications in fields requiring real-time object analysis. By doing so, they underscore YOLO's role as a foundational technology in various computer vision tasks, such as instance segmentation and 3D object detection.
- *Evaluating the evolution of YOLO models: a comprehensive benchmark study of YOLO11 and its predecessors* by Jegham et al. (2024) conducts a detailed benchmark analysis of YOLO models from YOLOv3 to YOLO11. It examines their performance across three diverse datasets: Traffic Signs, African Wildlife, and Ships and Vessels, focusing on different challenges like object size and aspect ratio. Employing metrics like Precision, Recall, mAP, Processing Time, GFLOPs, and Model Size, the study identifies the strengths and limitations of each version, with YOLO11m showing exceptional balance in accuracy and efficiency across the datasets.

For the comprehensive review articles, it is advantageous to pinpoint a specific gap that the proposed review will address. For instance, a common oversight in the existing literature is the omission of the latest YOLO iterations, particularly YOLOv9, YOLOv10, YOLOv11 and YOLOv12 or neglecting to cover the application domains of interest. Given the YOLO algorithm's ten-year milestone, there is a pressing need to systematically document and critically evaluate these newer models. Our review aims to fill this void by providing updated, in-depth insights and comparative analysis of YOLOv9 and YOLOv12, extending across various applications to serve the wider research and technical community. This state-of-the-art review intends to highlight the continued advancements and capabilities of these models within the dynamic field of object detection technology.

In this review paper, we adopt a unique reverse-chronological approach to analyze the progression of YOLO, beginning with the most recent versions and moving backwards. The analysis is divided into six distinct subsections. The first subsection covers the latest iterations, YOLOv12 and YOLOv11, The second subsection examines YOLOv10, YOLOv9, and YOLOv8, where we delve into the architecture and advancements that define the forefront of object detection technology. This approach not only shows the most cutting-edge developments but also sets the stage for understanding the incremental improvements that have been realized over time. The third subsection reviews YOLOv7, YOLOv6, and YOLOv5, tracing further back in the series to highlight the evolutionary steps contributing to the enhancements observed in the later versions. We analyze each model's technical and scientific aspects to provide a comprehensive view of the progress within these iterations. The fourth subsection addresses the earlier YOLO versions, offering a complete historical perspective that enriches the reader's understanding of the foundational technologies and the methodologies, refined through successive updates. The fifth subsection presents alternative versions derived from YOLO.

To close this section, we discuss the application of the YOLO models in reverse order across five critical real-world domains: autonomous vehicles and traffic safety, healthcare and medical image analysis, surveillance and security, industrial manufacturing and agriculture. For each application, we present a detailed examination and corresponding tabular data in reverse chronological order, showcasing how YOLO technologies have been adapted and implemented to meet specific industry needs and challenges. This reverse review strategy not only emphasizes the state-of-the-art but also provides a narrative of technological evolution, illustrating how each iteration builds upon the last to push the boundaries of what's possible in object detection. By understanding where YOLO technology stands today and how it got there, readers gain a comprehensive view of its capabilities and potential future directions. This methodical unpacking of the YOLO series not only highlights technological advancements but also offers insights into the broader implications and utility of these models in practical scenarios, setting the groundwork for anticipating future innovations in object detection technology.

## 4 Review of YOLO versions

This section reviews YOLO series models, starting from the advanced and latest version, YOLOv12, and progressively tracing back to the foundational YOLOv1. By first highlighting the most recent technological advancements, this approach enables immediate insights into the state-of-the-art capabilities of object detection. Subsequently, the narrative explores how earlier models laid the groundwork for these innovations.

### 4.1 YOLOv12 and YOLO11

YOLOv12 (Tian et al. 2025) is the most recent YOLO version introduced in February 2025, which marks a substantial advancement in real-time object detection by integrating attention mechanisms into the YOLO framework while maintaining competitive inference speeds. This attention-centric framework not only surpasses popular real-time detectors (e.g., Faster R-CNN, RetinaNet, Detectron 2) in accuracy but also achieves state-of-the-art performance through a combination of innovative attention methods, Residual Efficient Layer Aggregation Networks (R-ELAN), and several architectural optimizations.

Fig. 4a demonstrates latency comparisons on the MS COCO benchmark dataset, highlighting YOLOv12's significantly lower inference latency compared to YOLOv11, YOLOv10, YOLOv9, and YOLOv8. The curve reveals a substantial reduction in latency, enabling faster processing speeds while maintaining high detection accuracy (Tian et al. 2025). Complementing this, Fig. 4b presents comparison in terms of GFLOPs, which shows

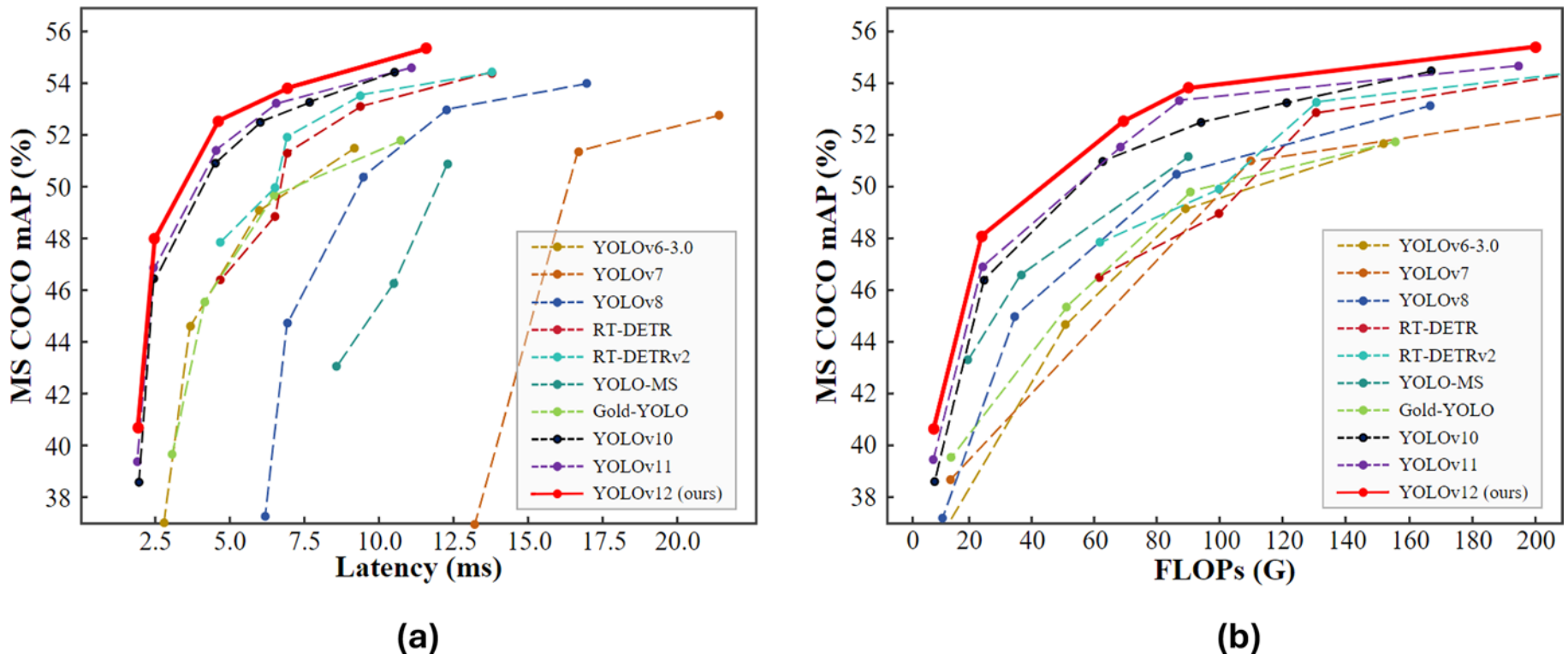

**Fig. 4** **a** Latency comparison on the MS COCO benchmark reveals significantly faster inference achieved with YOLOv12 compared to the same achieved with previous YOLO versions. **b** GFLOPs analysis also shows enhanced computational efficiency. (Tian et al. 2025)

YOLOv12 achieves higher computational efficiency, reflecting its ability to handle complex computations effectively. This balance between speed and computational power demonstrates YOLOv12's robust performance.

On the COCO dataset, YOLOv12 sets a new state-of-the-art standard, with the lightweight YOLOv12-N achieving 40.6% mAP and the larger YOLOv12-X reaching 55.2% mAP. These results, combined with the latency and GFLOPs improvements, establish YOLOv12 as a new benchmark in real-time object detection. The model framework is available in five scales: YOLOv12-N, S, M, L, and X, each optimized for specific applications. For instance, YOLOv12-N achieves 40.6% mAP at 1.64 ms on a T4 GPU, outperforming YOLOv10-N and YOLOv11-N by 2.1% and 1.2% mAP, respectively. Similarly, YOLOv12-S attains 48.0% mAP at 2.61 ms/image, surpassing YOLOv8-S, YOLOv9-S, YOLOv10-S, and YOLOv11-S by margins of 3.0%, 1.2%, 1.7%, and 1.1% mAP. The larger models in the YOLOv12 family continue to show improvements in efficiency and performance. Notably, YOLOv12-M achieves 52.5% mAP at 4.86 ms/image. In terms of computational effieciency, YOLOv12-L demonstrates a significant reduction in FLOPs, decreasing by 31.4G compared to its predecessor, YOLOv10-L. At the highest end of the scale, YOLOv12-X showcases superior performance, outperforming both YOLOv10-X and YOLOv11-X in detection accuracy (Tian et al. 2025).

YOLOv12 also surpasses end-to-end detectors like RT-DETR and RT-DETRv2. For example, YOLOv12-S runs 42% faster than RT-DETR-R18 and RT-DETRv2-R18, using only 36% of the computation and 45% of the parameters. Residual connections show minimal impact on convergence in smaller models (YOLOv12-N) but are critical for stable training in larger models (YOLOv12-L/X), with YOLOv12-X requiring a scaling factor of 0.01. The area attention module reduces inference time by 0.7 ms on an RTX 3080 with FP32 precision, while FlashAttention further accelerates inference by 0.3-—0.4 ms.

Visualization analyses confirm that YOLOv12 produces clearer object contours and more precise foreground activations than its predecessors. A convolution-based attention implementation proves to be faster than linear alternatives. Additionally, a hierarchical design, extended training (approximately 600 epochs), an optimized convolution kernel size ($7 \times 7$), absence of positional embedding, and an MLP ratio of 1.2 collectively enhance the framework's performance and efficiency.

As discussed before, the YOLOv12 architecture (Fig. 5a), demonstrates an advanced integration of $A^2$ (Area Attention) modules, R-ELAN (Residual Efficient Layer Aggregation Networks) blocks, and a streamlined detection head. This design optimizes the model's visual information processing while maintaining high accuracy. The major innovations on the YOLOv12 architecture are listed below.

#### 4.1.1 YOLOv12 architectural innovation

- Area attention ($A^2$) module: This module implements segmented feature processing with Flash Attention integration, reducing computational complexity by 50% through spatial reshaping while maintaining large receptive fields. AA enables real-time detection at fixed $n = 640$ resolution through optimized memory access patterns, as illustrated in Fig. 5a.
- Residual ELAN (R-ELAN) hierarchy: R-ELAN combines residual shortcuts (0.01 scaling) with dual-branch processing to mitigate the gradient vanishing problem. The model also features a streamlined final aggregation stage that reduces parameters by 18% and FLOPs by 24% compared to baseline architectures, as shown in Fig. 5b.

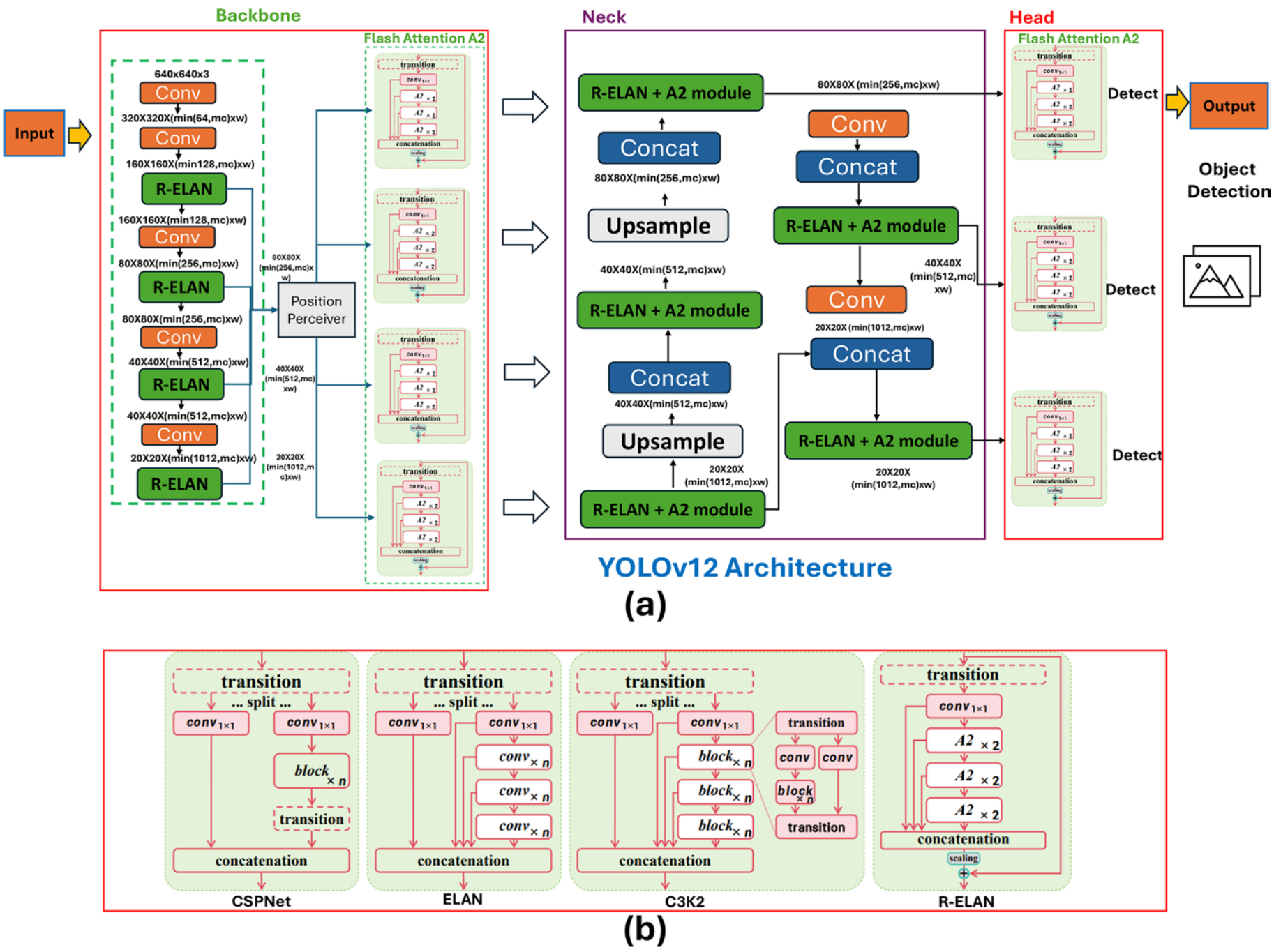

**Fig. 5** **a** Architecture of the YOLOv12 object detection model that integrated Area Attention ($A^2$) modules, R-ELAN blocks, and a streamlined detection head; **b** Comparison of "Attention Module" architectures: CSPNet (Wang et al. 2020), ELAN (Wang et al. 2022), C3K2 (used in YOLOv9) (Wang et al. 2024), and the novel R-ELAN (Tian et al. 2025) introduced with YOLOv12, which improved residual connections and enhanced feature aggregation, demonstrating superior performance

- Efficient architectural revisions: YOLOv12 replaces positional encoding with 7×7 depth-wise convolution for implicit spatial awareness. It also implements adaptive MLP ratio (1.2×) and shallow block stacking to balance the computational load, achieving 4.1 ms inference latency on V100 hardware.
- Optimized training framework: The model was trained over 600 epochs using SGD with cosine scheduling (initial lr = 0.01). The model also incorporates Mosaic-9 and Mixup augmentations with 12.8% mAP gain on COCO dataset, maintaining real-time performance through selective kernel convolution integration.

Figure 5b presents an architectural comparison of popular attention modules: CSPNet, ELAN, C3K2 (a case of GELAN), and the proposed R-ELAN. Brief summary of these modules is blow.

- CSPNet (Cross stage partial network): CSPNet enhances gradient flow by splitting feature maps into two paths, one for learning and one for propagation, reducing computational bottlenecks and improving inference speed. This model is visually depicted in Fig. 5b (leftmost module).
- ELAN (Efficient layer aggregation network): ELAN improves feature integration by aggregating multi-scale features efficiently, enhancing the model's ability to detect objects at various scales. However, as shown in Fig. 5b (second module), ELAN can introduce instability due to gradient blocking and lacks of residual connections, particularly in large-scale models.
- C3K2 (Compact GELAN): This module is a compact version of GELAN (Generalized Efficient Layer Aggregation Network) that offers a balance between computational efficiency and feature expressiveness, suitable for resource-constrained environments. The module is also illustrated in Fig. 5b (third module).
- R-ELAN (Residual ELAN): R-ELAN introduces residual connections and redesigns feature aggregation to address optimization challenges in attention-based models, combining the benefits of residual learning with efficient feature aggregation. As shown in Fig. 5b (rightmost module), R-ELAN applies a residual shortcut with a scaling factor (default 0.01) and processes the input through a transition layer, followed by a bottleneck structure for improved stability and performance.

The R-ELAN design, as depicted in Fig. 5b, addresses the limitations of ELAN by introducing residual connections and a revised aggregation approach. Unlike ELAN, which splits the input into two parts and processes them separately, R-ELAN applies a transition layer to adjust channel dimensions and processes the feature map through subsequent blocks before concatenation. This design mitigates gradient blocking and ensures stable convergence, particularly in large-scale models like YOLOv12-L and YOLOv12-X. The integration of residual connections and attention mechanisms in R-ELAN, as shown in Fig. 5b, highlights YOLOv12's architectural advancements in balancing efficiency and accuracy.

YOLO11: YOLO11 developed by Ultralytics represents the most recent version building upon the foundations established by its predecessors in the YOLO family. This latest iteration introduces several architectural innovations that enhance its performance across a wide spectrum of tasks as depicted in Fig. 6. The model incorporates the C3k2 (Cross Stage Partial with kernel size 2) block, which replaces the C2f block used in previous ver-sions, offering improved computational efficiency (Sapkota et al. 2024b). Additionally, YOLOv11 retains the SPPF (Spatial Pyramid Pooling-Fast) component and introduces the C2PSA (Convolutional block with Parallel Spatial Attention) module, collectively enhancing feature extraction capabilities (Sapkota et al. 2024a). These architectural enhancements enable YOLOv11 to capture intricate image details with greater precision, particularly in challenging scenarios involving small or occluded objects. The model's versatility is evident in its support for a broad range of computer vision tasks, including object detection, instance segmentation, pose estimation, image classification, and oriented bounding box (OBB) detection (Ultralytics 2024).

Empirical evaluations of YOLOv achieves a higher mean Average Precision (mAP) score on the COCO dataset while utilizing 22% fewer parameters compared to its YOLOv8m counterpart (Jegham et al. 2024). This reduction in parameter count contributes to faster model performance without significantly impacting overall accuracy. Furthermore, YOLOv11 exhibits inference times approximately 2% faster than YOLOv10, making it particularly well-suited for real-time applications. The model's efficiency extends across various deployment environments, including edge devices, cloud platforms, and systems supporting NVIDIA GPUs. YOLOv11 is available in multiple variants, ranging from nano to extra-large, catering to diverse computational requirements and use cases. These advancements position YOLOv11 as a state-of-the-art solution for industries requiring rapid and accurate image analysis, such as autonomous driving, surveillance, and industrial automation.

YOLOv10, incorporates advanced techniques like automated architecture search and more refined loss functions to enhance detection accuracy and speed, tailored for both edge and cloud computing environments. This version and its predecessors, YOLOv9 and YOLOv8, introduce substantial improvements in network architecture, such as the integration of cross-stage partial networks (CSPNets) and the use of transformer-based backbones for better feature extraction across different scales. YOLOv7 and YOLOv6 continued to build on these improvements by optimizing computational efficiency and expanding model scalability. Meanwhile, YOLOv5 introduced PyTorch support, which significantly enhanced the model's accessibility and adaptability, thus broadening its application in industry and academia. YOLOv4, on the other hand, marked a pivotal point in YOLO history by integrating

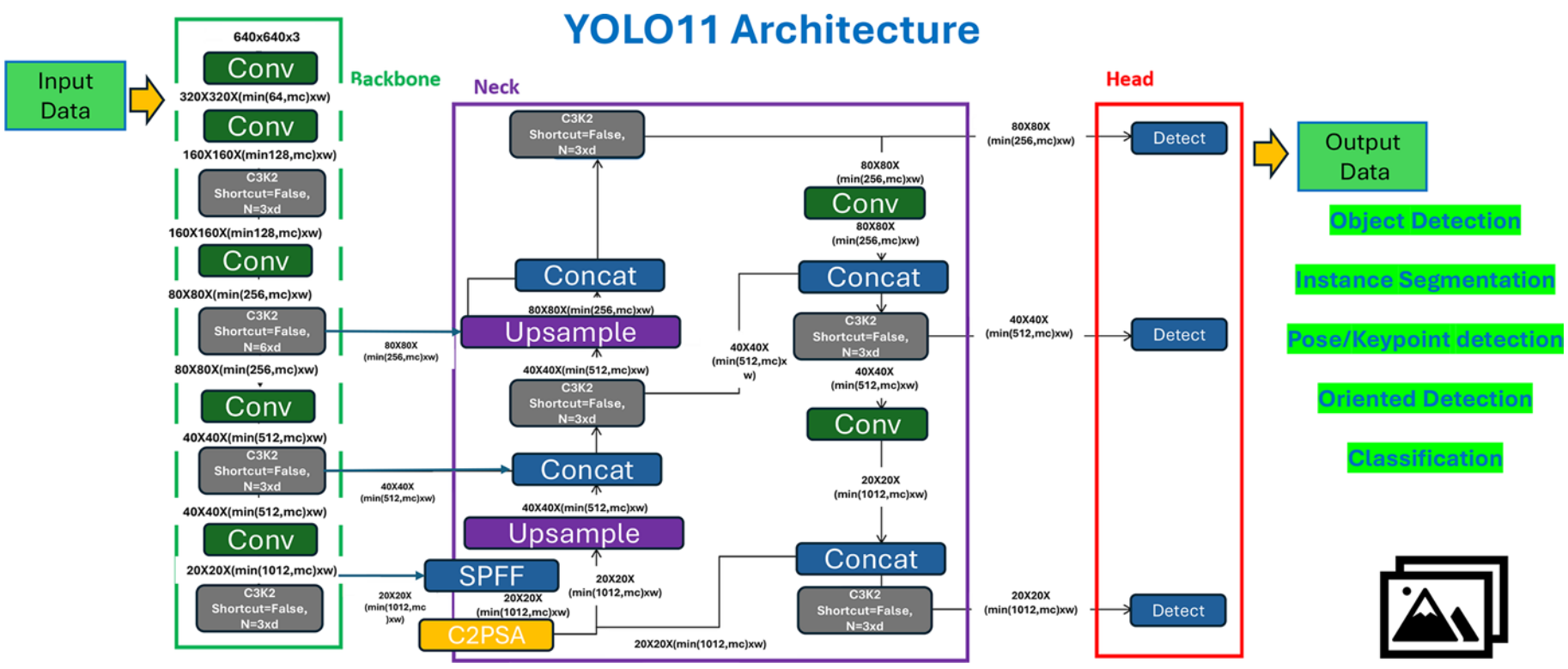

**Fig. 6** YOLOv11 architecture diagram: enhanced backbone with C3k2 blocks replacing C2f, SPPF for multi-scale feature extraction, C2PSA for attention mechanism, and optimized neck. Efficient feature processing through multiple scales, culminating in a detect head for multi-class object detection and localization

features like Mish activation and Cross-Stage Partial connections, setting new standards for speed and accuracy in real-time applications. The mid-generations, starting with YOLOv3, were notable for introducing multi-scale predictions and bounding box predictions across different layers, which greatly improved the model's ability to detect small objects-a long-standing challenge in earlier versions. YOLOv3 was also one of the first YOLO models to leverage deeper feature extractors like Darknet-53, which significantly boosted its performance over YOLOv2. YOLOv2 itself had introduced important features such as batch normalization and high-resolution classifiers, which enhanced the overall accuracy without compromising the speed. The original YOLO model, YOLOv1, was revolutionary, proposing a single-stage detection framework that unified the object detection process into a single neural network model.

Fig. 7a illustrates a sophisticated transformer-based model that simplifies the detection process by integrating dual label assignments and eliminating the need for non-max suppression (NMS), achieving a streamlined, end-to-end object detection. YOLOv9, as shown in Fig. 7b, introduces the Programmable Gradient Information (PGI) system to enhance model interpretability and robustness, significantly improving generalization across various tasks. Moving to YOLOv8 and YOLOv7, Figs. 7c and d respectively depict their architectures which incorporate elements like ELAN and CSPNet to boost performance and flexibility across computing devices. YOLOv6, highlighted in Fig. 7e, focuses on industry applications with enhancements in model quantization and real-time performance.

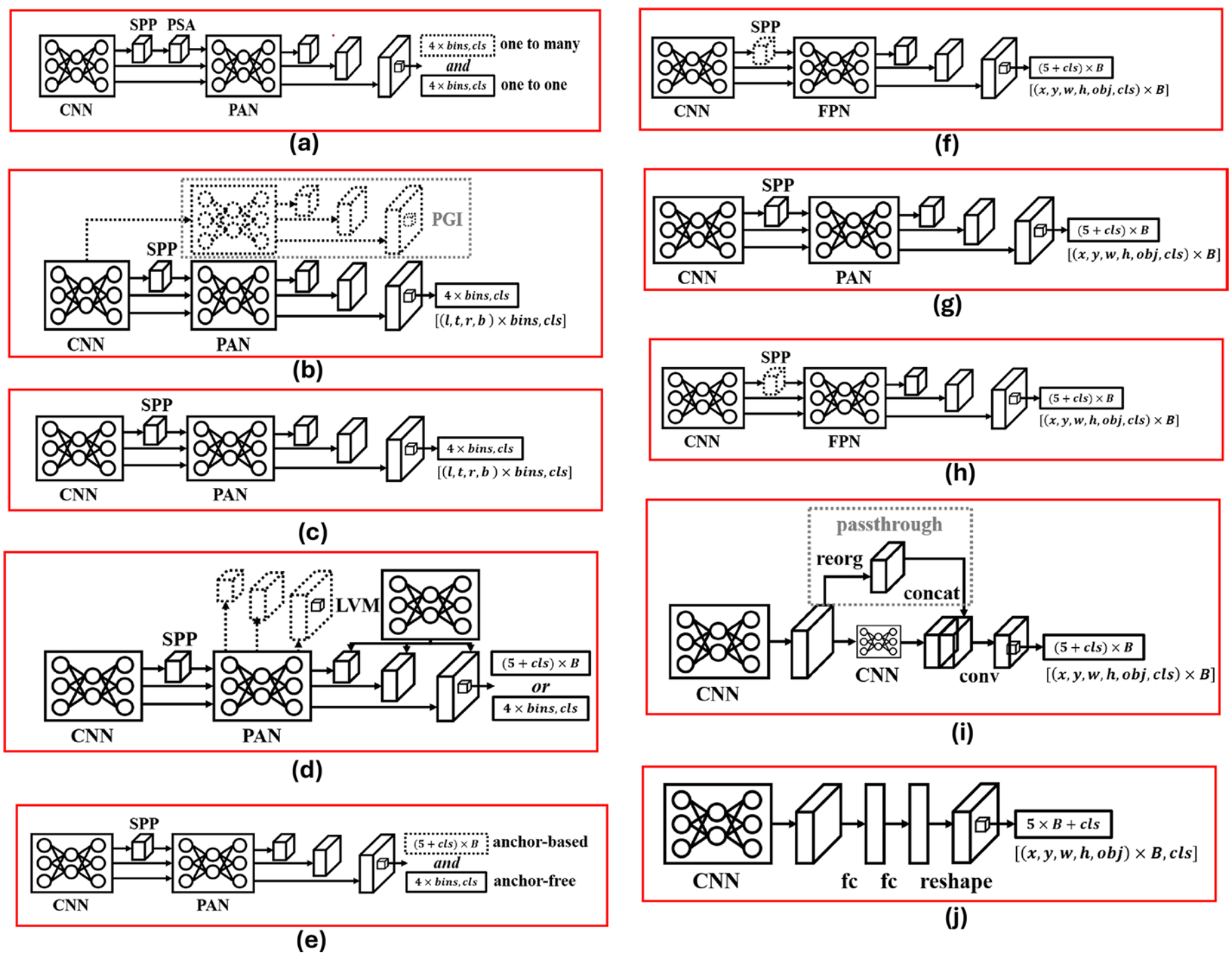

**Fig. 7** Simplified Architecture diagrams for: **a** YOLOv10; **b** YOLOv9; **c** YOLOv8; **d** YOLOv7; **e** YOLOv6; **f** YOLOv5; **g** YOLOv4; **h** YOLOv3; **i** YOLOv2; **j** YOLOv1. The diagrams are detailed in Wang et al. (2024)

YOLOv5, represented in Fig. 7f, marks a pivotal development with its adoption of PyTorch and improvements in training methods that enhance model accessibility and efficiency. In contrast, YOLOv4, shown in Fig. 7g, integrates technologies like CSPNet and Path Aggregation Network (PAN) (Liu et al. 2018) to optimize real-time detection. YOLOv3, visualized in Fig. 7h, introduces significant architectural changes with Darknet-53 and multi-scale predictions, which substantially enhance the detection of small objects. YOLOv2, depicted in Fig. 7i, advances the architecture with dimension clusters and fine-grained features, improving the model's efficiency and adaptability. Finally, YOLOv1, as outlined in Fig. 7j, revolutionizes object detection by integrating a single-stage detector that performs grid-based predictions in real-time, significantly reducing model complexity and enhancing speed.

### 4.2 YOLOv10, YOLOv9 and YOLOv8

YOLOv10 (Wang et al. 2024), developed at Tsinghua University, China, represents a breakthrough in the YOLO series for real-time object detection, achieving unprecedented performance. This version eliminates the need for non-maximum suppression (NMS) (Rothe et al. 2015), a traditional bottleneck in earlier models, thereby drastically reducing latency. YOLOv10 introduces a dual assignment strategy in its training protocol, which optimizes detection accuracy without sacrificing speed with the help of one-to-many and one-to-one label assignments, ensuring robust detection with lower latency (Li et al. 2023; Tian et al. 2024). The architecture of YOLOv10 includes several innovative components that enhance both computational efficiency and detection performance. Among these are lightweight classification heads (Bhagat et al. 2021) that reduce computational demands, spatial-channel decoupled downsampling to minimize information loss during feature reduction (Hu et al. 2023), and rank-guided block design that optimizes parameter use (Yang et al. 2023). These architectural advancements ensure that YOLOv10 operates synergistically across various scales-from YOLOv10-N (Nano) to YOLOv10-X (Extra Large), making it adaptable to diverse computational constraints and operational requirements (Wang et al. 2024). According to wang et al. (Wang et al. 2024), performance evaluations on benchmark datasets like MS-COCO (Lin et al. 2014) demonstrate that YOLOv10 not only surpasses its predecessors-YOLOv9 and YOLOv8-in both accuracy and efficiency but also sets new industry standards. For instance, YOLOv10-S substantially outperforms comparable models (e.g., YOLOv9 BASE, YOLOV9 Gelan, YOLOv8, YOLOv7) with an improved mAP and lower latency. This version also incorporates holistic efficiency-accuracy driven design, large-kernel convolutions, and partial self-attention modules, collectively improving the trade-off between computational cost and detection capability. The architecture diagrams of YOLOv10, YOLOv9, and YOLOv8 are summarized in Figs. 8, 9, and 10, respectively.

The YOLOv10 model offers various configurations, each tailored to specific performance needs within real-time object detection frameworks. Starting with YOLOv10-N (Nano), it demonstrates a rapid detection capability with a mAP of 38.5% at an exceptionally reduced latency to 1.84 ms, making it highly suitable for scenarios demanding quick responses. Progressing through the series, YOLOv10-S (Small) and YOLOv10-M (Medium) offer progressively higher mAP values of 46.3% and 51.1% at latencies of 2.49 ms and 4.74 ms, respectively, providing a balanced performance for versatile applications. The larger variants, YOLOv10-B (Balanced) and YOLOv10-L (Large), cater to environments requiring

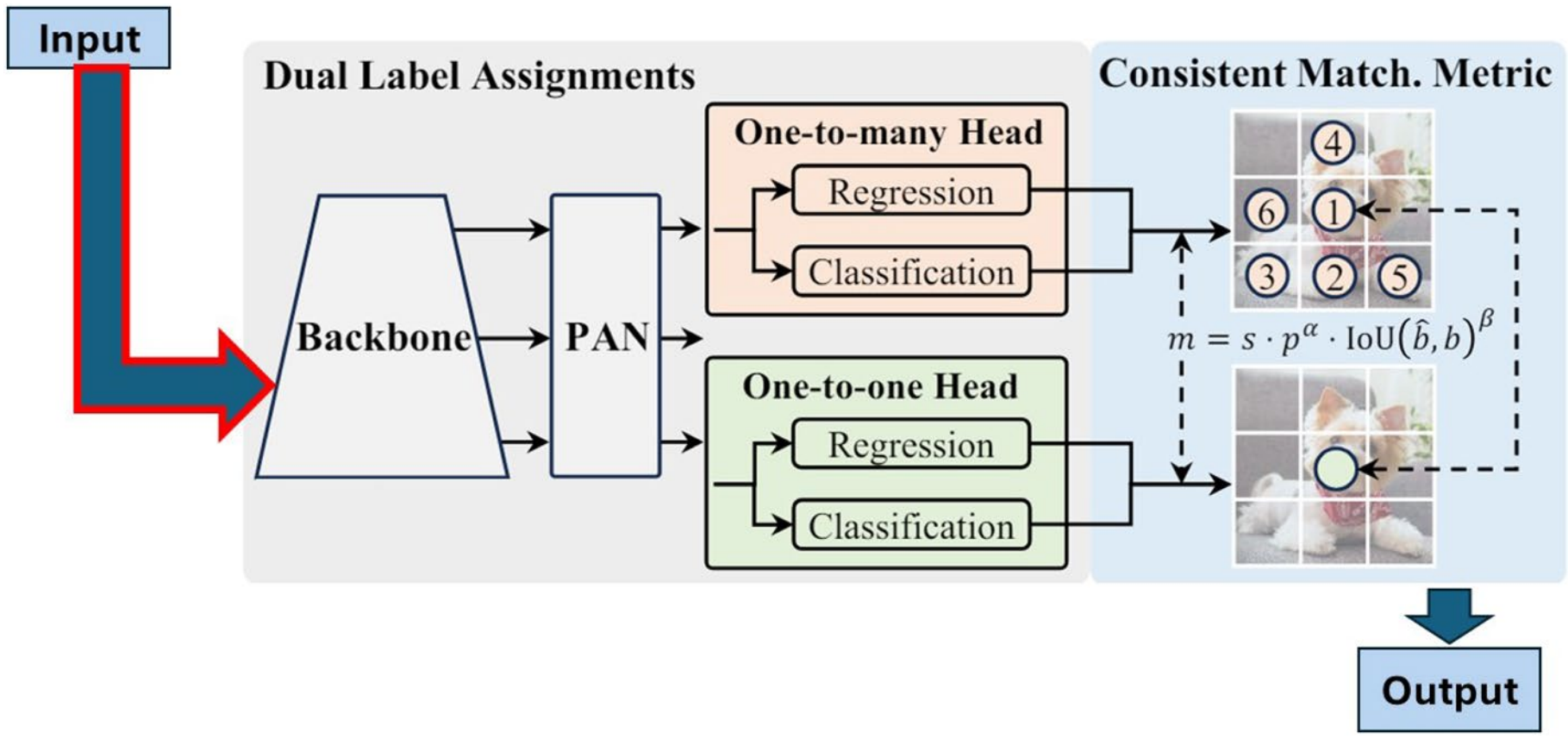

**Fig. 8** YOLOv10 architecture, which employs a dual label assignment strategy to improve detection accuracy. A backbone processes the input image, while PAN (Path Aggregation Network) enhances feature representation. Employed heads are (1) one-to-many head for regression and classification tasks, and (2) one-to-one head for precise localization (Wang et al. 2024)

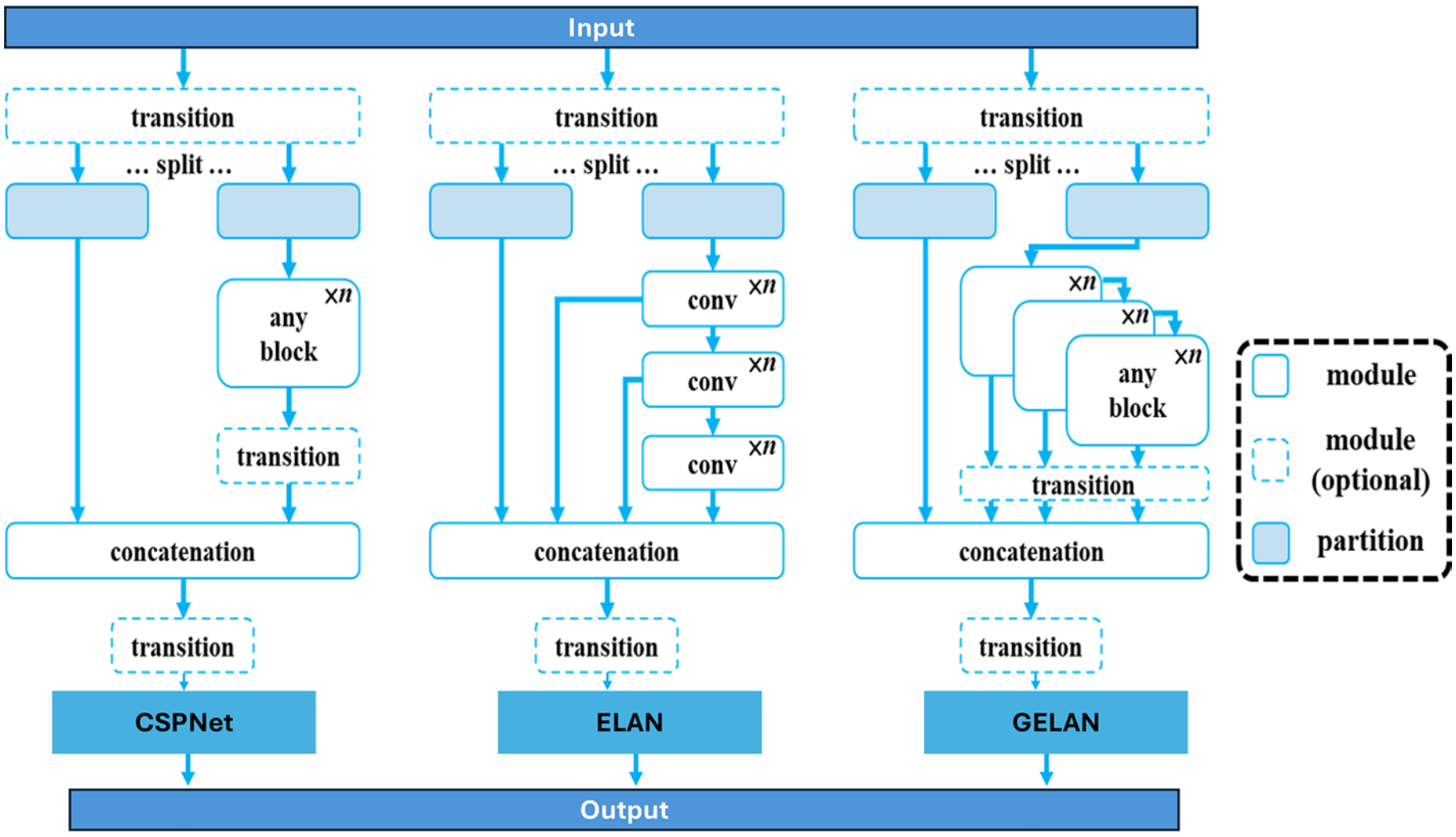

**Fig. 9** YOLOv9 architecture (Wang et al. 2024) with CSPNet, ELAN, and GELAN modules. CSPNet enhances gradient flow and reduces computational load through feature map partitioning. ELAN focuses on the linear aggregation of features for improved learning efficiency, while GELAN generalizes this approach to combine features from multiple depths and pathways, providing greater flexibility and accuracy in feature extraction

detailed detections, with mAPs of 52.5% and 53.2% and latencies of 5.74 ms and 7.28 ms respectively. The largest model, YOLOv10-X (Extra Large), excels with the highest mAP of 54.4% at a latency of 10.70 ms, designed for complex detection tasks where precision is paramount. These configurations underscore YOLOv10's adaptability across a spectrum of operational requirements.

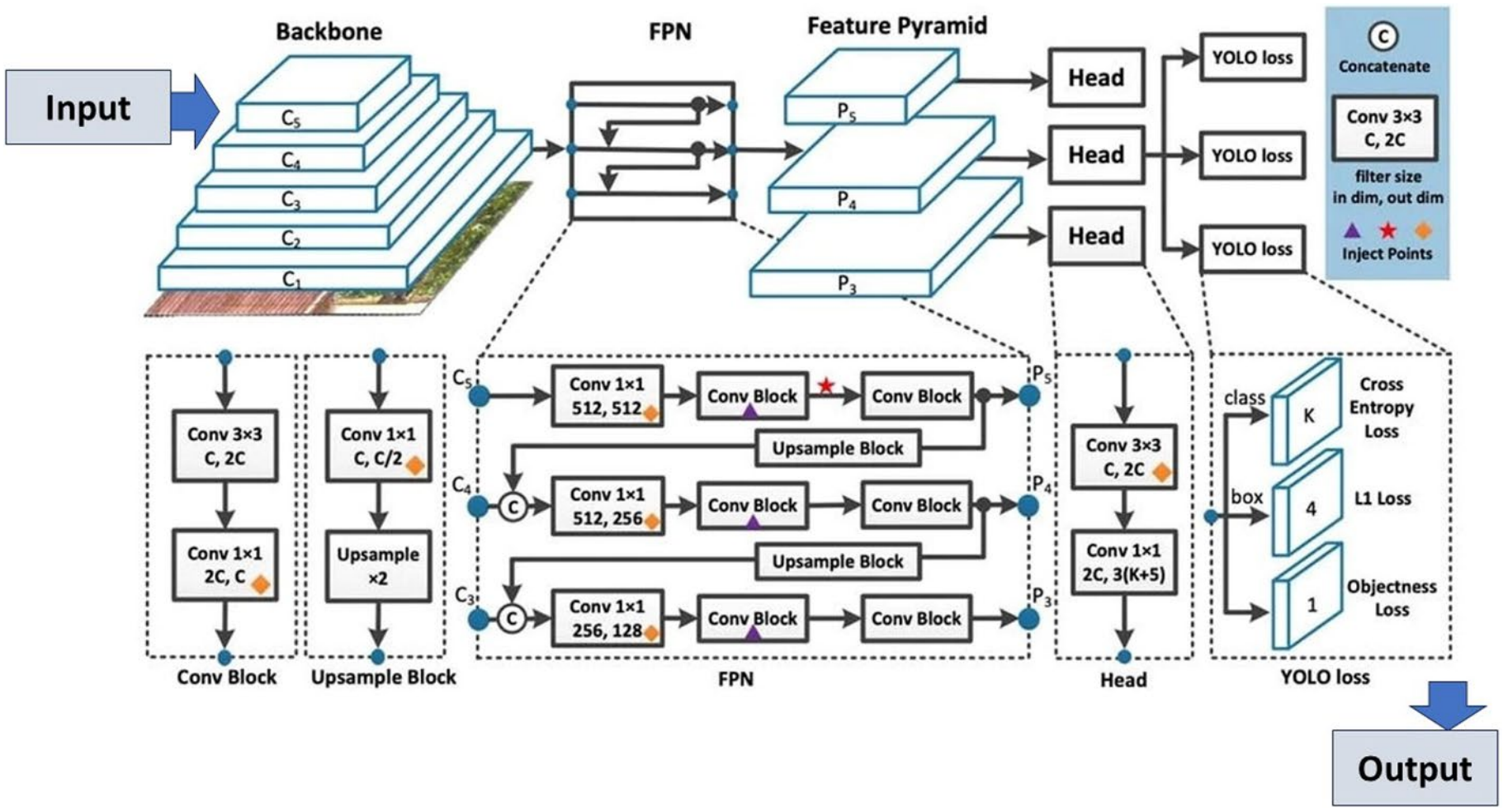

**Fig. 10** YOLOv8 architecture (Jocher et al. 2022): showcasing the key components and their connections. The backbone network processes the input image through multiple convolutional layers (C1 to C5), extracting hierarchical features. These features are then passed through the Feature Pyramid Network (FPN) to create a feature pyramid (P3, P4, P5), which enhances detection at different scales. The network heads perform final predictions, incorporating convolutional blocks and upsample blocks to refine features

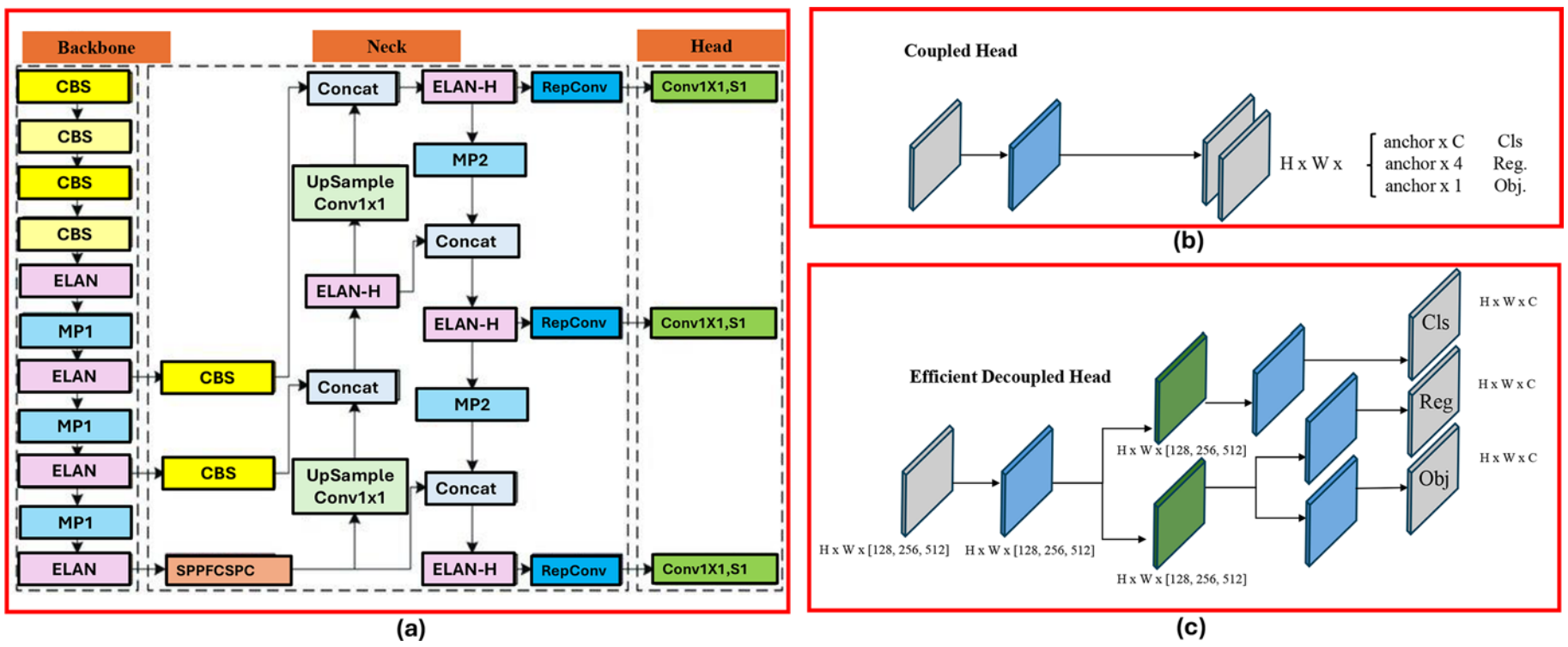

**Fig. 11** Architecture diagrams for **a** YOLOv7; **b** YOLOv6; and **c** YOLOv5

Reflecting on YOLO's evolution, starting from YOLOv1, which set the benchmark with an mAP of 63.4% and a latency of 45 ms, to the latest YOLOv10, significant technological strides have been evident. YOLOv10's predecessors, YOLOv9 and YOLOv8, display comparable mAP scores to YOLOv10 but with marginally higher latency, indicating the incremental enhancements YOLOv10 brings to the table. Specifically, YOLOv9 and YOLOv8 models, such as YOLOv9-N and YOLOv8-N, showcase mAPs of 39.5% and 37.3%, respectively, at latency indicative of their generational improvements. Meanwhile, the higher end of these series, YOLOv9-X, and YOLOv8-X, achieve mAPs of 54.4% and 53.9%, respectively, with YOLOv10 outperforming them in efficiency. The YOLO series, from YOLOv1 through YOLOv8, YOLOv9, and now YOLOv10, has continually advanced the frontier of real-time object detection, enhancing both the speed and accuracy of detections, and thus broadening the scope for practical applications in sectors like autonomous driving, surveillance, and real-time video analytics.

YOLOv9 (Wang et al. 2024) marks a significant advancement in real-time object detection by addressing the efficiency and accuracy challenges associated with earlier versions, particularly by mitigating information loss in deep neural processing. It introduces the innovative Programmable Gradient Information (PGI) and the Generalized Efficient Layer Aggregation Network (GELAN) architecture. These enhancements focus on preserving crucial information across the network, ensuring robust and reliable gradients that prevent data degradation, which is common in deep neural networks (Tishby and Zaslavsky 2015). Compared to its successor, YOLOv10, YOLOv9 sets a foundational stage by addressing the information bottleneck problem that typically hinders deep learning models. While YOLOv9's PGI strategically maintains data integrity throughout the processing layers, YOLOv10 builds upon this foundation by eliminating the need for NMS and further optimizing model architecture for reduced latency and enhanced computational efficiency. YOLOv10 also introduces dual assignment strategies for NMS-free training, significantly enhancing the system's response time without compromising accuracy, which reflects a direct evolution from the groundwork laid by YOLOv9's innovations (Zhang et al. 2023). Furthermore, YOLOv9's GELAN architecture represents a pivotal improvement in network design, offering a flexible and efficient structure that effectively integrates multi-scale features. While GELAN contributes significantly to YOLOv9's performance, YOLOv10 extends these architectural improvements to achieve even greater efficiency and adaptability (Chien et al. 2024). It reduces computational overhead and increases the model's applicability to various real-time scenarios, showcasing an advanced level of refinement that leverages and enhances the capabilities introduced by YOLOv9.

YOLOv8 was released in January 2023 by Ultralytics, marking a significant progression in the YOLO series with an introduction of multiple scaled versions designed to cater to a wide range of applications (Ultralytics 2024a, b). These versions included YOLOv8n (nano), YOLOv8s (small), YOLOv8m (medium), YOLOv8l (large), and YOLOv8x (extra-large), each optimized for specific performance and computational needs. This flexibility made YOLOv8 highly versatile, supporting many vision tasks such as object detection, segmentation, pose estimation, tracking, and classification, significantly broadening its application scope in real-world scenarios (Ultralytics 2024b). The architecture of YOLOv8 underwent substantial refinements to enhance its detection capabilities. It retained a similar backbone to YOLOv5 but introduced modifications in the CSP Layer, now evolved into the C2f module-a cross-stage partial bottleneck with dual convolutions that effectively combine high-level features with contextual information to bolster detection accuracy. YOLOv8 transitioned to an anchor-free model with a decoupled head, allowing independent processing of object detection, classification, and regression tasks, which, in turn, improved overall model accuracy (Ultralytics 2024). The output layer employed a sigmoid activation function for objectness scores and softmax for class probabilities, enhancing the precision of bounding box predictions. YOLOv8 also integrated advanced loss functions like CIoU (Du et al. 2021) and Distribution Focal Loss (DFL) (Xu et al. 2022) for bounding-box optimization and binary cross-entropy for classification, which proved particularly effective in enhancing detection performance for smaller objects. YOLOv8's architecture, demonstrated in detailed diagrams, features the modified CSPDarknet53 backbone with the innovative C2f module, augmented by a spatial pyramid pooling fast (SPPF) layer that accelerates computation by pooling features into a fixed-size map. This model also introduced a semantic segmentation variant, YOLOv8-Seg, which utilized the backbone and C2f module, followed by two segmentation heads designed to predict semantic segmentation masks efficiently. This segmentation model achieved state-of-the-art results on various benchmarks while maintaining high speed and accuracy, evident in its performance on the MS COCO dataset where YOLOv8x reached an AP of 53.9% at 640 pixels image size-surpassing the 50.7% AP of YOLOv5-with a remarkable speed of 280 FPS on an NVIDIA A100 using TensorRT. As we progress backwards through the YOLO series, from YOLOv10 to YOLOv8 and soon to YOLOv7, these architectural and functional advancements highlight the series' evolutionary trajectory in optimizing real-time object detection networks.

YOLOv8 was released in January 2023 by Ultralytics, marking a significant progression in the YOLO series with an introduction of multiple scaled versions designed to cater to a wide range of applications (Ultralytics 2024a, b). These versions included YOLOv8n (nano), YOLOv8s (small), YOLOv8m (medium), YOLOv8l (large), and YOLOv8x (extra-large), each optimized for specific performance and computational needs. This flexibility made YOLOv8 highly versatile, supporting many vision tasks such as object detection, segmentation, pose estimation, tracking, and classification, significantly broadening its application scope in real-world scenarios (Ultralytics 2024b). YOLOv8's architecture underwent significant upgrades to boost its detection performance, maintaining a backbone similar to YOLOv5 but enhancing it with the evolved C2f module, a cross-stage partial bottleneck with dual convolutions. This module integrates high-level features with contextual information, improving accuracy. The model transitioned to an anchor-free system with a decoupled head for independent objectness, classification, and regression tasks, enhancing accuracy (Ultralytics 2024). The output layer now uses sigmoid for objectness and softmax for class probabilities, refining bounding box precision. Additionally, YOLOv8 employs CIoU (Du et al. 2021), Distribution Focal Loss (DFL) (Xu et al. 2022) for bounding-box optimization, and binary cross-entropy for classification, significantly boosting performance, particularly for smaller objects.

This model also introduced a semantic segmentation variant, YOLOv8-Seg (Yue et al. 2023), which utilized the backbone and C2f module, followed by two segmentation heads designed to predict semantic segmentation masks efficiently. This segmentation model achieved state-of-the-art results on various benchmarks while maintaining high speed and accuracy, evident in its performance on the MS COCO dataset where YOLOv8x reached an AP of 53.9% at 640 pixels image size-surpassing the 50.7% AP of YOLOv5-with a remarkable speed of 280 FPS on an NVIDIA A100 using TensorRT. As we progress backwards through the YOLO series, from YOLOv10 to YOLOv8 and soon to YOLOv7, these architectural and functional advancements highlight the series' evolutionary trajectory in optimizing real-time object detection networks.

### 4.3 YOLOv7, YOLOv6 and YOLOv5

The YOLOv7 model introduces enhancements in object detection tailored for drone-captured scenarios, particularly through the Transformer Prediction Head (TPH-YOLOv5) variant (Zhu et al. 2021), which emphasizes improvements in handling scale variations and densely packed objects (Wang et al. 2023). By incorporating TPH and the Convolu-tional Block Attention Module (CBAM) (Woo et al. 2018), YOLOv7 substantially boosts its capacity to focus on relevant regions in cluttered environments. These features particularly enhance the model's ability to detect objects across varied scales, an essential trait for drone applications where altitude changes affect object size perception drastically. The model integrates sophisticated strategies like multi-scale testing (Hnewa and Radha 2023) and a self-trained classifier, which refines its performance on challenging categories by specifically addressing common issues in drone imagery, such as motion blur and occlusion. These adaptations have shown notable improvements, with YOLOv7 achieving competitive results in drone-specific datasets and challenges (Bai et al. 2024). The model's adaptability and robustness in such specialized conditions demonstrate its potential beyond conventional settings, catering effectively to next-generation applications like urban surveillance and wildlife monitoring.

YOLOv6 emerges as a robust solution in industrial applications by delivering a finely balanced trade-off between speed and accuracy, crucial for deployment across various hardware platforms (Li et al. 2022). It iterates on previous versions by incorporating cutting-edge network designs, training strategies, and quantization techniques to enhance its efficiency and performance significantly. This model has been optimized for diverse operational requirements with its scalable architecture, ranging from YOLOv6-N to YOLOv6-X, each offering different performance levels to suit specific computational budgets (Sirisha et al. 2023). Significant innovations in YOLOv6 include advanced label assignment techniques and loss functions that refine the model's predictive accuracy and operational efficiency. By leveraging state-of-the-art advancements in machine learning, YOLOv6 not only excels in traditional object detection metrics but also sets new standards in throughput and latency, making it exceptionally suitable for real-time applications in industrial and commercial domains.

YOLOv6 and YOLOv7 each introduced innovative features that build on the foundation set by YOLOv5. YOLOv6, released in October 2021, introduced lightweight nano models optimized for mobile and CPU environments alongside a more effective backbone for improved small object detection. YOLOv7 further advanced this development by incorporating a new backbone network, PANet (Wang et al. 2019), enhancing feature aggregation and representation, and introducing the CIOU loss function for better object scaling and aspect ratio handling. YOLO-v6 significantly shifts the architecture to an anchor-free design, incorporating a self-attention mechanism to better capture long-range dependencies and employing adaptive training techniques to optimize performance during training (Zhang et al. 2021). These versions collectively push the boundaries of object detection performance, emphasizing speed, accuracy, and adaptability across various deployment scenarios.

YOLOv5 has significantly contributed to the YOLO series evolution, focusing on user-friendliness and performance enhancements (Ultralytics 2024a, b). Its introduction by Ultralytics brought a streamlined, accessible framework that lowered the barriers to implementing high-speed object detection across various platforms. YOLOv5's architecture incorporates a series of optimizations including improved backbone, neck, and head designs which collectively enhance its detection capabilities. The model supports multiple size variants, facilitating a broad range of applications from mobile devices to cloud-based systems (Ultralytics 2024a). YOLOv5's adaptability is further evidenced by its continuous updates and community-driven enhancements, which ensure it remains at the forefront of object detection technologies. This version stands out for its balance of speed, accuracy, and utility, making it a preferred choice for developers and researchers looking to deploy state-of-the-art detection systems efficiently.

YOLOv5 marks a significant evolution in the YOLO series, focusing on production-ready deployments with streamlined architecture for real-world applications. This version emphasizes reducing the model's complexity by refining its layers and components, enhancing its inference speed without sacrificing detection accuracy. The backbone and feature extraction layers were optimized to accelerate processing, and the network's architecture was simplified to facilitate faster data throughput. Importantly, YOLO v5 enhances its deployment flexibility, catering to edge devices with limited computational resources through model modularity and efficient activations. These architectural refinements ensure YOLO v5 operates effectively in diverse environments, from high-resource servers to mobile devices, making it a versatile tool in the arsenal of object detection technologies.

### 4.4 YOLOv4, YOLOv3, YOLOv2 and YOLOv1

The introduction of YOLOv4 (Bochkovskiy et al. 2020) in 2020 marked the latest developments, employing CSPDarknet-53 (Mahasin and Dewi 2022) as its backbone. This modified version of Darknet-53 uses Cross-Stage Partial connections to reduce computational demands while enhancing learning capacity.

YOLOv4 incorporates innovative features such as Mish activation (Misra 2019), replacing traditional ReLU to maintain smooth gradients, and utilizes new data augmentation techniques such as Mosaic and CutMix (Yun et al. 2019). Additionally, it introduces advanced regularization methods, including DropBlock regularization (Ghiasi et al. 2018) and Class Label Smoothing to prevent overfitting (Müller et al. 2019), alongside optimization strategies termed BoF (Bag of Freebies) (Zhang et al. 2019) and BoS (Bag of Specials) that enhance training and inference efficiency. "YOLOv3, introduced in 2018 before the release of YOLOv4, employed the Darknet-53 architecture, incorporating principles of residual learning. Initially trained on ImageNet, this version excelled in detecting objects of various sizes due to its multi-scale detection capabilities within the architecture. The subsequent development of YOLOv4 built upon the success of YOLOv3, further enhancing the framework's robustness and accuracy.

YOLOv3 (Redmon and Farhadi 2018) improved detection accuracy, especially for small objects, by using three different scales for detection, thereby capturing essential features at various resolutions. Earlier, YOLOv2 and the original YOLO (YOLOv1) laid the groundwork for these advancements (Redmon et al. 2016).

Earlier, YOLOv2 and the original YOLO (YOLOv1) laid the groundwork for these advancements. Released in 2016, YOLOv2 introduced a new 30-layer architecture with anchor boxes from Faster R-CNN and batch normalization to speed up convergence and enhance model performance. YOLOv1, debuting in 2015 by Joseph Redmon, revolutionized object detection with its single-shot mechanism that predicted bounding boxes and class probabilities in one network pass, utilizing a simpler Darknet-19 architecture. This initial approach significantly accelerated the detection process, establishing the foundational techniques that would be refined in later versions of the YOLO series. YOLOv4 and YOLOv3, showcasing their advanced architectures and features, are illustrated in Fig. 12a and b, respectively, while YOLOv2 and YOLOv1 are depicted in Fig. 13a and b, showcasing the foundational developments in the series.

### 4.5 Alternative versions derived from YOLO

Several alternative YOLO models have been developed from different versions, with the five primary ones being YOLO-NAS, YOLO-X, YOLO-R, DAMO-YOLO, and Gold-YOLO.

#### 4.5.1 YOLO-NAS

YOLO-NAS, developed by Deci AI, represents a significant advancement in object detection technology (Terven et al. 2023). This model leverages Neural Architecture Search (NAS) (Ren et al. 2021) to address limitations of previous YOLO iterations such as YOLOv4, YOLOv5, YOLOv6 and YOLOv7 (Mithun and Jawhar 2024). YOLO-NAS introduces a quantization-friendly basic block, enhancing performance with minimal precision loss post-quantization. The architecture employs quantization-aware blocks and selective quantization, resulting in superior object detection capabilities. Notably, when converted to INT8, the model experiences only a slight precision drop, outperforming its predecessors. YOLO-NAS utilizes sophisticated training schemes and post-training quantization techniques, further improving its efficiency (Terven et al. 2023). The model is pre-trained on datasets such as COCO, Objects365, and Roboflow 100, making it suitable for various downstream object detection tasks. YOLO-NAS is available in three variants: Small (s), Medium (m), and Large (l), each optimized for different computational requirements. These variants offer a balance between Mean Average Precision (mAP) and latency, with the INT-8 versions demonstrating impressive performance metrics. The architecture of YOLO-NAS

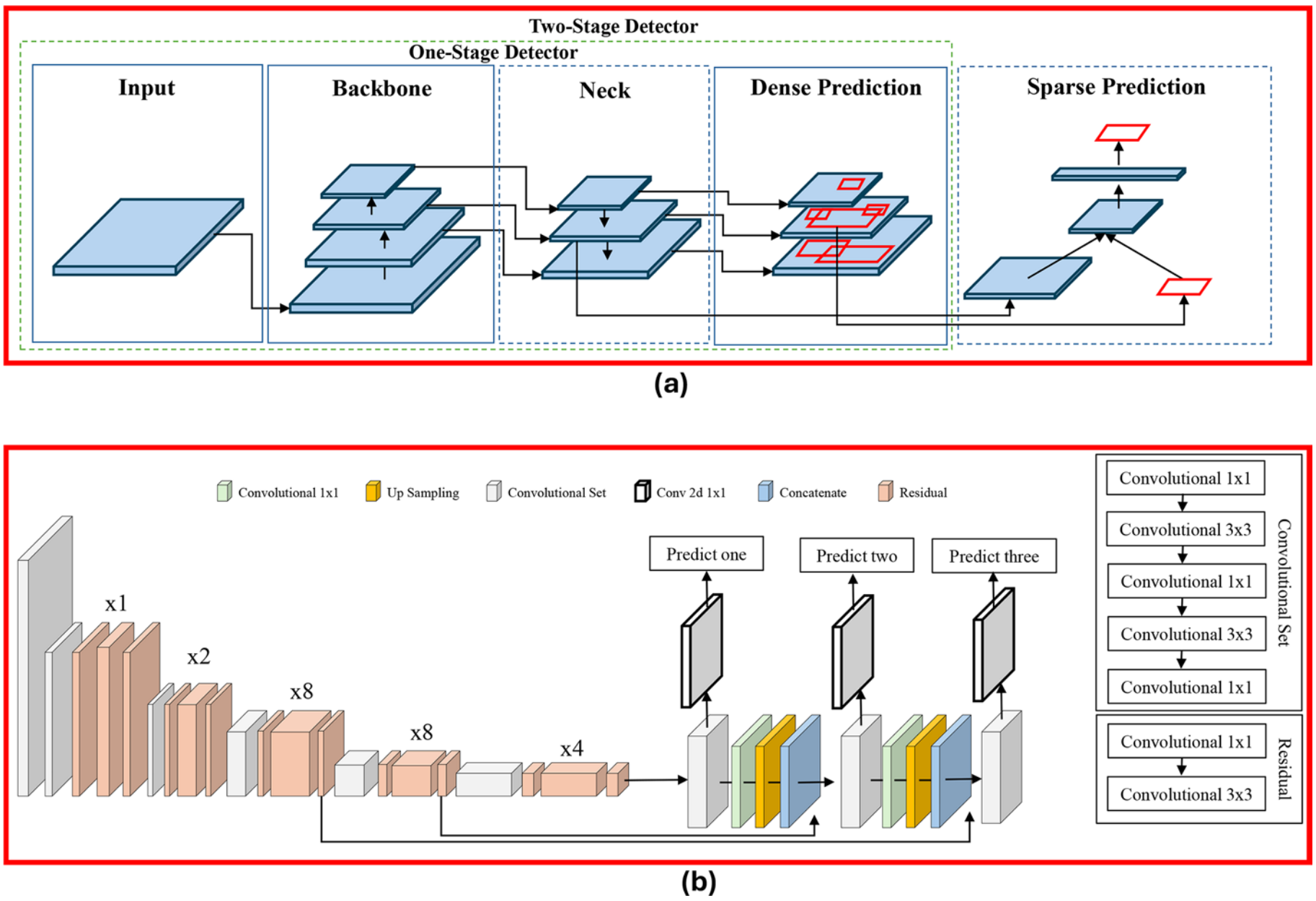

**Fig. 12** Comparison of YOLOv4 (Bochkovskiy et al. 2020) and YOLOv3 (Redmon and Farhadi 2018) architectures. **a** YOLOv4 architecture shows a two-stage detector with a backbone, neck, dense prediction, and sparse prediction modules. **b** YOLOv3 architecture features convolutional and upsampling layers that lead to multi-scale predictions. This highlights the structural advancements in object detection between the two versions

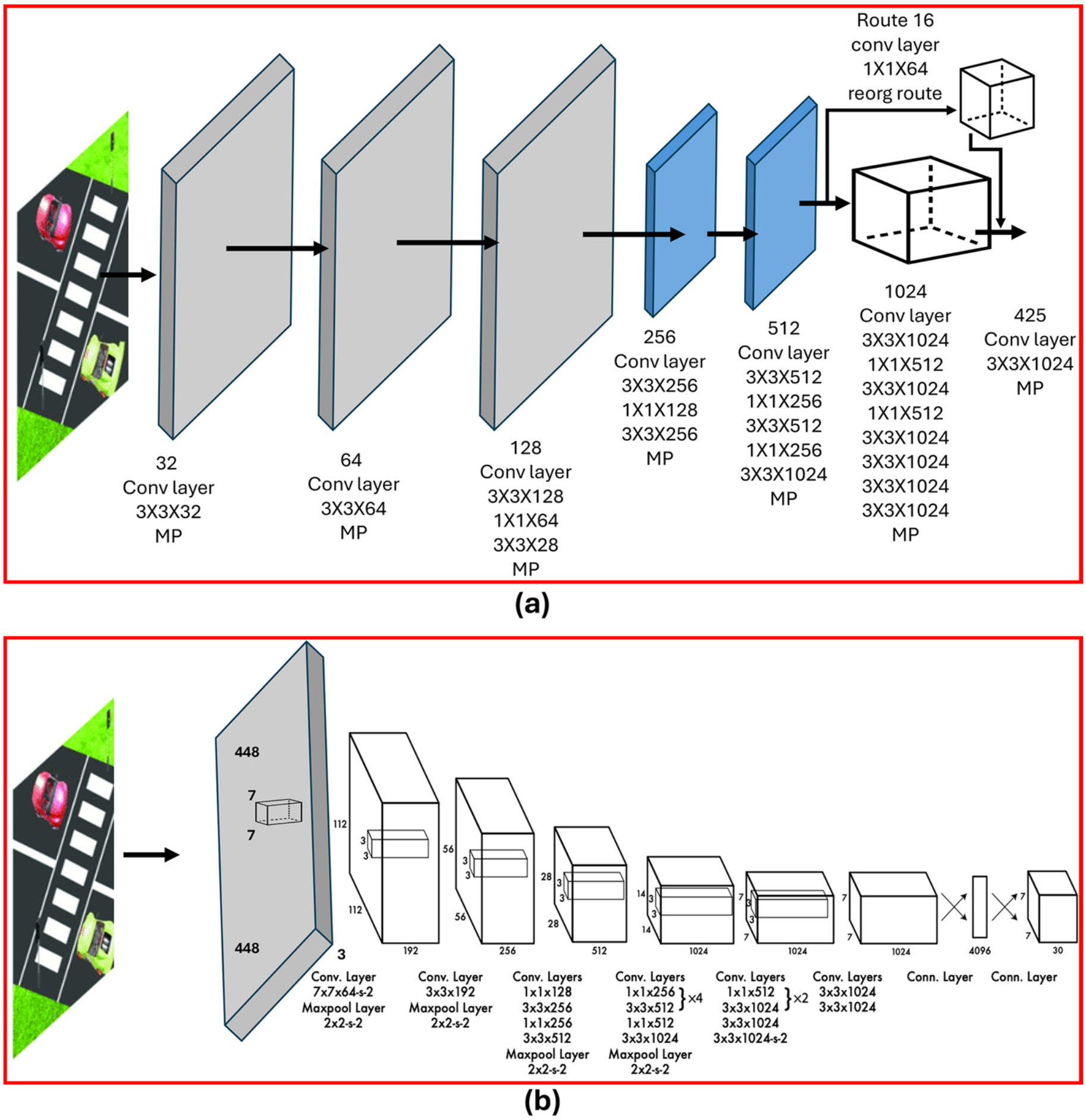

**Fig. 13** **a** YOLOv2 architecture (Redmon and Farhadi 2017), illustrating improvements such as the use of batch normalization, higher resolution input, and anchor boxes; **b** YOLOv1 architecture (Redmon et al. 2016), showing the sequence of convolutional layers, max-pooling layers, and fully connected layers used for object detection. This model performs feature extraction and prediction in a single unified step, aiming for real-time performance

(Fig. 14a) supports inference, validation, and export modes, though it does not support training. YOLO-NAS's innovative design and superior performance position it as a critical tool for developers and researchers in the field of computer vision.

### 4.5.2 YOLO-X

YOLOX, developed by Megvii Technology, represents a significant advancement in the YOLO series of object detectors. This model introduces several key improvements to enhance performance and efficiency. YOLOX adopts an anchor-free approach, departing from the anchor-based methods of its predecessors such as YOLOv6, YOLOv7 and YOLOv8 (Ge 2021). It incorporates a decoupled head, separating classification and regres-sion tasks to address the known conflict between these objectives in object detection (Zhang et al. 2022). The model also implements SimOTA, an advanced label assignment strategy, further improving its detection capabilities (Liu and Sun 2022). Architecturally (Fig. 14b), YOLOX-DarkNet53 builds upon the YOLOv3-SPP baseline, incorporating enhancements such as EMA weights updating, cosine learning rate scheduling, IoU loss, and an IoU-aware branch. The decoupled head consists of a 1x1 convolution layer for channel dimension reduction, followed by two parallel branches with two 3x3 convolution layers each (Ashraf et al. 2024). This design significantly improves convergence speed and is crucial for end-to-end detection performance. YOLOX demonstrates superior performance across various model sizes. The YOLOX-L variant achieves 50.0% AP on COCO at 68.9 FPS on Tesla V100, surpassing YOLOv5-L by 1.8% AP. Even the lightweight YOLOX-Nano, with only 0.91M parameters and 1.08 GFLOPs, attains 25.3% AP on COCO, outperforming NanoDet by 1.8% AP. These advancements position YOLOX as a state-of-the-art object detector, balancing accuracy and efficiency across a wide range of model scales (Ge 2021).

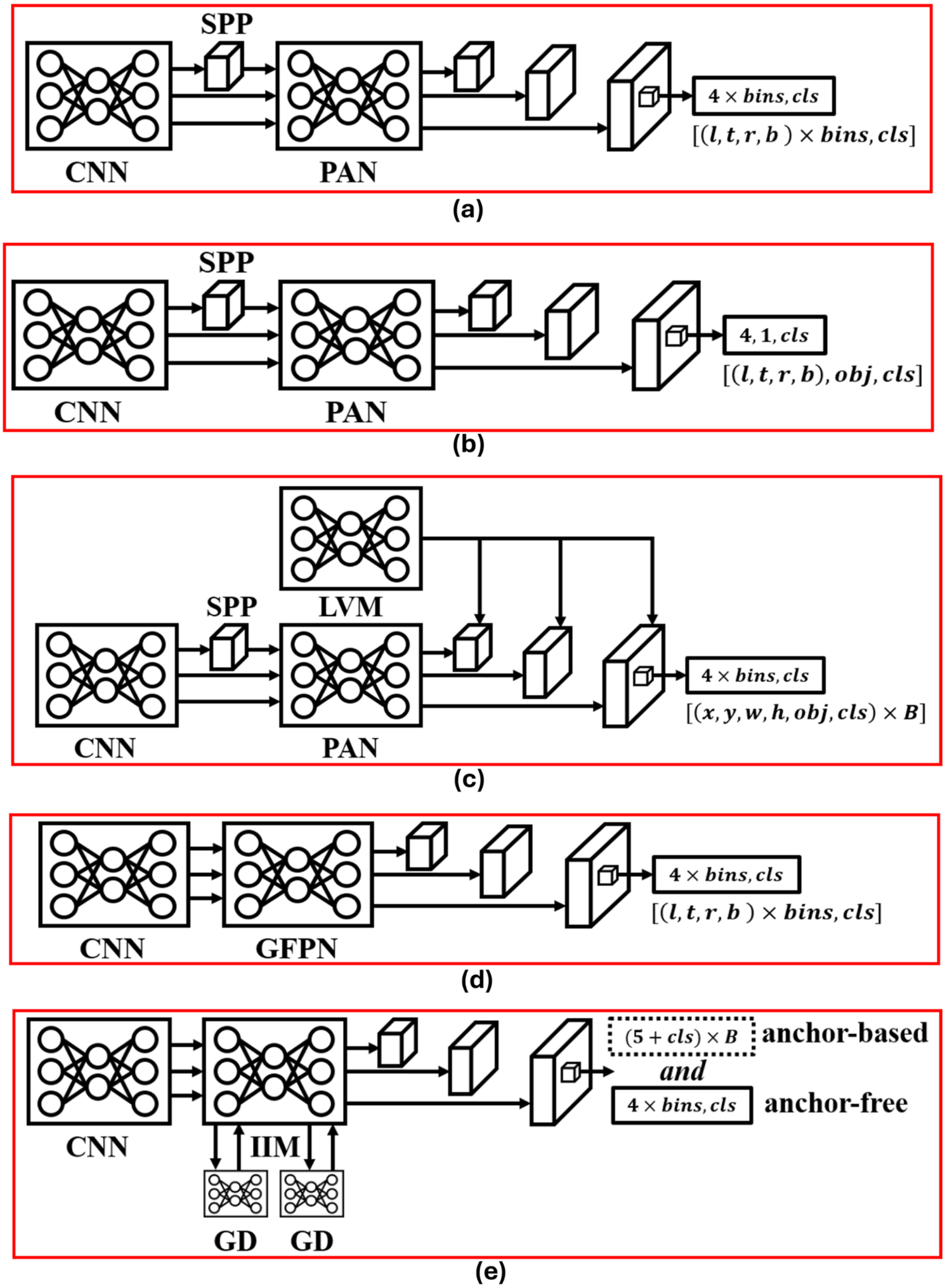

**Fig. 14** Architecture diagram of **a** YOLONAS; **b** YOLOX; **c** YOLOR; **d** DAMO YOLO; **e** GOLD YOLO

#### 4.5.3 YOLO-R

YOLOR (You Only Learn One Representation) is a novel object detection algorithm developed by Chang et al. (2023). Unlike other YOLO versions, YOLOR introduces a novel approach to multi-task learning by unifying implicit and explicit knowledge representation (Andrei-Alexandru et al. 2022). The algorithm's core concept is inspired by human cognition, aiming to process multiple tasks simultaneously given a single input. YOLOR's architecture (Fig. 14c) incorporates three key components: kernel space alignment, prediction refinement, and a CNN with multi-task learning capabilities. This unified network encodes both implicit knowledge (learned subconsciously from deep layers) and explicit knowledge (obtained from shallow layers and clear metadata), resulting in a more refined and generalized representation (Sun et al. 2024; Chang et al. 2023). Compared to other YOLO algorithms such as YOLOv9, YOLOv5 or YOLOv3, YOLOR significantly improves both speed and accuracy. It achieves comparable object detection accuracy to Scaled YOLOv4 while increasing inference speed by 88%, making it one of the fastest object detection algorithms in modern computer vision. On the MS COCO dataset, YOLOR outperforms PP-YOLOv2 by 3.8% in mean average precision at the same inference speed (Chang et al. 2023).

#### 4.5.4 DAMO-YOLO

DAMO-YOLO is developed by Alibaba's DAMO Academy, which significantly enhances performance by integrating novel technologies like Neural Architecture Search (NAS), a reparameterized Generalized-FPN (RepGFPN), and lightweight head architectures (Fig. 14d) with AlignedOTA label assignment and distillation enhancement (Xu et al. 2022). Leveraging MAE-NAS, the model employs a heuristic, training-free approach to architect detection backbones under strict latency and performance constraints, generating efficient structures akin to ResNet and CSPNet (Terven et al. 2023). The neck design emphasizes a robust "large neck, small head" architecture, optimizing the fusion of high-level semantic and low-level spatial features through an enhanced FPN. This approach effectively balances computational efficiency and detection accuracy, particularly notable in its deployment across various model scales, from lightweight versions for edge devices to more robust configurations for general industry applications. DAMO-YOLO's architectural prowess is showcased through impressive performance metrics, achieving mAP scores ranging from 43.6 to 51.9 on COCO datasets with relatively low latency on T4 GPUs. Moreover, the model's lightweight variants demonstrate substantial efficiency on edge devices, underscoring its adaptability and broad application potential. Such capabilities are further augmented by strategic enhancements in label assignment and knowledge distillation, addressing common challenges in object detection like label misalignment and model generalization.

### 4.5.5 Gold-YOLO

Gold-YOLO was developed by the team at Huawei Noah's Ark Lab to significantly enhance multi-scale feature fusion through an innovative Gather-and-Distribute (GD) mechanism (Wang et al. 2024). This mechanism, which utilizes convolution and self-attention operations, was implemented to optimize the exchange and integration of information across different levels of the feature pyramid. This approach facilitated a more effective balance between latency and detection accuracy (Wang et al. 2024). Furthermore, an MAE-style unsupervised pretraining was incorporated into the YOLO-series for the first time, which was reported to enhance learning efficiency and overall model performance. Gold-YOLO's architecture (Fig. 14e) aimed to address the limitations inherent in traditional Feature Pyramid Networks (FPNs) by preventing recursive information loss and enabling more direct and efficient feature fusion. This was achieved by a method where features from all levels were gathered to a central processing node, enhanced, and then redistributed, ensuring enriched feature maps that retained critical information across scales. The effectiveness of this novel design was demonstrated through good performance metrics; Gold-YOLO achieved a 39.9% AP on the COCO dataset with high throughput speeds on a T4 GPU, surpassing previous state-of-the-art models like YOLOv6−3.0-N. The contributions made by this paper were significant, as they not only enhanced the YOLO model's capabilities to handle various object sizes and complexities but also established a new benchmark for the integration of advanced neural network techniques with traditional convolutional architectures for real-time applications.

## 5 Applications

YOLO has many real-time practical applications (Vijayakumar and Vairavasundaram 2024; Chen et al. 2023), such as autonomous vehicles and traffic safety where the technology is used for obstacle detection, pedestrian pose estimation for intention prediction, and traffic sign recognition, enhancing safety and navigation (Gheorghe et al. 2024). Similarly, YOLO is employed in healthcare for detecting anomalies in medical images, aiding in accurate and efficient diagnostics (Vijayakumar and Vairavasundaram 2024; Ragab et al. 2024). Additionally, industrial manufacturing benefits from YOLO's capabilities, with applications in quality control and defect detection (Li et al. 2022; Hussain 2023), in surveillance for intrusion detection and anomaly identification (Mohod et al. 2022), while in agriculture, it supports crop stress detection, monitoring, and robotic fruit harvesting, among other use cases (Alibabaei et al. 2022; Badgujar et al. 2024; Wang et al. 2021).

The remainder of this section is categorically divided into five key application areas where YOLO models have demonstrated significant impact: Autonomous Vehicles and Traffic Safety, Healthcare and Medical Imaging, Security and Surveillance, Industrial Manufacturing, and Agriculture.

### 5.1 Autonomous vehicles and traffic safety

Each YOLO version has been pivotal in advancing the capabilities of autonomous vehicles and traffic safety by providing highly efficient and accurate real-time detection systems. Each iteration of YOLO has brought improvements that enhance the vehicle's ability to perceive its environment quickly and accurately, which is critical for safe navigation and decision-making (Benjumea et al. 2021; Malligere Shivanna and Guo 2024). Starting with YOLOv1 (Redmon et al. 2016), the YOLO algorithm revolutionized the approach by performing detection tasks directly from full images in a single network pass, allowing for the detection of objects at a remarkable speed (Sarda et al. 2021). This initial model was pivotal, setting a high standard for real-time object detection and establishing a framework that future versions would build upon. Subsequent iterations, including YOLOv2 and YOLOv3, continued to refine this approach by introducing concepts such as real-time multi-scale processing and improved anchor box adjustments, which enhanced the accuracy and robustness of the detections. These versions were particularly adept at handling the variable scales of objects seen in driving environments-from nearby pedestrians to distant road signs-making them invaluable for autonomous driving applications. YOLOv4 and later versions further pushed the boundaries by integrating advanced neural network techniques and optimizations that improved detection accuracy while maintaining the high-speed processing necessary for real-time applications (Cai et al. 2021; Zhao et al. 2022). These advancements in YOLO technology have not only bolstered the capabilities of autonomous vehicles in terms of environmental perception and decision-making but have also significantly contributed to advancements in automotive safety and operational reliability (Woo et al. 2022).

Ye et al. (2022) developed an end-to-end adaptive neural network control for autonomous vehicles that predicts steering angles using YOLOv5, enhancing vehicle navigation precision (Ye et al. 2022). Mostafa et al. (2022) compared the effectiveness of YOLOv5, YOLOX, and Faster R-CNN in detecting occluded objects for autonomous vehicles, improving detection reliability (Mostafa et al. 2022). Jia et al. (2023) proposed an enhanced YOLOv5 detector for autonomous driving, which offers increased speed and accuracy (Jia et al. 2023). Chen et al. (2023) utilized an improved YOLOv5-OBB algorithm for autonomous parking space detection in electric vehicles, enhancing operational efficiency (Chen et al. 2023). Liu and Yan (2022) customized YOLOv7 for vehicle-related distance estimation, providing essential metrics for safe navigation (Liu and Yan 2022). Mehla et al. (2023) evaluated YOLOv8 against EfficientDet in autonomous maritime vehicles, highlighting the superior detection capabilities of YOLOv8 (Mehla et al. 2023).

Further advancements with YOLOv8 have led to significant improvements in object detection in adverse weather conditions, an area of particular concern for autonomous driving. Applying transfer learning techniques using datasets from diverse weather conditions has markedly increased the detection performance of YOLOv8, ensuring reliable recognition of crucial road elements like pedestrians and obstacles under challenging weather scenarios (Kumar and Muhammad 2023). Additionally, the development of YOLOv8 for specific tasks such as brake light status detection illustrates the algorithm's flexibility and its potential in enhancing interpretability and safety for autonomous vehicles (Oh and Lim 2023). These innovations underscore the critical role of YOLOv8 and YOLOv9 in pushing the boundaries of what is possible in the autonomous vehicle industry, highlighting their impact in meeting the rigorous demands for safety and reliability in self-driving technolo-gies (Afdhal et al. 2023). YOLOv8 and YOLOv9 are at the forefront of transforming the landscape of autonomous vehicle technologies, playing a pivotal role in enhancing the operational safety and efficiency of self-driving cars. These models have excelled in real-time object detection, a crucial aspect of autonomous driving, especially under the challenging and variable conditions typical in real-world traffic environments. An enhanced version, addresses the need for detecting smaller objects such as traffic signs and signals, demonstrating its utility with a notable accuracy rate and efficiency in processing, making it ideal for high-speed driving scenarios (Wang et al. 2024). Table 2 illustrates different applications of YOLO in the autonomous vehicle industry, presented in reverse chronological order from the most recent versions to the older ones.

#### 5.1.1 Pedestrian pose estimation

Ali et al. (2023) presented a Bayesian Generalized Extreme Value Model to evaluate real-time pedestrian crash risks at signalized intersections, leveraging advanced AI-based video analytics. This framework employs deep learning algorithms like YOLO for precise object detection and DeepSORT for effective tracking. The model concentrates on crucial safety indicators such as Post Encroachment Time (PET). Through this approach, the study underscores the significant role of AI-driven video analysis in boosting intersection safety by delivering real-time risk assessments. This development signifies a substantial advancement in the proactive management of traffic safety. Hussain et al. (2024) explored the enhancement of pedestrian crash estimation using machine learning techniques focused on anomaly detection. Their study addresses the limitations of traditional Extreme Value Theory (EVT) models by applying unconventional sampling methods, thereby increasing the accuracy and reducing uncertainty in crash risk estimations. The use of YOLO for object detection and DeepSORT for tracking is pivotal in this methodology, enhancing detection accuracy and tracking reliability in real-time scenarios. Ghaziamin et al. (2024) developed a privacy-preserving real-time passenger counting system for bus stops using overhead fisheye cameras. This innovative system employs YOLOv4, Detecnet-V2 and Faster-RCNN for detection purposes and DeepSORT for tracking. The system processes data in real-time at 30 frames per second (FPS) when utilizing YOLOv4 as the detection model. This technology significantly enhances transit planning by providing accurate passenger counts, while also maintaining passenger privacy and energy efficiency.

Additionally, Pedestrian-vehicle conflict prediction was explored by Zhang et al. (2020) proposed a model employing a Long Short-Term Memory (LSTM) neural network to forecast pedestrian-vehicle conflicts at signalized intersections by analyzing video data. The model uses YOLOv3 for object detection and Deep SORT for tracking, achieving impressive accuracy rates and demonstrating the transformative potential of LSTM networks in collision warning systems. This approach suggests a proactive enhancement of pedestrian safety in connected vehicle environments.

Crossing intention prediction and behavioral analysis was explored by Zhang et al. (2020) utilized an LSTM neural network to predict pedestrian red-light crossing intentions at intersections by analyzing video data of real traffic scenarios. The model uses YOLOv3 for detection and DeepSORT for tracking, recognizing patterns that indicate potential red-light crossings with a high accuracy rate. This capability aims to improve traffic safety through vehicle-to-infrastructure communication systems that alert drivers to potential pedestrian violations, thereby preventing accidents. Yang et al. (2022) introduced the VENUS smart node, a cooperative traffic signal assistance system for non-motorized users and individuals with disabilities. This novel infrastructure leverages computer vision and edge AI to integrate real-time data on pedestrian movement and intent. The system employs YOLOv4 for detection and OpenPose for pose estimation, achieving high accuracy in detecting crossing intentions and mobility status across various test sites. This innovation has significant potential for widespread use in smart city infrastructures, greatly enhancing safety and accessibility.

**Table 2** Studies on YOLO applications focus on object detection and real-time performance improvements to enhance autonomous vehicles and traffic safety

| Title of paper | Description of work | Purpose and YOLO usage | Version | Refs. and Year |
|---|---|---|---|---|
| "Multi-Class Vehicle Detection and Classification with YOLO11 on UAV-Captured Aerial Imagery" | Utilizes YOLO11 for real-time vehicle detection and classification in UAV-captured traffic images, with a comparative analysis against YOLOv10 | Focuses on robust vehicle detection in dynamic UAV-captured images, enhancing real-time processing and accuracy in aerial traffic monitoring | YOLO11 | Bakirci et al. (2024), 2024 |
| "YOLOv10-Based Real-Time Pedestrian Detection for Autonomous Vehicles" | Presents a real-time pedestrian detection method enhancing YOLOv10 with EfficientNet backbone, C2F-DM, BiFormer, and multi-scale feature fusion detection head for complex environments | Aims to enhance autonomous vehicle safety by providing efficient multi-scale pedestrian detection | YOLOv10 | Li et al. (2024), 2024 |
| "Improved YOLOv10 for Visually Impaired: Balancing Model Accuracy and Efficiency in the Case of Public Transportation" | Introduces Improved-YOLOv10 with Coordinate Attention and Adaptive Kernel Convolution for enhanced bus detection and POV classification for the visually impaired | Enhances accessibility in public transportation for visually impaired individuals through improved detection and efficiency | YOLOv10 | Arifando et al. (2025), 2025 |
| "Transforming Aircraft Detection Through LEO Satellite Imagery and YOLOv9 for Improved Aviation Safety" | Utilizes YOLOv9 with LEO satellite imagery for enhanced aircraft detection in wide-area airport environments | Aims to improve airport security and aviation safety by integrating advanced YOLO-based object detection with satellite imagery | YOLOv9 | Bakirci and Bayraktar (2024), 2024 |
| "YOLOv8-QSD: An Improved Small Object Detection Algorithm for Autonomous Vehicles Based on YOLOv8" | Developed an anchor-free, BiFPN-enhanced YOLOv8 model for better small object detection in driving scenarios | Enhances detection of small objects for autonomous vehicles with reduced computational demands, tested on SODA-A dataset | YOLOv8-QSD | Wang et al. (2024), 2024 |
| "Object Detection in Dense and Mixed Traffic for Autonomous Vehicles With Modified YOLO" | Adapted YOLOv7 with deformable layers and softNMS for object detection in heavy Indonesian traffic | Enhances detection and classification of objects around autonomous vehicles using a modified YOLOv7, tested on a novel Indonesian traffic dataset | YOLOv7-MOD | Wibowo et al. (2023), 2023 |
| "Object Tracking for Autonomous Vehicle Using YOLOV3" | Evaluated YOLOv3 for object tracking in autonomous vehicles | Two models were provided, one trained using only the online COCO dataset and the other with additional images from various locations at Universiti Malaysia Pahang (UMP). | YOLOv3 | Hung et al. (2022), 2022 |
| "The improvement in obstacle detection in autonomous vehicles using YOLO non-maximum suppression fuzzy algorithm" | Employed a hybrid of fuzzy logic and NMS in YOLO for better obstacle detection in autonomous driving | Enhances obstacle detection accuracy and speed using a modified YOLO algorithm | YOLOv3 | Zaghari et al. (2021), 2021 |

Jiao and Fei (2023) conducted a study on monitoring pedestrian walking speeds at the street level using drones. The research utilized UAV-based video footage to measure the walking speeds of pedestrians on a commercial street. Deep learning algorithms, particularly YOLOv5 for object detection and DeepSORT for tracking, were employed in this study. Speed calculations were adjusted for geometric distortions using the SIFT and RANSAC algorithms, achieving high accuracy. The study found that 90.5% of the corrected speeds had an absolute error of less than 0.1 m/s, providing a precise and non-intrusive method for analyzing pedestrian walking speeds. Wang et al. (2024) used drone-captured video footage to examine "safe spaces" for pedestrians and e-bicyclists at urban crosswalks. The study discovered that e-bicyclists maintain larger semi-elliptical safe zones that are sensitive to speed changes compared to the semi-circular zones maintained by pedestrians. By quantifying these safe spaces and examining variations due to speed and traffic presence, the study offers valuable insights for enhancing crosswalk safety and managing urban traffic more effectively. The use of YOLOv3 for object detection and DeepSORT for tracking plays a critical role in this analysis. Zhou et al. (2023) developed an innovative model that integrates a pedestrian-centric environment graph with Graph Convolutional Networks (GCNs) and a pedestrian-state encoder. This model effectively captures dynamic interactions between pedestrians and their environments, providing advanced safety warnings by predicting crossing intentions up to three seconds in advance. This model holds significant potential for applications in intelligent transportation systems. The integration of YOLOv5 for detection, DeepSORT for tracking, and HRNet for pose estimation enhances the model's predictive accuracy and real-time application. Table 3 illustrates different applications of YOLO usage in pedestrian pose estimation, for intention prediction and behavioral analysis.

#### 5.1.2 Traffic sign detection

Traffic sign detection and recognition systems play a pivotal role in enhancing road safety and are essential for the advancement of autonomous driving. These systems enable drivers, or autonomous vehicles, to effectively respond to road conditions, ensuring the safety of all road users (Flores-Calero et al. 2024). However, the complexity and variability of traffic environments, such as adverse weather conditions, combined with the small size of traffic signs present significant challenges for accurately detecting small traffic signs in real-world scenarios (Li et al. 2023; Mahaur and Mishra 2023; Zhang et al. 2020)

For instance, Li et al. (2022) presented a classical YOLO-based architecture for traffic sign recognition. First, traffic signs are categorized and preprocessed according to their specific characteristics. The processed images are then input into an optimized convolutional neural network for finer category classification. The proposed recognition algorithm was tested using a dataset based on the German traffic sign recognition standard, and its per-

**Table 3** Studies on YOLO usage in pedestrian pose estimation, for intention prediction and behavioral analysis

| Title of paper | Description of work | Purpose and YOLO usage | Version | Refs. and Year |
|---|---|---|---|---|
| "Multi-Object Pedestrian Tracking Using Improved YOLOv8 and OC-SORT" | Proposes a comprehensive approach for pedestrian tracking by combining the improved YOLOv8 object detection algorithm with the OC-SORT tracking algorithm, integrating advanced techniques such as SoftNMS, GhostConv, and C3Ghost Modules | Aimed to enhance multi-object pedestrian tracking for autonomous driving systems by improving detection accuracy and model efficiency with YOLOv8, and integrating it with the OC-SORT tracking algorithm for robust tracking in challenging scenarios | YOLOv8 | Xiao and Feng (2023), 2023 |
| "Revisiting The Hybrid Approach of Anomaly Detection and Extreme Value Theory for Estimating Pedestrian Crashes Using Traffic Conflicts Obtained from AI-Based Video Analytics" | Focuses on improving pedestrian crash predictions by utilizing machine learning for anomaly detection and integrating it with Extreme Value Theory (EVT). The approach leverages YOLOv7 for extracting traffic conflicts and relevant data, providing a robust framework for predicting pedestrian crashes from traffic incidents | YOLOv7 is used for the extraction of traffic conflicts and relevant data needed for the EVT model | YOLOv7 | Hussain et al. (2024), 2024 |
| "Pedestrian Walking Speed Monitoring at Street Scale by An In-Flight Drone" | This study introduces a method for measuring pedestrian walking speeds on a commercial street using drone video. Pedestrians are detected and tracked using YOLOv5 and the DeepSORT algorithm, with distance calculations performed using Scale-Invariant Feature Transform (SIFT) and random sample consensus (RANSAC) algorithms, followed by geometric correction | Provide an accurate and cost-effective method for monitoring pedestrian walking speeds over large areas. YOLOv5 is utilized for pedestrian detection in drone footage, which is crucial for tracking and speed calculation | YOLOv5 | Jiao and Fei (2023), 2023 |
| "Pedestrian Crossing Intention Prediction From Surveillance Videos For Over-The-Horizon Safety Warning" | Prediction of pedestrian crossing intentions using surveillance camera footage. The framework constructs a pedestrian-centric environment graph, uses a Graph Convolutional Network (GCN) for environment encoding, and employs a pedestrian-state encoder for extracting behaviour features. An intention prediction decoder is then used to determine crossing probabilities | YOLOv5 is used to identify and track pedestrians within the surveillance footage, facilitating the extraction of visual and behavioural features necessary for the prediction framework | YOLOv5 | Zhou et al. (2023), 2024 |
| "A Bayesian Generalised Extreme Value Model to Estimate Real-Time Pedestrian Crash Risks at Signalised Intersections Using Artificial Intelligence-Based Video Analytics" | Develop a Bayesian Generalised Extreme Value (GEV) model for estimating real-time pedestrian crash risks at signalized intersections. The model uses AI-based video analytics to identify and track vehicles and pedestrians, extracting traffic conflicts and relevant covariates for risk assessment | Identify crash-prone conditions and implement timely risk mitigation strategies to enhance pedestrian safety at intersections. YOLOv4 is used to detect and track pedestrians and vehicles, facilitating the extraction of necessary data for the Bayesian EVT model | YOLOv4 | Ali et al. (2023), 2022 |

**Table 3** (continued)

| Title of paper | Description of work | Purpose and YOLO usage | Version | Refs. and Year |
|---|---|---|---|---|
| "A Privacy-Preserving Edge Computing Solution For Real-Time Passenger Counting At Bus Stops Using Overhead Fisheye Camera" | Development of a privacy-preserving edge computing solution for real-time passenger counting at bus stops using fisheye camera footage. The system leverages YOLOv4 for passenger detection and is designed to operate efficiently on edge devices powered by solar panels | YOLOv4 creates an automated, efficient, and privacy-respecting passenger counting system suitable for smart city bus stops. Model is identified as the superior object detection model in this study, outperforming DetectNet and Faster-RCNN | YOLOv4 | Ghaziamin et al. (2024), 2024 |
| "Cooperative Traffic Signal Assistance System for Non-Motorized Users and Disabilities Empowered by Computer Vision and Edge Artificial Intelligence" | Create the VENUS smart node, a system designed to aid non-motorized and disabled users at intersections. The system integrates computer vision and edge AI technologies for object recognition, user localization, and pose direction estimation, ensuring accurate detection and tracking of various users | Enhancing traffic signal assistance for non-motorized and disabled users by providing real-time data and interaction capabilities. YOLOv4 is used for object recognition, user localization, and pose direction estimation of non-motorized users | YOLOv4 | Yang et al. (2022), 2022 |
| "Examining Safe Spaces for Pedestrians and E-Bicyclists at Urban Crosswalks: An Analysis Based on Drone-Captured Video" | Assess the safety zones for pedestrians and e-bicyclists at urban crosswalks using drone footage. YOLOv3 is employed for accurate identification and tracking of individuals, allowing for detailed analysis of their movements and interactions | Enhancing the safety of pedestrians and e-bicyclists at crosswalks by analyzing their safe spaces. YOLOv3 is used for the study of user's dynamic interactions | YOLOv3 | Wang et al. (2024), 2024 |
| "Forecast Pedestrian-Vehicle Collisions at Traffic Lights" | Implements YOLOv3 for detecting pedestrian-vehicle interactions and classifying them into safe interactions, slight conflicts, and severe conflicts | Enhance pedestrian safety at intersections by modelling and predicting potential pedestrian-vehicle conflicts | YOLOv3 | Zhang et al. (2020), 2020 |
| "Prediction of Pedestrian Crossing Intentions at Intersections Based On Long Short-Term Memory Recurrent Neural Network" | Prediction pedestrian red-light crossing behaviour at intersections using LSTM networks. YOLOv3 is employed to detect pedestrians and extract relevant characteristics from video data, which are then used for behavioural prediction | YOLOv3 is used to identify pedestrians and extract relevant characteristics from the video data, which are then passed into the LSTM neural network for prediction | YOLOv3 | Zhang et al. (2020), 2020 |

formance was compared with other baseline algorithms. Results show that the algorithm significantly improves processing speed while maintaining high classification accuracy, making it better suited for traffic sign recognition systems.

Zhang et al. (2017) presented a Chinese traffic sign detection algorithm based on a deep convolutional network. To enable real-time detection, they proposed an end-to-end convolutional network inspired by YOLOv2. Considering the characteristics of traffic signs, they incorporated multiple $1\times1$ convolutional layers in the intermediate network layers while reducing the number of convolutional layers in the top layers to decrease computational complexity. For effective small traffic sign detection, the input images are divided into dense grids to capture finer feature maps. Additionally, they expanded the Chinese Traffic Sign Dataset (CTSD) and enhanced the marker information available online. Experimental results using both the expanded CTSD and the German Traffic Sign Detection Benchmark (GTSDB) demonstrate that the proposed method is faster and more robust.

On the other hand, Zhang et al. (2020) proposed a new detection scheme, MSA_YOLOv3, for accurate real-time localization and classification of small traffic signs. The approach begins with data augmentation using image mixup technology. A multi-scale spatial pyramid pooling block is incorporated into the Darknet53 network, enabling more comprehensive learning of object features. Additionally, a bottom-up augmented path is designed to enhance the feature pyramid in YOLOv3, allowing effective utilization of fine-grained features in the lower layers for precise object localization. Tests on the TT100K dataset show that MSA_YOLOv3 outperforms YOLOv3 in detecting small traffic signs.

Recently, later versions of YOLO are being applied to detect traffic signs. For example, Mahaur and Mishra (2023) presented a new version called iS-YOLOv5 model, which increases the mean Average Precision (mAP) by 3.35% on the BDD100K dataset to detect traffic sign and traffic lights. While, Bai et al. (2023) introduced two innovative traffic sign detection models, called YOLOv5-DH and YOLOv5-TDHSA, based on the YOLOv5s model with the following modifications (YOLOv5-DH uses only the second modification): (1) replacing the last layer of the 'Conv + Batch Normalization + SiLU' (CBS) structure in the YOLOv5s backbone with a transformer self-attention module (T in the YOLOv5-TDHSA's name), and also adding a similar module to the last layer of its neck, so that the image information can be used more comprehensively, (2) replacing the YOLOv5s coupled head with a decoupled head (DH in both models' names) to increase the detection accuracy and speed up the convergence, and (3) adding a small-object detection layer (S in the YOLOv5-TDHSA's name) and an adaptive anchor (A in the YOLOv5-TDHSA's name) to the YOLOv5s neck to improve the detection of small objects. Their experiments were conducted using the TT100K dataset.

Similarly, Li et al. (2023) proposed a small object detection algorithm for traffic signs based on the improved YOLOv7 called SANO-YOLOv7. First, the small target detection layer in the neck region was added to augment the detection capability for small traffic sign targets. Simultaneously, the integration of self-attention and convolutional mix modules (ACmix) was applied to the newly added small target detection layer, enabling the capture of additional feature information through the convolutional and self-attention channels within ACmix. Furthermore, the feature extraction capability of the convolution modules was enhanced by replacing the regular convolution modules in the neck layer with omni-dimensional dynamic convolution (ODConv). To further enhance the accuracy of small target detection, the normalized Gaussian Wasserstein distance (NWD) metric was introduced to mitigate the sensitivity to minor positional deviations of small objects. SANO-YOLOv7 was trained on the highly challenging TT100K public dataset.

More interesting traffic sign detection studies were presented with YOLOv8 and later YOLO versions. In the Robotaxi-Full Scale Autonomous Vehicle Competition, YOLOv8 was specifically adapted to recognize and interpret traffic signs, providing real-time alerts that are essential for safe driving (Soylu and Soylu 2024). Next, Zhang (2024) introduced an enhanced traffic sign detection algorithm based on YOLOv9. AKConv replaces the Conv module in RepNCSPELAN4, maintaining detection accuracy while reducing weight. Focal-EIoU Loss replaces the original regression loss function, Clou Loss, accelerating convergence and improving accuracy by dividing the aspect ratio's loss term into the difference between the minimum outer frame's width and height and the predicted width and height. Additionally, the network's feature extraction capability and detection accuracy are further strengthened by incorporating the Convolutional Block Attention Module (CBAM) attention mechanism. The public TT100K traffic sign dataset was used for training and evaluation.

Up until now, YOLO versions 10, 11 and 12 have not been applied in detecting traffic signs. This presents an exciting opportunity for future research and application, as these versions may offer improvements in accuracy, speed, and adaptability for detecting traffic signs in complex environments (4).

### 5.2 Healthcare and medical imaging

YOLO has marked a significant technological advancement in healthcare applications as well, especially with the introduction of newer versions such as YOLOv7 and YOLOv8 (Pandey et al. 2023; Ju and Cai 2023; Inui et al. 2023). The recent iterations of YOLO, particularly YOLOv7, YOLOv8, and YOLOv9, could significantly enhance medical diagnostics by offering advanced computational efficiency and improved feature extraction capabilities, making them suitable for real-time medical imaging applications. Such capabilities are crucial in urgent care scenarios, where swift diagnosis can be pivotal. For instance, YOLOv8's sophisticated algorithms excel in accurately delineating complex biological structures, vital for identifying pathologies in conditions like vascular diseases or tumors. Similarly, YOLOv9's rapid processing power enables immediate analysis of medical images, essential in emergency medical responses where timely intervention is critical. These versions have the potential to revolutionize healthcare by facilitating early detection of diseases and supporting continuous patient monitoring, transforming the traditional approach of healthcare diagnostics into one that integrates accurate, swift diagnostics seamlessly with routine medical examinations. Unlike the traditional methods which depend heavily on manual annotation and are prone to errors and subjectivity, YOLO algorithms automate the detection and localization of medical anomalies such as tumors, lesions, and other pathological markers across various imaging modalities. This automation is driven by YOLO's unique architecture that efficiently predicts multiple bounding boxes and class probabilities in a single analysis, enhancing diagnostic accuracy and reducing the potential for human error.

In the field of medical imaging and diagnostics, the adoption of the YOLO object detection algorithm has showcased promising improvements in accuracy and efficiency, particularly with its latest versions like YOLOv5, YOLOv6, YOLOv7, and YOLOv8. For instance,

**Table 4** Studies on YOLO usage in traffic sign detection and recognition

| Title of paper | Description of work | Purpose and YOLO usage | Version | Refs. and Year |
|---|---|---|---|---|
| "FINE-GRAINED CLASSIFICATION OF MILITARY AIRCRAFT USING PRE-TRAINED DEEP LEARNING MODELS AND YOLO11" | Examines the use of pre-trained CNN models and YOLO11 for fine-grained classification of military aircraft, using a dataset of 24,164 images for training and 6,042 for testing | Aims to enhance military aviation applications by achieving accurate aircraft identification and tail number extraction using advanced AI-powered image recognition | YOLO11x-cls | Karaca and Atasoy (2025), 2025 |
| "Research on traffic sign detection based on improved YOLOv9" | An enhanced traffic sign detection algorithm based on YOLOv9 was introduced to address issues of low accuracy, missed detection, and the omission of small targets in complex environments. The TT100K traffic sign dataset was used for training and evaluation | YOLOv9 was modified by replacing the Conv module with AKConv, reducing weight while maintaining accuracy. Focal-EIoU Loss improves convergence and accuracy, and the CBAM attention mechanism enhances feature extraction and detection performance | YOLOv9 | Zhang (2024), 2024 |
| "A performance comparison of YOLOv8 models for traffic sign detection in the Robotaxi-full scale autonomous vehicle competition" | To develop a traffic sign recognition system using YOLOv8 for real-time detection and classification of objects, and to compare the performance of YOLOv8 models for traffic sign detection within the Robotaxi-full framework | YOLOv8 was trained on a dataset of traffic sign images to develop a model capable of accurately recognizing and classifying various types of traffic signs | YOLOv8 | Soylu and Soylu (2024), 2024 |
| "A Small Object Detection Algorithm for Traffic Signs Based on Improved YOLOv7" | A traffic sign detection algorithm for small-sized signs was proposed, built upon the enhanced YOLOv7 model, improving its ability to accurately identify and classify small traffic signs in challenging environments. Their effectiveness was validated on a public dataset | YOLOv7 was enhanced with a small target detection layer incorporating ACmix modules for improved feature extraction via self-attention and convolution. ODConv replaced standard convolution in the neck layer, boosting feature extraction. Additionally, the NWD metric was introduced to reduce sensitivity to positional deviations, improving small traffic sign detection accuracy | YOLOv7 | Li et al. (2023), 2023 |
| "Local Regression Based Real-Time Traffic Sign Detection using YOLOv6" | The capabilities of YOLOv6 are combined with an optimized Logistic Regression (LR)-based classifier, delivering improved accuracy and performance tailored to resource-constrained environments | YOLOv6 is well-suited for applications on embedded systems and smartphones, where power efficiency is a critical constraint | YOLOv6 | Kaur and Singh (2022), 2022 |

**Table 4** (continued)

| Title of paper | Description of work | Purpose and YOLO usage | Version | Refs. and Year |
|---|---|---|---|---|
| "Two Novel Models for Traffic Sign Detection Based on YOLOv5s" | Based on the YOLOv5s architecture, two novel traffic sign detection models, YOLOv5-DH and YOLOv5-TDHSA, have been proposed. Their effectiveness was validated on a public dataset | Proposed YOLOv5-DH and YOLOv5-TDHSA, enhancing YOLOv5s with transformer self-attention modules, decoupled heads, small-object detection layers, and adaptive anchors. YOLOv5-DH focuses on accuracy and convergence, while YOLOv5-TDHSA improves small object detection | YOLOv5 | Bai et al. (2023), 2023 |
| "Small-object detection based on YOLOv5 in autonomous driving systems" | Investigated and refined YOLOv5 for improved detection of small objects such as traffic signs and traffic lights, tested on BDD100K, TT100K, and DTLD datasets | Introduced architectural changes to the popular YOLOv5 model to improve its performance in the detection of small objects without sacrificing the detection accuracy of large objects | YOLOv5 | Mahaur and Mishra (2023), 2023 |
| "Deep convolutional neural network for enhancing traffic sign recognition developed on YOLO V4" | Analyzed YOLO V4 and YOLO V4-tiny with SPP for better feature extraction in traffic sign recognition | Compared improving traffic sign recognition performance by integrating SPP into YOLO V4 backbones | YOLOv4 | Dewi et al. (2022), 2022 |
| "Real-Time Detection Method for Small Traffic Signs Based on YOLOv3" | YOLOv3 was enhanced for real-time localization and classification of small traffic signs, effectively utilizing fine-grained features in the lower layers for accurate object detection. Experiments were conducted using the TT100K dataset | MSA_YOLOv3, was proposed for real-time localization and classification of small traffic signs. It incorporates image mixup for data augmentation, integrates a multi-scale spatial pyramid pooling block into Darknet53 for comprehensive feature learning, and enhances YOLOv3's feature pyramid with a bottom-up augmented path for accurate object localization | YOLOv3 | Zhang et al. (2020), 2020 |
| "A real-time Chinese traffic sign detection algorithm based on modified YOLOv2" | A Chinese traffic sign detection algorithm based on YOLOv2 was presented. To effectively detect small traffic signs, input images are divided into dense grids to capture finer feature maps. Experiments were conducted using the Chinese Traffic Sign Dataset (CTSD) | YOLOv2 was modified by adding multiple $1\times1$ convolutional layers in the intermediate layers and reducing the convolutional layers in the top layers to reduce computational complexity. To enhance small traffic sign detection, input images are divided into dense grids, capturing finer feature maps | YOLOv2 | Zhang et al. (2017), 2017 |

**Table 4** (continued)

| Title of paper | Description of work | Purpose and YOLO usage | Version | Refs. and Year |
|---|---|---|---|---|
| "YOLO-Based Traffic Sign Recognition Algorithm" | A modified YOLO architecture for traffic sign recognition, which categorizes and preprocesses traffic signs before inputting them into an optimized CNN for finer classification. Tested on a German traffic sign dataset, the algorithm improves speed and accuracy, outperforming baseline models for traffic sign recognition systems | A classical YOLO architecture was modified by enhancing the pooling layer, which reduces the spatial resolution of the input and eliminates redundant information, improving efficiency and focus on relevant features | YOLO | Li et al. (2022), 2022 |

Luo et al. (2021) leveraged YOLOv5 in conjunction with ResNet50 to enhance chest abnormality detection, demonstrating the algorithm's proficiency in identifying subtle medical conditions (Luo et al. 2021). Similarly, Wu et al. (2022) developed Me-YOLO, an adapted version of YOLOv5, to improve the detection of medical personal protective equipment, highlighting the model's adaptability to varied medical use cases (Wu et al. 2022). Moreover, advancements like the CSFF-YOLOv5 by Zhao et al. (2024) introduced modifications for better feature fusion, significantly boosting the detection accuracy in femoral neck fracture cases (Zhao et al. 2024). This specificity is further explored by Goel and Patel (2024), who enhanced YOLOv6 for lung cancer detection using an advanced PSO optimizer, underscoring the potential of YOLO algorithms in facilitating early disease diagnosis and treatment (Goel and Patel 2024). Additionally, the extension of YOLOv6 by Norkobil Saydirasulovich et al. (2023) for improved fire detection in smart city environments exemplifies the algorithm's versatility beyond traditional medical applications, proving its efficacy in diverse environmental conditions (Norkobil Saydirasulovich et al. 2023). Each of these developments not only enhances specific medical diagnostic processes but also paves the way for integrating these advanced object detection systems into broader healthcare applications, as illustrated by the innovative uses of YOLOv7 and YOLOv8 in detecting whole body bone fractures and enhancing hospital efficiency (Zou and Arshad 2024; Salinas-Medina and Neme 2023). These studies collectively demonstrate the significant advancements brought by YOLO in the healthcare sector, ensuring more precise, efficient, and versatile diagnostic solutions.

Recent versions such as YOLOv7, YOLOv8 and YOLOv9 have been effectively demonstrated across a variety of healthcare applications. Razaghi et al. (2024) utilized YOLOv8 for the innovative diagnosis of dental diseases, highlighting its precision in identifying dental pathologies (Razaghi et al. 2024). Similarly, Pham and Le (2024) leveraged YOLOv8 for the detection and classification of ovarian tumors from ultrasound images, showcasing the model's adaptability to different medical imaging modalities (Pham and Le 2024). Krishnamurthy et al. (2023) applied custom YOLO architectures to enhance object detection capabilities during endoscopic surgeries, illustrating the potential of YOLO in surgical settings (Krishnamurthy et al. 2023). Furthermore, Palanivel et al. (2023) discussed the application of YOLOv8 in cancer diagnosis through medical imaging, further cementing YOLO's role in critical healthcare applications (Palanivel et al. 2023).

Continuing with advancements, Karaköse et al. (2024) introduced CSFF-YOLOv5, an improved YOLO model for femoral neck fracture detection, utilizing advanced feature fusion techniques (Karaköse et al. 2024). Inui et al. (2023) demonstrated YOLOv8's effectiveness in detecting elbow osteochondritis dissecans in ultrasound images, which supports its use in orthopedic diagnostics (Inui et al. 2023). Bhojane et al. (2023) employed YOLOv8 for detecting liver lesions from MRI and CT images, underscoring the algorithm's capability across various imaging technologies (Bhojane et al. 2023). Additionally, Zhang et al. (2023) developed an improved detection model for microaneurysms using YOLOv8, which illustrates continuous enhancements in YOLO's application to highly specific medical tasks (Zhang et al. 2023).

Table 5 illustrates the different uses of YOLO versions in security and survelliance:

### 5.3 Security and surveillance

In the ever-evolving field of security systems, YOLO's application extends to detecting unauthorized entries and identifying potential threats swiftly, thereby bolstering security measures (Majeed et al. 2022; Aboah et al. 2021). Recent YOLO models such as YOLOv6 enhanced security and surveillance applications by improving detection accuracy through deeper network layers that process images with greater precision (AFFES et al. 2023). Meanwhile, YOLOv7 offers advanced customization options that allow security systems to be finely tuned to specific surveillance needs, enhancing the adaptability and effectiveness of threat detection (AFFES et al. 2023; Cao and Ma 2023). These YOLO versions support high-resolution video feeds, ensuring that security personnel can engage with real-time data to make informed decisions quickly. Further advancements in surveillance systems are embodied by YOLOv8 and YOLOv9, which introduce significant innovations in deep learning for security applications (Chatterjee et al. 2024; Sandhya and Kashyap 2024; Tran et al. 2024). YOLOv8's architecture is designed to handle complex environments where traditional surveillance systems may fail, such as varying lighting and weather conditions. This version's robust performance in diverse scenarios enhances its utility in comprehensive security strategies. On the other hand, YOLOv9 pushes the boundaries of speed and accuracy, providing unparalleled real-time analysis and detection capabilities. Its deployment in surveillance systems ensures that even the subtlest anomalies are detected, reducing the likelihood of security breaches. The integration of recent versions of YOLO such as YOLOv8 and YOLOv9 into security frameworks not only streamlines operations but also ensures a proactive approach to threat management, keeping public and private spaces safer across the globe (Bakirci and Bayraktar 2024a, b; Shoman et al. 2024c).

The application of YOLO models in surveillance and security systems highlights their pivotal role in enhancing real-time response and precision. Majeed et al. (2022) investigated the effectiveness of a YOLOv5-based security system within a real-time environment, underscoring its capability to significantly improve operational efficiency in dynamic settings. Similarly, AFFES et al. (2023) conducted a comparative study across YOLOv5, YOLOv6, YOLOv7, and YOLOv8, focusing on their performance in intelligent video surveillance systems. Their analysis demonstrated the incremental improvements in detection accuracy and processing speed, crucial for real-time security applications. Further advancing the field, Cao and Ma (2023) utilized a refined YOLOv7 model to enhance campus security through improved target detection capabilities, highlighting the model's precision in identifying potential threats in densely populated environments. Chatterjee et al. (2024) introduced a YOLOv8-based intrusion detection system specifically tailored for physical security and surveillance, which significantly contributes to safeguarding assets and individuals by detecting unauthorized entries or activities effectively. Additionally, Sandhya and Kashyap (2024) employed YOLOv8 for real-time object-removal tampering localization in surveillance videos, a crucial technology for maintaining the integrity of video evidence and ensuring the reliability of surveillance feeds. Together, these studies showcase the robustness of YOLO architectures in addressing diverse and complex security challenges, providing substantial improvements in both the efficacy and efficiency of surveillance operations.

Recent studies have significantly leveraged advanced YOLO models to enhance surveillance and security across various domains. Bakirci and Bayraktar (2024a) discussed optimizing ground surveillance for aircraft monitoring using YOLOv9, highlighting its efficacy

**Table 5** Studies on YOLO applications in healthcare and medicine, emphasizing object detection for diagnostic imaging and real-time medical analysis

| Title of paper | Description of work | Purpose and YOLO usage | Version | Refs. and Year |
|---|---|---|---|---|
| "Development of an AI-Supported Clinical Tool for Assessing Mandibular Third Molar Tooth Extraction Difficulty Using Panoramic Radiographs and YOLO11 Sub-Models" | Develops an AI tool using YOLO11 sub-models to evaluate mandibular third molar extraction difficulty from panoramic radiographs, incorporating different YOLO11 sizes for scenario-based evaluations | Enhances dental practice by providing an AI tool for accurate and reliable assessment of tooth extraction difficulty, facilitating improved decision-making and patient management | YOLO11 | Akdoğan et al. (2025), 2025 |
| "A Machine Learning-Based Model for the Detection of Skin Cancer Using YOLOv10" | Presents a machine learning model for skin cancer detection using YOLOv10, involving preprocessing, augmentation, and training on two datasets for different skin conditions | Aims to improve early detection and survival rates for skin cancer through advanced YOLO-based image detection | YOLOv10 | Ali et al. (2024), 2024 |
| "Efficient Skin Lesion Detection using YOLOv9 Network" | Utilized YOLOv9 for advanced skin lesion detection, leveraging deep learning to enhance diagnostic accuracy and speed | Developed improved skin lesion identification using YOLOv9, showcasing significant advances in detection performance | YOLOv9 | Ju and Cai (2023), 2023 |
| "Fracture detection in pediatric wrist trauma X-ray images using YOLOv8 algorithm" | Employed YOLOv8 with data augmentation on the GRAZPEDWRI-DX dataset for detecting fractures in pediatric wrist X-ray images | Enhanced fracture detection in pediatric wrist trauma using YOLOv8, achieving superior mAP compared to previous versions. Designed an app for surgical use | YOLOv8 | Ju and Cai (2023), 2023 |
| "Chapter 4 - Medical image analysis of masses in mammography using deep learning model for early diagnosis of cancer tissues" | Utilizes YOLOv7 to detect and diagnose cancerous tissues in mammogram images, leveraging advancements in deep learning for early cancer detection | Aims to enhance early detection of breast cancer using YOLOv7, improving diagnostic accuracy with deep learning integration. Performance measured by Precision, Recall, and F1-score | YOLOv7 | Julia et al. (2024), 2024 |
| "Improving YOLOv6 using advanced PSO optimizer for weight selection in lung cancer detection and classification" | Enhanced YOLOv6 with Particle Swarm Optimization for weight optimization in lung cancer detection from CT scans | Utilized advanced PSO to optimize YOLOv6 for higher accuracy in detecting lung cancer, significantly outperforming previous methods on the LUNA 16 Dataset | YOLOv6 | Goel and Patel (2024), 2024 |
| "One-Stage methods of computer cision object detection to classify carious lesions from smartphone imaging" | Utilized YOLO v5, YOLO v5X, and YOLO v5M to detect and classify carious lesions from smartphone images | Aimed to automate caries detection with enhanced accuracy using YOLO. mAP, P, and R metrics validated performance | YOLOv5, YOLOv5X, YOLOv5M | Salahin et al. (2023), 2023 |
| "An Improved Method of Polyp Detection Using Custom YOLOv4-Tiny" | Customized YOLOv4-tiny with Inception-ResNet-A block for enhanced detection of polyps in wireless endoscopic images | Developed to improve the detection performance of polyp detection using a modified YOLOv4-tiny. Demonstrated significant performance improvement | YOLOv4-Tiny | Doniyorjon et al. (2022), 2022 |

**Table 5** (continued)

| Title of paper | Description of work | Purpose and YOLO usage | Version | Refs. and Year |
|---|---|---|---|---|
| "Detection of dental caries in oral photographs taken by mobile phones based on the YOLOv3 algorithm" | Utilized YOLOv3 for detecting dental caries from mobile phone images, employing image augmentation and enhancement for improved accuracy | Enhanced detection and diagnosis of dental caries using YOLOv3, with evaluation of diagnostic precision, recall, and F1-score across different datasets | YOLOv3 | Ding et al. (2021), 2021 |
| "Automatic thyroid nodule recognition and diagnosis in ultrasound imaging with the YOLOv2 neural network" | Employed YOLOv2 for automatic detection and diagnosis of thyroid nodules in ultrasound images, enhancing diagnostic precision | Compared AI performance with radiologists using YOLOv2, showing improved accuracy and specificity in thyroid nodule diagnosis. ROC curve analysis confirms effectiveness | YOLOv2 | Wang et al. (2019), 2019 |
| "Real-Time Facial Features Detection from Low Resolution Thermal Images with Deep Classification Models" | Developed a method to localize facial features from low-resolution thermal images by modifying existing deep classification networks for real-time detection | Demonstrates how spatial information can be restored and utilized from classification models for facial feature detection, significantly reducing dataset preparation time while maintaining high precision | Custom Deep Classification Model and YOLO | Kwaśniewska et al. (2018), 2018 |

**Table 6** Studies on YOLO usage in security and surveillance, for real-time threat detection and enhanced monitoring to improved safety measures

| Title of paper | Description of work | Purpose and YOLO usage | Version | Refs. and Year |
|---|---|---|---|---|
| "Assessment of YOLO11 for Ship Detection in SAR Imagery Under Open Ocean and Coastal Challenges" | Evaluated YOLO11's performance for ship detection in challenging SAR imagery, distinguishing between open ocean and coastal scenarios | Aims to enhance maritime surveillance and safety by leveraging the improved detection capabilities of YOLO11 in both open ocean and coastal environments | YOLO11 | Bakirci and Bayraktar (2024), 2024 |
| "Deploying YOLOv10 for Affordable Real-Time Handgun Detection" | Investigated the deployment of YOLOv10 on cost-effective hardware like Raspberry Pi for real-time handgun detection in various scenarios, analyzing detection parameters and model performance | Enhances public safety by enabling affordable, real-time firearm detection technology in everyday environments, using YOLOv10 to adapt AI capabilities for low-cost devices | YOLOv10 | Žigulić et al. (2024), 2024 |
| "YOLOv9-Enabled Vehicle Detection for Urban Security and Forensics Applications" | Implements YOLOv9 for aerial vehicle detection via UAVs, enhancing urban security and forensic capabilities | Focus on utilizing YOLOv9 for real-time vehicle monitoring, facilitating efficient law enforcement and forensic analysis in urban settings | YOLOv9 | Bakirci and Bayraktar (2024b), 2024 |
| "SC-YOLOv8: A Security Check Model for the Inspection of Prohibited Items in X-ray Images" | Developed a custom YOLOv8 model for X-ray image analysis to detect prohibited items. Enhanced model accuracy using a novel backbone structure and data augmentation | Aimed to improve security screening effectiveness and reduce error rates in detecting prohibited items. Showcases an innovative use of YOLOv8 in security applications | YOLOv8 | Han et al. (2023), 2023 |
| "Detection of Prohibited Items Based upon X-ray Images and Improved YOLOv7" | Improved YOLOv7 with spatial attention for contraband detection in X-ray images. Implemented large kernel attention mechanisms to improve texture and feature extraction to boost accuracy | Aims to automate security inspections and enhance public safety by improving prohibited item detection with modified YOLOv7. Demonstrates YOLOv7's adaptability in security systems | YOLOv7 | Yuan et al. (2022), 2022 |
| "Suspicious Activity Trigger System using YOLOv6 Convolutional Neural Network" | Implements YOLOv6 to detect and classify suspicious activities in CCTV footage, enhancing home surveillance systems. Utilizes deep learning to automatically trigger alerts, improving response times and security effectiveness | Aims to reduce property theft by integrating YOLOv6 into home security systems to auto-detect suspicious behavior and alert users. Demonstrates YOLOv6's effectiveness in real-world security applications | YOLOv6 | Awang et al. (2023), 2023 |
| "Real-time Object Detection for Substation Security Early-warning with Deep Neural Network based on YOLO-V5" | Utilizes YOLO-v5 to enhance substation security by detecting multiple threats like fire, unauthorized entry, and vehicle misplacement in real-time. Combines deep learning with video surveillance to reduce the need for extra hardware | Designed to improve substation security management without costly additional equipment by detecting various security threats simultaneously using YOLO-v5. Demonstrates the application of YOLO-v5 in critical infrastructure protection | YOLOv5 | Xiao et al. (2022), 2022 |

**Table 6** (continued)

| Title of paper | Description of work | Purpose and YOLO usage | Version | Refs. and Year |
|---|---|---|---|---|
| "Fighting against terrorism: A real-time CCTV autonomous weapons detection based on improved YOLO v4" | Improved YOLOv4 with SCSP-ResNet backbone and F-PaNet module for detecting weapons in CCTV footage, integrating synthetic and real-world data to enhance detection | Aims to bolster security and counter-terrorism efforts by accurately identifying weapons in CCTV using an advanced YOLOv4 architecture, demonstrating significant performance improvements | YOLOv4 | Wang et al. (2023), 2023 |
| "Automatic tracking of objects using improvised YOLOv3 algorithm and alarm human activities in case of anomalies" | Utilizes an enhanced YOLOv3 model to automatically track objects and alert for anomalies in live video feeds, comparing performance with CNNs and decision trees | Designed to enhance surveillance systems by detecting and alerting on anomalies like bag stealing and lock-breaking, demonstrating rapid processing and high detection accuracy | YOLOv3 | Kashika and Venkatapur (2022), 2022 |
| "Multi-Object Detection using Enhanced YOLOv2 and LuNet Algorithms in Surveillance Videos" | Employs a novel YOLOv2-LuNet combination for efficient multi-object tracking in video surveillance, enhancing feature extraction and object detection accuracy | Designed to improve real-time surveillance by enabling robust multi-object tracking in challenging conditions. Highlights the effectiveness of combined YOLOv2 and LuNet approach | YOLOv2 | Mohandoss and Rangaraj (2024), 2024 |
| "From Silence to Propagation: Understanding the Relationship between 'Stop Snitchin' and 'YOLO'" | Examines the cultural shift from 'Stop Snitchin" to 'YOLO' in urban hip-hop culture, highlighting the role of social media in promoting individualism and exceptionalism | Aims to explore how social media influences criminal behavior and public perception, applying cultural criminology to assess changes in social interactions and deviance | N/A | Smiley (2015), 2015 |

in real-time security applications. Similarly, Chakraborty et al. (2024) explored a multi-model approach for violence detection, incorporating YOLOv8 to improve public safety through automated surveillance. These advancements indicate a shift towards reliable and efficient security systems for complex scenarios.

Chen et al. (2024) delve into the application of an enhanced YOLOv8 model for large-scale security and low-altitude drone-based law enforcement, demonstrating its potential in managing security risks effectively. Further, Pashayev et al. (2023) utilize YOLO8 for intelligent face recognition in smart cameras, contributing to the development of smarter, more responsive surveillance technologies. Additionally, Kaç et al. (2024) investigate image-based security techniques for critical water infrastructure surveillance, employing YOLO models to ensure robust monitoring. Lastly, Gao et al. (2024) introduce an improved YOLOv8s network model for contraband detection in X-ray images, underscoring the versatility and precision of YOLO models in enhancing contraband security measures.

Recent advancements in surveillance technologies have leveraged the YOLO's capabilities, particularly in managing crowd dynamics and detecting critical events. Antony et al. (2024) explored the use of YOLOv8 alongside ByteTrack for crowd management, emphasizing the system's efficiency in improving surveillance and public safety. This integration marks a significant step towards enhancing real-time monitoring capabilities during large public gatherings. Concurrently, Zhang (2024) utilized a YOLO model to detect fire and smoke in IoT surveillance systems, showcasing the model's ability to respond swiftly to emergency situations, thus bolstering safety protocols within environments.

In security, Khin and Htaik (2024) conducted a comparative study of YOLOv8 with other models like RetinaNet and EfficientDet for gun detection, emphasizing YOLOv8's superior accuracy in detecting firearms within a custom dataset. It underlines the critical role of precise object detection to prevent potential threats. Additionally, Nkuzo et al. (2023) provided a comprehensive analysis of the YOLOv7 in detecting car safety belts in real-time, illustrating its importance in enforcing road safety measures. Moreover, Chang et al. (2023) developed an improved YOLOv7, equipped with feature fusion and attention mechanisms, tailored for detecting safety gear violations in high-risk environments like construction, to enhance workplace safety standards. Table 6 presents the various YOLO usage in security and surveillance.

## 5.4 Manufacturing

In the landscape of industrial manufacturing, the deployment of YOLO algorithms can significantly enhance various processes and quality assessment tasks such as the development of automated optical inspection (AOI) systems. Each iteration of the YOLO family, from YOLOv2 to YOLOv5, and beyond into the latest versions like YOLOv6 and YOLOv7, brings forward substantial improvements in detecting defects across various manufacturing domains (Hussain 2023; Ahmad and Rahimi 2022; Pendse et al. 2023; Yi et al. 2024). The high accuracy and real-time processing capabilities of YOLOv6 and YOLOv7, for instance, allow for immediate identification of production flaws, crucial for maintaining workflow efficiency on fast-paced production lines (Wang et al. 2023; Ludwika and Rifai 2024). Advancing into the domain of smart manufacturing, YOLO algorithms are pivotal in revolutionizing quality control mechanisms (Beak et al. 2023; Zhao et al. 2024). The continuous evolution from YOLOv5 to YOLOV6, YOLOv7, YOLOv8, and upto 10th version of YOLO exemplifies the adaptation of deep learning to meet the stringent quality demands of modern manufacturing processes. These algorithms reduce the need for labor-intensive manual inspections, thereby minimizing the margin for human error and enhancing the overall speed of quality assessments (Hussain 2023; Ahmad and Rahimi 2022; Pendse et al. 2023; Yi et al. 2024; Beak et al. 2023; Zhao et al. 2024).

For instance, (Liu and Ye 2023) pioneered YOLO-IMF, an enhanced version of YOLOv8 tailored for precise surface defect detection in industrial settings, exemplifying the algorithm's efficacy in real-time environments. This refinement aims to cater to the high demands for accuracy in manufacturing sectors where defects can significantly impact quality and safety. Continuing this trend, (Wen and Wang 2024) introduced YOLO-SD, which utilizes simulated feature fusion for few-shot learning, enhancing YOLOv8's capability in detecting industrial defects under varied conditions. Similarly, (Karna et al. 2023) extended YOLOv8's utility in monitoring 3D printing processes by optimizing hyperparameters to detect faults more accurately, reflecting a targeted approach to maintaining production integrity. Li et al. (2024) adapted YOLOv8 to inspect cylindrical parts, a critical aspect of quality control in specialized manufacturing. Lastly, Hu et al. (2024) leveraged a conditioned version of YOLOv8, named Cond-YOLOv8-seg, to assess the uniformity of industrially produced materials, showcasing the model's versatility across different manufacturing scenarios. These innovations underscore the pivotal role of YOLO algorithms in driving forward the capabilities of industrial inspection systems, highlighting their impact on enhancing operational efficiency and product quality.

Additionally (Yang et al. 2024) introduced DCS-YOLOv8, a variant optimized for detecting steel surface defects, demonstrating its effectiveness in addressing the complexities of steel manufacturing. This adaptation ensures that even minor imperfections are identified, crucial for maintaining the structural integrity of steel products.Likewise, Wang et al. (2023) further refined YOLOv8 to develop BL-YOLOv8, focusing on road defect detection. This model enhances the safety and maintenance of transportation infrastructure by enabling more accurate and real-time detection of road surface anomalies. Similarly, Luo et al. (2023) presented a "Hardware-Friendly" YOLOv8 model designed for foreign object identification on belt conveyors, crucial for preventing equipment damage in materials handling. This version of YOLOv8 is tailored to perform well on the limited computational resources typical of industrial hardware systems. Finally, Wang et al. (2024) employed an improved YOLOv8 algorithm for the detection of defects in automotive adhesives, a critical quality control measure for ensuring vehicle safety and durability. These applications of YOLOv8 exemplify its adaptability and precision in industrial settings, where high accuracy and efficiency are paramount for operational success and safety compliance.

The recent advancements in YOLOv7 have paved the way for significant improvements in industrial inspection and monitoring systems. Wu et al. (2023) developed an enhanced YOLOv7 model specifically tailored for detecting objects in complex industrial equipment scenarios, highlighting its application in real-world settings (Wu et al. 2023). Similarly, Kim et al. (2022) implemented YOLOv7 in a real-time inspection system that leverages Moire patterns to detect defects in highly reflective injection molding products, demonstrating the algorithm's capability in manufacturing quality control (Kim et al. 2022). Further, Chen et al. (2023) explored the defect detection capabilities of YOLOv7 for automotive running lights, contributing to safer automotive systems through precise quality assurance techniques (Chen et al. 2023).

Hussain et al. (2022) applied domain feature mapping with YOLOv7 to automate inspections of pallet racking in storage facilities, enhancing safety and efficiency in logistics operations (Hussain et al. 2022). Zhu et al. (2023) extended YOLOv7's utility to the identification and classification of surface defects in belt grinding processes, aiding in maintaining the integrity of manufacturing workflows (Zhu et al. 2023). Lastly, Zhang et al. (2024) innovated with YOLO-RDP, a lightweight version of YOLOv7, optimized for detecting steel defects in real-time, showcasing the adaptability of YOLOv7 to resource-constrained environments and promoting sustainable manufacturing practices (Zhang 2024). Table 7 illustrates the different use of YOLO versions in the field of industrial manufacturing:

## 5.5 Agriculture

In agricultural environments, advanced object detection techniques such as YOLOv5 (Wang et al. 2022; Badgujar et al. 2023; Shoman et al. 2022), YOLOv6 (Bhat et al. 2023; Bist et al. 2023), YOLOv7 (Jiang et al. 2022; Kumar and Kumar 2023), YOLOv8 (Chen et al. 2023; Zhang et al. 2023) and YOLOv8 to YOLO11(Sharma et al. 2024) have proven to be instrumental in transforming traditional farming into smart, precision agriculture(Badgujar et al. 2024). YOLOv5, for example, has been adept at weed detection (Chen et al. 2023; Junior and Ulson 2021), enabling farmers to apply herbicides more effectively and economically by precisely identifying and localizing weed species amidst crops. This level of precision not only conserves resources but also mitigates the adverse environmental impact of excessive chemical usage. Furthermore, YOLOv6, YOLOv7 and YOLOv8 have enhanced capabilities in broader agricultural applications such as monitoring and analyzing crop health and growth patterns, significantly improving yield predictions and crop management strategies (Yu et al. 2024; Khalid et al. 2023; Gallo et al. 2023).

The recent introduction of YOLOv7 and YOLOv8 has further pushed the boundaries of agricultural innovation. YOLOv7 (Jia et al. 2023; Vaidya et al. 2023; Umar et al. 2024) and YOLOv8 (Zhang et al. 2023; Yue et al. 2023; Zayani et al. 2024) have been specifically refined to detect small pests and subtle disease symptoms on crops, which are often overlooked by human inspectors. Its enhanced deep learning framework allows for integrating complex image recognition tasks that facilitate early detection, thereby preventing widespread crop damage. On the other hand, YOLOv8 has made significant strides in fruit and other crop canopy detection tasks. Its application in orchards for detecting fruits such as apples and apple tree branches (Ma et al. 2024) supports optimal harvesting by determining the right stage of fruit maturity. This technique helps maximize the harvest quality and ensures that the fruits are picked at their nutritional peak, thereby enhancing their market value. The application of these advanced YOLO models including YOLOv5, YOLOv6, YOLOv7, and YOLOv8 represents a leap toward a more sustainable and efficient agricultural sector.

Recent studies have demonstrated the efficacy of YOLO-based models in enhancing various aspects of smart farming and agricultural automation solutions. Junos et al. (2021) optimized a YOLO-based object detection model to improve crop harvesting systems, showcasing the potential to boost yield and reduce labor costs (Junos et al. 2021). Zhao et al. (2024) extended this application to real-time object detection combined with robotic manipulation, further aligning agricultural practices with advanced automation technologies (Zhao et al. 2024). Chen et al. (2021) developed an apple detection method using a

**Table 7** Studies on YOLO applications in the manufacturing industry, focusing on real-time defect detection and process optimization for improved efficiency

| Title of paper | Description of work | Purpose and YOLO usage | Version | Refs. and Year |
|---|---|---|---|---|
| "Automated Dual-Side Leather Defect Detection and Classification Using YOLOv11: A Case Study in the Finished Leather Industry" | Explores leather defect detection on both the grain and flesh sides using YOLOv11, with a custom-designed light chamber to enhance quality control and reduce waste | Aims to improve defect detection accuracy and leather utilization in the leather industry by utilizing advanced AI capabilities of YOLOv11 for dual-side analysis | YOLOv11 | Banduka et al. (2024), 2024 |
| "A Novel YOLOv10-Based Algorithm for Accurate Steel Surface Defect Detection" | Presents YOLOv10n-SFDC, a new system improving steel surface defect detection by integrating DualConv, SlimFusionCSP modules, and Shape-IoU loss function for enhanced accuracy | Enhances steel manufacturing processes by offering a more accurate and efficient defect detection system, demonstrating significant improvements over traditional methods | YOLOv10 | Liao et al. (2025), 2025 |
| "An Improved YOLOv9 and Its Applications for Detecting Flexible Circuit Boards Connectors" | Enhances YOLOv9 with Multi-scale Dilated Attention and Deformable Large Kernel Attention for detecting defects in FPC connectors, improving feature capture and boundary adaptation | Aims to address challenges in automatic defect detection of flexible circuit board connectors by significantly enhancing detection precision and computational efficiency with modified YOLOv9 | YOLOv9 | Huang et al. (2024), 2024 |
| "YOLO-IMF: An Improved YOLOv8 Algorithm for Surface Defect Detection in Industrial Manufacturing Field" | Proposes an enhanced YOLOv8, YOLO-IMF, for surface defect detection on aluminum plates. Replaces CIOU with EIOU loss function to better handle small and irregularly shaped targets, achieving significant improvements in precision | Demonstrates YOLOv8's extended applicability in industrial settings by enhancing accuracy and defect detection capabilities | YOLOv8 | Liu and Ye (2023), 2023 |
| "YOLOv7-SiamFF: Industrial Defect Detection Algorithm Based on Improved YOLOv7" | Introduces YOLOv7-SiamFF, an advanced defect detection framework employing YOLOv7 with Siamese network enhancements for superior defect identification and background noise suppression | Enhances industrial defect detection by integrating attention mechanisms and feature fusion modules, achieving higher accuracy in pinpointing defect locations | YOLOv7 | Yi et al. (2024), 2024 |
| "A Novel Finetuned YOLOv6 Transfer Learning Model for Real-Time Object Detection" | Enhances real-time object detection by integrating a transfer learning approach with a pruned and finetuned YOLOv6 model, significantly boosting detection accuracy and speed | Focuses on improving YOLOv6 for efficient object detection in embedded systems, using advanced pruning techniques for reduced model size without sacrificing performance | YOLOv6 | Gupta et al. (2023), 2023 |
| "Real-time Tool Detection in Smart Manufacturing Using YOLOv5" | Utilizes YOLOv5 for advanced real-time tool detection in manufacturing environments, optimizing object detection capabilities for precise tool localization | Aims to enhance smart manufacturing by leveraging YOLOv5 for accurate and real-time detection of various tools, contributing significantly to Industry 4.0 initiatives | YOLOv5 | Zendehdel et al. (2023), 2023 |

**Table 7** (continued)

| Title of paper | Description of work | Purpose and YOLO usage | Version | Refs. and Year |
|---|---|---|---|---|
| "Efficient Automobile Assembly State Monitoring System Based on Channel-Pruned YOLOv4" | Implements a channel-pruned YOLOv4 algorithm to optimize monitoring in automobile assembly, enhancing detection speed without compromising accuracy | Designed to streamline assembly monitoring in industrial environments, showcasing YOLOv4's utility in enhancing operational efficiency and deployment readiness | YOLOv4 | Jiang et al. (2024), 2024 |
| "YOLO V3 + VGG16-based Automatic Operations Monitoring in Manufacturing Workshop" | Utilizes a combined YOLO V3 and VGG16 framework to recognize and monitor industrial operations accurately for Industry 4.0 manufacturing workshops | Aims to enhance production efficiency and quality by automating action analysis and process monitoring using advanced YOLO V3 and VGG16 technologies | YOLO V3, VGG16 | Yan and Wang (2022), 2022 |
| "Improvements of Detection Accuracy by YOLOv2 with Data Set Augmentation" | Employs YOLOv2 with an innovative data set augmentation method to enhance the detection accuracy and confidence in identifying defective areas in industrial products | Seeks to optimize defect detection and visualization on production lines, demonstrating YOLOv2's effectiveness with limited data augmentation options | YOLOv2 | Arima et al. (2023), 2023 |

tailored YOLOv4 algorithm, specifically designed to support harvesting robots operating in complex environments, which significantly enhances the precision and efficiency of fruit picking (Chen et al. 2021).

Further contributions include work by Nergiz (2023), who utilized YOLOv7 to enhance strawberry harvesting efficiency, providing practical solutions for small to medium-sized enterprises in the agricultural (NERGİZ 2023). Wang et al. (2024) focused on planning harvesting operations in large strawberry fields using a deep learning-based image processing method, demonstrating the scalability of YOLO for larger agricultural operations (Wang et al. 2024). Lastly, Zhang et al. (2023) introduced DCF-YOLOv8, an improved algorithm for agricultural pests and diseases detection by aggregating low-level features, which helps in early detection and management of crop health (Zhang et al. 2023). These studies collectively illustrate the transformative impact of YOLO-based models in modernizing agricultural practices, ensuring higher productivity and sustainability.

In orchard automation, the YOLO object detection models have been specifically pivotal in enhancing the accuracy and efficiency of fruit detection (Chen et al. 2022; Mirhaji et al. 2021; Sapkota et al. 2024c), flower identification (Wu et al. 2020; Wang et al. 2021; Khanal et al. 2023), and automated harvesting processes (Xiao et al. 2023; Junos et al. 2021; Yijing et al. 2021). These models adeptly identify and classify fruits at various stages of ripeness, detect flowers and other canopy objects such as branches with high precision, and facilitate efficient harvesting operations. The development of YOLO models has introduced significant improvements that cater specifically to the challenges of agricultural environments. For instance, YOLOv5's introduction of multi-scale predictions improved the detection of small and clustered objects like flowers and young fruits, which are critical during the early stages of crop yield management (Zhang et al. 2024). As the models advanced, YOLOv7 and YOLOv8 incorporated better segmentation techniques, which enhanced the differentiation between fruit types and maturity stages, critical for targeted harvesting (Zhou et al. 2023; Xiuyan and ZHANG 2023).

In addition to their application in fruit detection and harvesting, YOLO object detection models are increasingly vital across other agricultural practices such as pruning (Sapkota et al. 2024a), thinning (Sapkota et al. 2024c), pollination and harvest (He et al. 2021) management. In pruning, YOLO models facilitate the accurate identification of non-fruiting branches, aiding in the automation of pruning tasks to optimize plant health and productivity. For thinning processes, these models can distinguish between fruit clusters, allowing for precise thinning that improves fruit size and quality at harvest. Furthermore, YOLO models are being adapted to recognize pollination patterns, helping to monitor and enhance pollination efficiency, which is crucial for maximizing crop yield. Each advancement in YOLO technology contributes to more sustainable and effective agricultural operations, underscoring the critical role of AI-driven technologies in modern farming practices.

Moreover, recent iteration, YOLOv9 have leveraged advanced algorithms with spatial pyramid pooling and attention mechanisms, which have refined the detection capabilities in plant disease detection (Boudaa et al. 2024). Boudaa et al. (2024) performed a comparative study on different important versions of YOLO (v5, v8 and v9) on a real-world dataset for tomato plant disease detection and suggested that YOLOv9 outperforms YOLOv5 and YOLOv8. A study on weed species control was presented by Sharma et al. (2024), where a comparative analysis of different YOLO versions (v8, v9, v10, and 11) and Faster R-CNN was conducted. The study utilized an annotated image database containing five weed spe-cies: cocklebur (Xanthium strumarium L.), dandelion (Taraxacum officinale), common waterhemp (Amaranthus tuberculatus), Palmer amaranth (Amaranthus palmeri), and common lambsquarters (Chenopodium album L.).

It is also noted that autonomous driving has been an active area of research and development in agriculture as well (Andreyanov et al. 2022; Jung et al. 2020), similar to self-driving cars. YOLO models have played crucial roles in advancing this technology for farming for example in peach crop (Xu and Rai 2024). Object detection based on YOLO has enhanced the navigation systems of autonomous agricultural vehicles, enabling them to operate efficiently in diverse field conditions. By accurately detecting and classifying field boundaries, obstacles such as rocks and trees, and other critical elements like rows of crops and water sources, YOLO-powered systems significantly improve the precision and safety of operations. This capability is particularly vital in precision agriculture, where exactness in planting, fertilizing, and treatment applications can substantially impact crop health and yield. The integration of YOLO models into agricultural drones and robots also supports tasks such as crop monitoring, disease detection, and targeted pesticide application, further automating and optimizing farm management practices (Alibabaei et al. 2022).

Table 8 illustrates the different use of YOLO versions in the field of Agriculture:

## 6 Challenges and limitations

### 6.1 Challenges and limitations of each version

YOLOv12:

- Computational efficiency and performance trade-offs: YOLOv12 introduces the Area Attention ($A^2$) module and FlashAttention mechanism, which, while theoretically efficient, present practical challenges. Empirical studies demonstrate a reduction in inference speed (30 FPS compared to YOLO11's 40 FPS) and an increase in training time by approximately 20% (Tian et al. 2025). The effectiveness of these attention mechanisms varies with hardware architecture, potentially limiting performance gains on certain systems. Moreover, YOLOv12 requires lower confidence thresholds ($0.3-0.4$) for effective detection, which may increase false positive rates. These factors collectively contribute to the ongoing challenge of optimizing the trade-off between detection accuracy and real-time performance, particularly in resource-constrained environments and complex detection scenarios.
- Detection limitations and environmental sensitivity: Despite architectural advancements, YOLOv12 continues to face challenges in accurately detecting small objects and objects in crowded scenes, a limitation persistent from previous iterations. This is primarily due to three factors: 1) The downsampling process in YOLO architectures can lead to loss of fine-grained details, particularly affecting small object detection (Diwan et al. 2023), 2) Limited contextual information in a single forward pass makes it difficult for the model to differentiate between closely packed or overlapping objects in crowded scenes (Ji et al. 2023; Fang et al. 2019), and 3) The use of fixed anchor boxes with predefined scales and aspect ratios may not optimally capture objects with highly variable sizes or unusual shapes (Sirisha et al. 2023).

**Table 8** Studies on YOLO usage in agriculture, emphasizing automated crop monitoring, pest detection, and yield estimation for enhanced productivity

| Title of paper | Description of work | Purpose and YOLO usage | Version | Refs. and Year |
|---|---|---|---|---|
| "Improved YOLOv12 with LLM-Generated Synthetic Data for Enhanced Apple Detection and Benchmarking Against YOLOv11 and YOLOv10" | Evaluates YOLOv12 for apple detection in commercial orchards using LLM-generated synthetic datasets and benchmarks it against YOLOv11 and YOLOv10 | Demonstrates that YOLOv12 achieves the highest precision (0.916), recall (0.969), and mAP@50 (0.978), surpassing YOLOv11 and YOLOv10 in accuracy while YOLO11 remains the fastest. Highlights efficiency in processing speeds and cost-effectiveness by reducing the need for extensive manual data collection | YOLOv12 | Sapkota and Karkee (2025), 2025 |
| "Synthetic meets authentic: Leveraging llm generated datasets for YOLO11 and YOLOv10-based apple detection through machine vision sensors" | Utilizes LLM-generated datasets to train YOLOv10 and YOLO11 for detecting apples, showcasing significant improvements in detection metrics and processing times | Aims to streamline data collection and enhance object detection in orchards with minimal fieldwork by employing YOLO11 and YOLOv10, demonstrating superior performance in real-world tests | YOLO11 | Sapkota et al. (2024b), 2024 |
| "Comprehensive Performance Evaluation of YOLO11, YOLOv10, YOLOv9 and YOLOv8 on Detecting and Counting Fruitlet in Complex Orchard Environments" | Conducts a thorough performance evaluation of YOLOv8, YOLOv9, YOLOv10, and YOLO11 on green fruit detection and counting across multiple configurations, highlighting in-field validation with an iPhone and machine vision sensors | Provides comparative insights on various YOLO configurations for optimizing fruitlet detection in commercial orchards, recommending YOLO11 for its speed and accuracy | YOLO11 | Sapkota et al. (2024d), 2024 |
| "YOLOv10-pose and YOLOv9-pose: Real-time Strawberry Stalk Pose Detection Models" | Introduces YOLOv10-pose and YOLOv9-pose for high-precision strawberry stalk pose detection, comparing them with previous YOLO versions to optimize agricultural automation tasks | Enhances the efficiency of robotic harvesting and other agricultural applications by providing accurate pose detection, crucial for automated operations in the agricultural industry | YOLOv10 | Meng et al. (2025), 2025 |
| "Comparing YOLOv11 and YOLOv8 for Instance Segmentation of Occluded and Non-occluded Immature Green Fruits in Complex Orchard Environment" | Evaluates YOLOv11 and YOLOv8 for instance segmentation capabilities of immature green fruits, focusing on occluded and non-occluded scenarios in orchards | Enhances understanding of YOLO11 and YOLOv8's segmentation performance, particularly their efficacy in detecting and segmenting immature green fruits amidst complex environmental conditions | YOLO11 | Sapkota and Karkee (2024), 2024 |
| "Automating Tomato Ripeness Classification and Counting with YOLOv9" | Implements YOLOv9 to automate and enhance the accuracy of classifying and counting ripe tomatoes, replacing labor-intensive visual inspections | Aims to streamline tomato ripeness monitoring and counting, to enhance agricultural productivity and quality. Utilizes YOLOv9 for high accuracy in detection | YOLOv9 | Vo et al. (2024), 2024 |

**Table 8** (continued)

| Title of paper | Description of work | Purpose and YOLO usage | Version | Refs. and Year |
|---|---|---|---|---|
| "A Lightweight YOLOv8 Tomato Detection Algorithm Combining Feature Enhancement and Attention" | Enhances YOLOv8 for tomato detection in agriculture using depthwise separable convolution and dual-path attention gate modules. Optimizes real-time detection for robotic tomato picking | Aims to advance agricultural automation by boosting YOLOv8's efficiency and accuracy in tomato harvesting. Demonstrates improved performance over earlier YOLO versions | YOLOv8 | Yang et al. (2023), 2023 |
| "An Attention Mechanism-Improved YOLOv7 Object Detection Algorithm for Hemp Duck Count Estimation" | Implements CBAM-YOLOv7 to enhance feature extraction capabilities within YOLOv7 for precise hemp duck counting in agriculture, outperforming SE-YOLOv7 and ECA-YOLOv7 in precision and mAP | Enhances livestock management by automating duck count with advanced object detection, reducing labor and improving accuracy | YOLOv7 | Jiang et al. (2022), 2022 |
| "Detecting Crops and Weeds in Fields Using YOLOv6 and Faster R-CNN Object Detection Models" | Utilizes YOLOv6 and Faster R-CNN to detect crops and weeds for precise management | Aims to boost agricultural productivity and environmental sustainability by improving accuracy in weed detection using YOLOv6 | YOLOv6 | Bhat et al. (2023), 2023 |
| "An improved YOLOv5-based vegetable disease detection method" | Enhances YOLOv5 for precise detection of vegetable diseases by upgrading CSP, FPN, and NMS modules to handle complex environmental interference | Aims to improve food security by boosting the accuracy and speed of disease detection in vegetables using an improved YOLOv5 algorithm | YOLOv5 | Li et al. (2022), 2022 |
| "Using channel pruning-based YOLO v4 deep learning algorithm for the real-time and accurate detection of apple flowers in natural environments" | Implements a channel pruned YOLOv4 model to enhance efficiency and accuracy in detecting apple flowers, supporting the development of flower thinning robots | Aims to optimize apple flower detection in orchards by applying channel pruning to YOLOv4, significantly reducing model size and improving processing speed while maintaining high accuracy | YOLOv4 | Wu et al. (2020), 2020 |
| "Fast and accurate detection of kiwifruit in orchard using improved YOLOv3-tiny model" | Enhances YOLOv3-tiny with additional convolutional kernels for improved kiwifruits detection in orchards, in occlusions and varying lighting conditions | Focus on increasing the efficiency of kiwifruit detection in dynamic orchard environments with a modified YOLOv3-tiny, demonstrating high performance | YOLOv3-tiny | Fu et al. (2021), 2021 |
| "A Detection Method for Tomato Fruit Common Physiological Diseases Based on YOLOv2" | Implements YOLOv2 to detect and identify healthy and diseased tomato, using advanced image processing and data augmentation to enhance detection accuracy | Aims to boost tomato yield and quality control through efficient detection of physiological diseases, demonstrating the effectiveness of YOLOv2 in agriculture | YOLOv2 | Zhao and Qu (2019), 2019 |

**Table 8** (continued)

| Title of paper | Description of work | Purpose and YOLO usage | Version | Refs. and Year |
|---|---|---|---|---|
| "A Vision-Based Counting and Recognition System for Flying Insects in Intelligent Agriculture" | Utilizes YOLO for initial detection and counting, and SVM for fine classification of flying insects, for efficient insect pest control | Demonstrates a robust, efficient system for insect monitoring, greatly enhancing accuracy and speed in pest management | YOLO, SVM | Zhong et al. (2018), 2018 |
| "Comparative performance of YOLOv8, YOLOv9, YOLOv10, YOLO11 and Faster R-CNN models for detection of multiple weed species" | Compare the performance of YOLOv8, YOLOv9, YOLOv10, YOLO11, and Faster R-CNN algorithms in terms of speed and accuracy | Demonstrates the robust efficiency of several YOLO versions in detecting weed species using a database comprising cocklebur (Xanthium strumarium L.), dandelion (Taraxacum officinale), common waterhemp (Amaranthus tuberculatus), Palmer amaranth (Amaranthus palmeri), and common lambsquarters (Chenopodium album L.) | YOLOv8, YOLOv9, YOLOv10, YOLOv11 and Faster R-CNN | Sharma et al. (2024), 2024 |

These issues contribute to the model's potential struggle with objects at vastly different scales within the same image, indicating a need for improved scale-invariant feature extraction.

YOLOv11:

As the most recent addition to the YOLO series, this version must address and overcome

- Challenges in Detecting Small and Rotated Objects: Despite advancements, YOLOv11s-till struggles with small, low-resolution objects and those with varied orientations. This limitation is due to its architectural constraints, which may not fully capture the complexities of such objects, leading to potential inaccuracies in detection.
- Susceptibility to Overfitting: YOLOv11 is prone to overfitting, particularly when trained on limited or homogeneous datasets. This overfitting can adversely affect the model's performance on new or varied datasets, indicating a need for improved generalization capabilities and robust training approaches.
- Computational Efficiency vs. Accuracy Trade-off: While YOLOv11 has improved computational efficiency, there remains a trade-off with accuracy, particularly in complex detection environments. This trade-off highlights the ongoing challenge of balancing speed and accuracy to support real-time applications.

YOLOv10:

- YOLOv10 has not yet seen widespread adoption in published research. Its release promises cutting-edge improvements in object detection capabilities, but the lack of extensive testing and real-world application data makes it difficult to ascertain its full potential and limitations.
- Preliminary evaluations suggest that while YOLOv10 might offer advancements in speed and accuracy, integrating it into existing systems could present challenges due to compatibility and computational demands. Potential users may hesitate to adopt this version until more comprehensive studies and benchmarks are available, which articulate its advantages over previous models.
- The expectation with YOLOv10, much like its predecessors, is that it will drive further research in object detection technologies. Its eventual widespread implementation could pave the way for addressing complex detection scenarios with higher accuracy, particularly in dynamic environments. However, as with any new technology, the adaptation phase will be crucial in understanding its practical limitations and operational challenges.

YOLOv9:

- Despite YOLOv9's enhancements in detection capabilities, it has only been featured in a handful of studies, which limits a comprehensive understanding of its performance across diverse applications. This lack of extensive validation may deter organizations from adopting it until more empirical evidence and comparative analyses establish its efficacy and efficiency over earlier versions.
- YOLOv9 significantly rectified the computational efficiency challenges faced by

YOLOv8. YOLOv9 introduces several improvements that enhance its efficiency while maintaining or improving accuracy. The model achieves a 49% reduction in parameters and a 43% reduction in computational load compared to YOLOv8, while improving accuracy by 0.6% on benchmark dataset. The introduction of the Generalized Efficient Layer Aggregation Network and Programmable Gradient Information enhances feature extraction and gradient flow, leading to improved efficiency. YOLOv9-E demonstrates a 16% reduction in parameters and a 27% reduction in FLOPs compared to YOLOv8-X, while also gaining a 1.7% improvement in mAP value.

- While YOLOv9 improves upon the speed and accuracy of its predecessors, it may still struggle with detecting small or overlapping objects in cluttered scenes. This is a recurring challenge in high-density environments like crowded urban areas or complex natural scenes in transportation and agriculture, where precise detection is critical for applications such as autonomous driving, wildlife monitoring and robotic fruit picking.
- Future developments for YOLOv9 could focus on enhancing its robustness in adverse conditions, such as varying weather, lighting, or occlusions. Integrating more adaptive and context-aware mechanisms could help in mitigating false positives and improving the reliability of the system under different operational conditions. The implementation of advanced training techniques such as federated learning could also be explored to enhance its adaptability and learning efficiency from decentralized data sources.

YOLOv8:

- YOLOv8 has shown significant improvements in object detection tasks, particularly in real-time applications. However, it continues to face challenges in terms of computational efficiency and resource consumption when deployed on lower-end hardware (Ye et al. 2024). This can limit its applicability in resource-constrained environments where deploying advanced hardware solutions is not feasible (Soylu and Soylu 2024).
- The future direction for YOLOv8 could involve optimizing its architectural design to reduce computational load without compromising detection accuracy. Enhancing its scalability to efficiently process images of varying resolutions and conditions can broaden its application scope. Moreover, incorporating adaptive scaling and context-aware training methods could potentially address the detection challenges in complex scenes, making it more robust against diverse operational challenges.

YOLOv7:

- Although YOLOv7 introduces significant improvements in detection accuracy and speed, its adoption across varied real-world applications reveals a persistent challenge in handling highly dynamic scenes. For instance, in environments with rapid motion or in scenarios involving occlusions, YOLOv7 can still experience drops in performance. The algorithm's ability to generalize across different types of blur and motion artifacts remains an area for further research and enhancement.
- The complexity of YOLOv7's architecture, while beneficial for accuracy, imposes a substantial computational burden. This makes it less ideal for deployment on edge devices or platforms with limited processing capabilities, where maintaining a balance between speed and power efficiency is crucial (Olorunshola et al. 2023; AFFES et al.

2023). Efforts to streamline the model for such applications without significant loss of performance are necessary.

- Looking forward, there is significant potential in expanding YOLOv7's capabilities through the integration of semi-supervised or unsupervised learning paradigms. This would enable the model to leverage unlabeled data effectively, a common challenge in the real-world where annotated datasets are often scarce or expensive to produce. Additionally, enhancing the model's resilience to adversarial attacks and variability in data quality could further solidify its utility in security-sensitive applications like surveillance and fraud detection.

YOLOv6:

- One of the notable challenges with YOLOv6 is its handling of scale variability within images, which can affect its efficacy in environments where objects appear at diverse distances from the camera. While YOLOv6 shows improved accuracy and speed over its predecessors, it sometimes struggles with small or partially occluded objects, which are common in crowded scenes or complex industrial environments (Norkobil Saydirasulovich et al. 2023; Li et al. 2023). This limitation can be critical in applications such as automated surveillance or advanced manufacturing monitoring.
- YOLOv6, while efficient, still requires considerable computational resources when compared to other models optimized for edge devices. Its deployment in resource-constrained environments such as mobile or embedded systems often requires a trade-off between detection performance and operational efficiency. Further optimizations and model pruning are necessary to achieve the best of both worlds-real-time performance with reduced computational demands.
- Future enhancements for YOLOv6 could focus on incorporating more advanced feature extraction techniques that improve its robustness to variations in object appearance and environmental conditions. Additionally, integrating more adaptive and context-aware learning mechanisms could help overcome some of the challenges related to background clutter and similar adversities. Enhancing the model's capacity to learn from a limited number of training samples, through techniques such as few-shot learning or transfer learning, could address the scarcity of labeled training data in specialized applications.

YOLOv5:

- YOLOv5 has made significant strides in improving detection speed and accuracy, but it faces challenges in consistently detecting small objects due to its spatial resolution constraints. This is particularly evident in fields such as medical imaging or satellite image analysis, where precision is crucial for identifying fine details. Techniques such as spatial pyramid pooling or enhanced up-sampling may be needed to increase the receptive field and improve the detection of smaller objects without compromising the model's efficiency (Benjumea et al. 2021; Jung and Choi 2022; Wang et al. 2022).
- While YOLOv5 offers faster training and inference times compared to previous versions, its deployment on edge devices is limited by high memory and processing requirements (Wu et al. 2021; Jia et al. 2023). Although optimized models like YOLOv5s provide a solution, they sometimes do so at the cost of detection accuracy. Optimizing network architecture through neural architecture search (NAS) could potentially offer a more balanced solution, enhancing both performance and efficiency for real-time object detection applications.

- The adaptability of YOLOv5 to varied environmental conditions and different types of data distribution remains an area for development. Future research could focus on enhancing the robustness of YOLOv5 through advanced data augmentation techniques and domain adaptation strategies. This would enable the model to maintain high accuracy levels across diverse application settings, from urban surveillance to complex natural environments, effectively handling variations in lighting, weather, and seasonal changes.

YOLOv4, YOLOv3, YOLOv2 and YOLOv1:

- While YOLOv4 introduced notable enhancements in speed and accuracy, it still exhibits performance inconsistencies across different datasets, particularly with class imbalance and the detection of rare objects. The model's high computational demand also restricts its deployment on low-power devices. Continued efforts to improve model compression and increase adaptability to varying environmental conditions are essential to extend its practical utility in diverse real-world applications.
- YOLOv3 improved upon the balance of speed and accuracy, yet it struggles with small object detection due to its grid limitation. Its computational efficiency poses challenges for deployment in resource-constrained environments, prompting research towards optimization techniques to improve efficiency without sacrificing performance. Additionally, enhancing the model's robustness to environmental variations could improve its reliability for applications like autonomous driving and urban surveillance.
- Despite the incremental improvements introduced in YOLOv2, it faces challenges in detecting small objects, balancing speed with accuracy, and maintaining relevance with the advent of more capable successors. This version's reliance on a fixed grid system hampers its ability to perform in high-precision detection tasks. Future developments may shift towards adapting YOLOv2's core strengths in new architectures that enhance its spatial resolution and dynamic scaling capabilities.

For the versions of YOLO under YOLOv5, their use may decrease and discontinue in the future as newer versions are replacing the older YOLO versions in overall performance and efficiency.

- The potential for YOLOv4, YOLOv3, and YOLOv2 in future research involves exploring adaptive mechanisms that can tailor learning rates and augment data to better handle diverse operational scenarios. Integrating these models with newer technologies like model pruning and feature fusion may address existing inefficiencies and extend their applicability to a wider range of applications.
- YOLOv1 was revolutionary for its time, introducing real-time object detection by processing the entire image at once as a single regression problem. However, it faces significant challenges in dealing with small objects due to each grid cell predicting only two boxes and the probabilities for the classes. This structure often leads to poor performance on groups of small objects that are close together, such as flocks of birds or traffic scenes with multiple vehicles at a distance. Improvements in subsequent models focus on increasing the number of predictions per grid and incorporating finer-grained feature maps to enhance small object detection.

- Another limitation of YOLOv1 is the spatial constraints of its bounding boxes. Since each cell in the grid can only predict two boxes and has limited context about its neighboring cells, the precision in localizing objects, especially those with complex or irregular shapes, is often compromised. This challenge is particularly evident in medical imaging and satellite image analysis, where the exact contours of the objects are crucial. Advances in convolutional neural network designs and cross-layer feature integration in later versions seek to address these drawbacks.
- Although YOLOv1 laid the groundwork for real-time object detection, its direct usage has significantly diminished, with advancements in the field largely driven by more recent iterations such as YOLOv4 and beyond. These newer models have not only retained the core principles of YOLOv1 but have also introduced improved mechanisms for handling diverse object sizes and aspect ratios. Current and future research is less likely to concentrate on YOLOv1 and earlier versions like YOLOv3, but rather on advancing these later iterations or developing hybrid models that might incorporate elements of YOLOv1's architecture to benefit applications where high speed and low latency are paramount, despite potential trade-offs in detection precision and detail.
- Future iterations could focus on dynamic grid systems, lighter network architectures, and advanced scaling features to tackle the challenges of small object detection and computational limitations. These improvements could enhance their deployment in emerging areas such as edge computing, where real-time processing and low power consumption are crucial.
- As YOLO continues to evolve with newer iterations like YOLOv8 and YOLOv9, the core principles of earlier versions such as YOLOv4, YOLOv3, and YOLOv2 still hold significant value for developing hybrid models and specialized applications. The research community is increasingly focused on harnessing the rapid detection capabilities of these older versions while addressing their limitations in detection accuracy through composite and hybrid modeling strategies. This trend is evidenced by innovations that integrate YOLO with other architectures to enhance overall performance. For example, a hybrid model that combines CNNs, YOLO, and Vision Transformers (ViTs) has demonstrated enhanced detection accuracy and reduced inference times by utilizing CNNs for robust feature extraction, YOLO for quick object detection, and ViTs for capturing global context (Ali et al. 2024). Similarly, the DA-ActNN-YOLOv5 model merges YOLOv5 with advanced data augmentation and ActNN's model compression techniques to optimize both accuracy and efficiency across diverse operational environments (Zhu et al. 2021; Bashir et al. 2023)

### 6.2 Challenges in statistical metrics for evaluation

Threat: Evaluating YOLO detection systems requires a unique approach, as each version, from the original YOLO to the latest YOLOv12, targets different aspects of detection capability, such as speed, accuracy, or computational efficiency. For a comprehensive evaluation, it is essential to employ a diverse array of metrics, including precision, recall, GFLOPs, and model size. This approach allows for a more complete comparison and understanding of each model's strengths and weaknesses in various real-world applications. This multi-metric evaluation is crucial to assessing the practical utility and technological advancement of the YOLO series. Future YOLO versions are expected to introduce novel evaluation metrics that capture emerging capabilities in edge computing, multi-modal fusion, and adaptive architecture optimization, necessitating an even more sophisticated evaluation framework to accurately assess their performance across diverse deployment scenarios.

Mitigation: Despite this limitation, our main premise is that the selected metrics enable us to compare various YOLO systems and adequately assess their overall effectiveness. Recognizing the inherent limitations of statistical summaries is crucial when conducting a comprehensive evaluation of detection systems across different applications. Therefore, we aim to improve the clarity and reliability of our review by openly acknowledging these potential threats to construct validity. This approach provides a more nuanced understanding of the limitations associated with various aspects of YOLO techniques for object detection in diverse domains.

Spectral versus RGB images: Beyond traditional RGB imaging, spectral features encompass a broader spectrum, including infrared, ultraviolet, and even multispectral and hyperspectral imaging. These advanced spectral techniques can significantly enhance YOLO's object detection capabilities by providing additional information not visible in the RGB spectrum. For example, hyperspectral imaging can detect subtle variations in plant health for agricultural applications or distinguish between materials based on their spectral signatures in industrial settings. This expansion into wider spectral data not only improves detection accuracy but also opens up new avenues for application-specific optimizations, reinforcing YOLO's versatility and potential across various fields.

Over the past decade, the series of YOLO models have significantly impacted various sectors, demonstrating the powerful capabilities of deep learning in real-world applications. As a pioneering object detection algorithm, YOLO has facilitated rapid advancements across diverse fields by offering high-speed, real-time detection with commendable accuracy. One of the most notable applications has been in public safety and surveillance, where YOLO models have improved the efficacy of monitoring systems, enhancing the detection of suspicious activities and ensuring public safety more efficiently. In the realm of automotive technology, YOLO has been integral in developing advanced driver-assistance systems (ADAS) (Malligere Shivanna and Guo 2024), contributing to object detection that supports collision avoidance systems and pedestrian safety. Furthermore, YOLO has transformed the healthcare sector by accelerating medical image analysis, enabling quicker and more accurate detection of pathologies which is critical for diagnostics and treatment planning. In industrial settings, YOLO has optimized quality control processes by identifying defects in manufacturing lines in real-time, thereby reducing waste and increasing production efficiency. Additionally, in the retail sector, YOLO has supported inventory management through automated checkouts and stock monitoring, enhancing customer experience and operational efficiency, whereas in agriculture, YOLO has played a key role to enhance timely crop stress detection, pest localization and precision crop management while improving worker health and safety.

# 7 Future directions in object detection with YOLO

## 7.1 YOLO deployment on edge and IoT devices

The deployment of YOLO on edge devices unlocks several promising avenues for future research and development. One potential direction involves enhancing the algorithm's efficiency and accuracy for even more constrained environments, such as ultra-low-power Microcontrollers and embedded systems. This can be achieved through further optimization techniques, including model pruning, quantization, and the development of specialized hardware accelerators. Additionally, integrating YOLO with advanced communication protocols, edge computing frameworks and IoT devices could facilitate more seamless collaboration between edge devices and centralized cloud services, enhancing the overall system performance and scalability. Exploring the integration of YOLO with other AI-driven functionalities, such as anomaly detection and predictive analytics, may unlock new applications in areas like healthcare, smart cities, and industrial automation. As edge computing continues to evolve, the adaptation of YOLO to support federated learning paradigms could ensure the data privacy while enabling continuous learning and improvement of object detection models. These future directions will not only expand the capabilities of YOLO but also contribute significantly to the advancement of intelligent edge computing systems (Yang et al. 2022; Ghaziamin et al. 2024; Hussain et al. 2022; Zhang 2024).

## 7.2 YOLO and embodied artificial intelligence

Embodied Artificial Intelligence (EAI) refers to AI systems integrated with physical entities or bodies, enabling them to interact with the real world in a natural and human-like manner (Pfeifer and Iida 2004). Incorporating YOLO into these systems significantly enhances their sensory capabilities, allowing for more efficient and accurate interaction with the physical environment. Applications of YOLO in EAI include autonomous vehicles, drones, robots (Sanket 2021), human-robot interaction (Wang et al. 2024). Additionally, it plays a significant role in healthcare, particularly with robotic surgical assistants (Lakshmipathy et al. 2024), among other innovative uses (Li et al. 2024).

# 8 Expanding YOLO object detection into broader AI domains

## 8.1 YOLO and artificial general intelligence

Artificial General Intelligence (AGI) refers to an intelligent agent with human-level or higher intelligence, capable of solving a variety of complex problems in diverse domains (Pande et al. 2024; Qu et al. 2024). In this context, an AGI system would need to integrate object detection capabilities, similar to those provided by YOLO, with other essential cognitive functions, such as advanced natural language understanding, reasoning, and decision-making. This fusion will enable the system to handle a broad spectrum of tasks in real-time, adapting to dynamic environments and complex scenarios effectively, thus advancing the current AI systems towards achieving a true AGI.

### 8.2 YOLO integration with large language models

One effective way to advance AGI capabilities is to integrate YOLO with Large Language Models (LLMs) by seamlessly merging advanced visual data interpretation with sophisticated natural language understanding, reasoning, and contextual awareness. This fusion would allow the AGI to not only recognize and analyze objects in real-time but also engage in meaningful interactions with stakeholders (or end users), making more informed decisions and adapting to complex tasks with greater autonomy and precision. This synergy would enable AGI systems to operate in complex environments handling multi-modal inputs simultaneously, such as navigating through an environment using visual cues identified by YOLO while interpreting and acting on spoken commands through capabilities provided by LLMs (Rouhi et al. 2025; Sapkota et al. 2024e, b). Such integration is expected to lead to a highly versatile and intelligent system, capable of performing real-time, multi-faceted operations across diverse application domains. By combining advanced visual recognition with robust language processing, it brings us a step closer to realizing true AGI, where the system can autonomously adapt, learn, and perform complex tasks with human-like flexibility and reasoning. Future YOLO versions, such as YOLOv13, YOLOv14, YOLOv15, and beyond, are anticipated to advance toward AGI, integrating enhanced reasoning, adaptability, and autonomous learning capabilities beyond traditional object detection.

## 9 YOLO and environmental impact

Training and retraining YOLO is extremely energy-intensive, leading to substantial energy and water consumption, as well as significant carbon dioxide emissions. This environmental impact underscores concerns about the sustainability of AI development, emphasizing the urgent need for more efficient practices to reduce the ecological footprint of large-scale model training (Xu et al. 2024; Dhar 2020).

## 10 Conclusion

In this comprehensive review, we explored the evolution of the YOLO models from the most recent YOLOv12 to the inaugural YOLOv1, including alternative versions of YOLO as YOLO-NAS, YOLO-X, YOLO-R, DAMO-YOLO, and Gold-YOLO. This retrospective analysis covered a decade of advancements, highlighting theapplied use of each version and their respective impacts across five critical application areas: autonomous vehicles and traffic safety, healthcare and medical imaging, security and surveillance, manufacturing, and agriculture. Our review outlined the significant enhancements in detection speed, accuracy, and computational efficiency that each iteration brought, while also addressing the specific challenges and limitations faced by earlier versions. Furthermore, we identified gaps in the current capabilities of YOLO models and proposed potential directions for future research, such as trade-off between detection speed versus accuracy, handling small and overlapping Objects, and generalization across diverse datasets and domains. Predicting the trajectory of YOLO's development, we anticipate a shift towards multimodal data processing, leveraging advancements in large language models and natural language processing to enhance object detection systems. This fusion is expected to broaden the utility of YOLO models, enabling more sophisticated, context-aware applications that could revolutionize the interaction between AI systems and their environments using Generative AI and multi-modal LLMs. Thus, this review not only serves as a detailed chronicle of YOLO's evolution but also sets a prospective blueprint for its integration into the next generation of technological innovations.

**Acknowledgements** This work was supported in part by the National Science Foundation (NSF) and the United States Department of Agriculture (USDA), National Institute of Food and Agriculture (NIFA), through the "Artificial Intelligence (AI) Institute for Agriculture" program under Award Numbers AWD003473 and AWD004595, and USDA-NIFA Accession Number 1029004 for the project titled "Robotic Blossom Thinning with Soft Manipulators." Additional support was provided through USDA-NIFA Grant Number 2024-67022-41788, Accession Number 1031712, under the project "Expanding UCF AI Research To Novel Agricultural Engineering Applications (PARTNER)." Additionally, this work is partially supported by the Hong Kong Innovation and Technology Commission (InnoHK Project CIMDA) and by EcuTS2025 from UFA-ESPE.

**Author contributions** Ranjan Sapkota: principal conceptualizer, research design, formal analysis, original draft preparation, manuscript writing, and editing. Marco Flores-Calero, Rizwan Qureshi, Chetan Badgujar, Upesh Nepal, Alwin Poulose, Peter Zeno, Uday Bhanu Prakash Vaddevolu, Sheheryar Khan, Maged Shoman, Hong Yan: methodology refinement, critical revisions, manuscript review, and editing. Manoj Karkee: Funding and supervision, methodology refinement, critical revisions, manuscript review, and editing. Ranjan Sapkota and Manoj Karkee: Corresponding authors.

**Data availability** No datasets were generated or analysed during the current study.

## Declarations

**Conflict of interest** The authors declare no conflict of interest.

## References

Liu L, Ouyang W, Wang X, Fieguth P, Chen J, Liu X, Pietikäinen M (2020) Deep learning for generic object detection: a survey. Int J Comput Vision 128:261–318

Badgujar CM, Poulose A, Gan H (2024) Agricultural object detection with You Only Look Once (YOLO) algorithm: a bibliometric and systematic literature review. Comput Electron Agric 223:109090. https://doi.org/10.1016/j.compag.2024.109090

Ahmad HM, Rahimi A (2022) Deep learning methods for object detection in smart manufacturing: a survey. J Manuf Syst 64:181–196

Gheorghe C, Duguleana M, Boboc RG, Postelnicu CC (2024) Analyzing real-time object detection with YOLO algorithm in automotive applications: a review. CMES - Comput Model Eng Sci 141(3):1939–1981 https://doi.org/10.32604/cmes.2024.054735

Arkin E, Yadikar N, Xu X, Aysa A, Ubul K (2023) A survey: object detection methods from CNN to transformer. Multim Tools Appl 82(14):21353–21383

Fernandez RAS, Sanchez-Lopez JL, Sampedro C, Bavle H, Molina M, Campoy P (2016) Natural user interfaces for human-drone multi-modal interaction. In: 2016 International Conference on Unmanned Aircraft Systems (ICUAS), pp. 1013–1022. IEEE
Wang RJ, Li X, Ling CX (2018) PELEE: A real-time object detection system on mobile devices. Adv Neural Inform Process Syst 31
Ren S, He K, Girshick R, Sun J (2015) Faster R-CNN: Towards real-time object detection with region proposal networks. Adv Neural Inform Process Syst 28
Tang J, Ye C, Zhou X, Xu L (2024) YOLO-fusion and internet of things: advancing object detection in smart transportation. Alex Eng J 107:1–12
Chen H, Guan J (2022) Teacher-student behavior recognition in classroom teaching based on improved YOLO-V4 and internet of things technology. Electronics 11(23):3998
Ragab MG, Abdulkader SJ, Muneer A, Alqushaibi A, Sumiea EH, Qureshi R, Al-Selwi SM, Alhussian H (2024) A comprehensive systematic review of YOLO for medical object detection (2018 to 2023). IEEE Access
Flippo D, Gunturu S, Baldwin C, Badgujar C (2023) Tree trunk detection of eastern red cedar in rangeland environment with deep learning technique. Croatian J For Eng 44(2):357–368. https://doi.org/10.5552/crojfe.2023.2012
Malligere Shivanna V, Guo J-I (2024) Object detection, recognition, and tracking algorithms for ADASs—a study on recent trends. Sensors. https://doi.org/10.3390/s24010249
Flores-Calero M, Astudillo CA, Guevara D, Maza J, Lita BS, Defaz B, Ante JS, Zabala-Blanco D, Armingol Moreno JM (2024) Traffic sign detection and recognition using YOLO object detection algorithm: a systematic review. Mathematics. https://doi.org/10.3390/math12020297
Guerrero-Ibáñez J, Zeadally S, Contreras-Castillo J (2018) Sensor technologies for intelligent transportation systems. Sensors 18(4):1212
Shoman M, Wang D, Aboah A, Abdel-Aty M (2024) Enhancing traffic safety with parallel dense video captioning for end-to-end event analysis. In: Proceedings of the IEEE/CVF Conference on Computer Vision and Pattern Recognition (CVPR) Workshops, pp 7125–7133
Hnewa M, Radha H (2023) Integrated multiscale domain adaptive YOLO. IEEE Trans Image Process 32:1857–1867
Hussain R, Zeadally S (2018) Autonomous cars: Research results, issues, and future challenges. IEEE Commun Surv Tutor 21(2):1275–1313
Shoman M, Lanzaro G, Sayed T, Gargoum S (2024) Autonomous vehicle-pedestrian interaction modeling platform: a case study in four major cities. J Transport Eng Part A: Syst 150(9):04024045. https://doi.org/10.1061/JTEPBS.TEENG-8097
Kaushal M, Khehra BS, Sharma A (2018) Soft computing based object detection and tracking approaches: state-of-the-art survey. Appl Soft Comput 70:423–464
Xiang J, Fan H, Liao H, Xu J, Sun W, Yu S (2014) Moving object detection and shadow removing under changing illumination condition. Math Probl Eng 1:827461
Xiao Y, Jiang A, Ye J, Wang M-W (2020) Making of night vision: object detection under low-illumination. IEEE Access 8:123075–123086
Seoni S, Shahini A, Meiburger KM, Marzola F, Rotunno G, Acharya UR, Molinari F, Salvi M (2024) All you need is data preparation: a systematic review of image harmonization techniques in multi-center/device studies for medical support systems. Comput Methods Programs Biomed 108200
Khan SM, Shah M (2008) Tracking multiple occluding people by localizing on multiple scene planes. IEEE Trans Pattern Anal Mach Intell 31(3):505–519
Mostafa T, Chowdhury SJ, Rhaman MK, Alam MGR (2022) Occluded object detection for autonomous vehicles employing YOLOV5, YOLOX and faster R-CNN. In: 2022 IEEE 13th Annual Information Technology, Electronics and Mobile Communication Conference (IEMCON), pp 0405–0410. IEEE
Gupta A, Anpalagan A, Guan L, Khwaja AS (2021) Deep learning for object detection and scene perception in self-driving cars: survey, challenges, and open issues. Array 10:100057
Zou Z, Chen K, Shi Z, Guo Y, Ye J (2023) Object detection in 20 years: a survey. Proc IEEE 111(3):257–276
Park K, Patten T, Prankl J, Vincze M (2019) Multi-task template matching for object detection, segmentation and pose estimation using depth images. In: 2019 International Conference on Robotics and Automation (ICRA), pp 7207–7213. IEEE
Liu S, Liu D, Srivastava G, Połap D, Woźniak M (2021) Overview and methods of correlation filter algorithms in object tracking. Compl Intell Syst 7:1895–1917
Teutsch M, Kruger W (2015) Robust and fast detection of moving vehicles in aerial videos using sliding windows. In: Proceedings of the IEEE Conference on Computer Vision and Pattern Recognition Workshops, pp 26–34
Lienhart R, Maydt J (2002) An extended set of haar-like features for rapid object detection. In: Proceedings. International Conference on Image Processing, vol. 1, IEEE

Jun-Feng G, Yu-Pin L (2009) A comprehensive study for asymmetric adaboost and its application in object detection. Acta Automatica Sinica 35(11):1403–1409

Li Q, Niaz U, Merialdo B (2012) An improved algorithm on viola-jones object detector. In: 2012 10th International Workshop on Content-Based Multimedia Indexing (CBMI), pp 1–6. IEEE

Hu X-d, Wang X-q, Meng F-j, Hua X, Yan Y-j, Li Y-y, Huang J, Jiang X-l (2020) Gabor-CNN for object detection based on small samples. Def Technol 16(6):1116–1129

Surasak T, Takahiro I, Cheng C-h, Wang C-e, Sheng P-y (2018) Histogram of oriented gradients for human detection in video. In: 2018 5th International Conference on Business and Industrial Research (ICBIR), pp 172–176. IEEE

Karis MS, Razif NRA, Ali NM, Rosli MA, Aras MSM, Ghazaly MM (2016) Local binary pattern (lbp) with application to variant object detection: A survey and method. In: 2016 IEEE 12th International Colloquium on Signal Processing & Its Applications (CSPA), pp. 221–226. IEEE

Mita T, Kaneko T, Hori O (2005) Joint haar-like features for face detection. In: Tenth IEEE International Conference on Computer Vision (ICCV'05) Volume 1, vol 2, pp 1619–1626. IEEE

Yan J, Lei Z, Wen L, Li SZ (2014) The fastest deformable part model for object detection. In: Proceedings of the IEEE Conference on Computer Vision and Pattern Recognition, pp 2497–2504

Piccinini P, Prati A, Cucchiara R (2012) Real-time object detection and localization with sift-based clustering. Image Vis Comput 30(8):573–587

Li J, Zhang Y (2013) Learning surf cascade for fast and accurate object detection. In: Proceedings of the IEEE Conference on Computer Vision and Pattern Recognition, pp 3468–3475

Chiu H-J, Li T-HS, Kuo P-H (2020) Breast cancer-detection system using PCA, multilayer perceptron, transfer learning, and support vector machine. IEEE Access 8:204309–204324

Mienye ID, Sun Y (2022) A survey of ensemble learning: concepts, algorithms, applications, and prospects. IEEE Access 10:99129–99149. https://doi.org/10.1109/ACCESS.2022.3207287

Xiang Y, Mottaghi R, Savarese S (2014) Beyond pascal: a benchmark for 3D object detection in the wild. In: IEEE Winter Conference on Applications of Computer Vision, pp 75–82. IEEE

Krizhevsky A, Sutskever I, Hinton GE (2012) Imagenet classification with deep convolutional neural networks. Adv Neural Inform Process Syst 25

Xie X, Cheng G, Wang J, Yao X, Han J (2021) Oriented R-CNN for object detection. In: Proceedings of the IEEE/CVF International Conference on Computer Vision, pp 3520–3529

Tang S, Yuan Y (2015) Object detection based on convolutional neural network. In: International Conference-IEEE–2016

Zhiqiang W, Jun L (2017) A review of object detection based on convolutional neural network. In: 2017 36th Chinese Control Conference (CCC), pp 11104–11109. IEEE

Li X, Song D, Dong Y (2020) Hierarchical feature fusion network for salient object detection. IEEE Trans Image Process 29:9165–9175

Crawford E, Pineau J (2019) Spatially invariant unsupervised object detection with convolutional neural networks. Proc AAAI Conf Artif Intell 33:3412–3420

Tan M, Pang R, Le QV (2020) Efficientdet: Scalable and efficient object detection. In: Proceedings of the IEEE/CVF Conference on Computer Vision and Pattern Recognition, pp 10781–10790

Carion N, Massa F, Synnaeve G, Usunier N, Kirillov A, Zagoruyko S (2020) End-to-end object detection with transformers. In: European Conference on Computer Vision, pp 213–229. Springer

Li Y, Mao H, Girshick R, He K (2022) Exploring plain vision transformer backbones for object detection. In: European Conference on Computer Vision, pp 280–296. Springer

Girshick R, Donahue J, Darrell T, Malik J, Mercan E (2014) R-CNN for object detection. In: IEEE Conference

Bhat S, Shenoy KA, Jain MR, Manasvi K (2023) Detecting crops and weeds in fields using YOLOv6 and Faster R-CNN object detection models. In: 2023 International Conference on Recent Advances in Information Technology for Sustainable Development (ICRAIS), pp 43–48. IEEE

Girshick R (2015) Fast R-CNN. In: Proceedings of the IEEE International Conference on Computer Vision, pp 1440–1448

Liu W, Anguelov D, Erhan D, Szegedy C, Reed S, Fu C-Y, Berg AC (2016) Ssd: Single shot multibox detector. In: Computer Vision–ECCV 2016: 14th European Conference, Amsterdam, The Netherlands, October 11–14, 2016, Proceedings, Part I 14, pp 21–37. Springer

Redmon J, Divvala S, Girshick R, Farhadi A (2016) You only look once: Unified, real-time object detection. In: Proceedings of the IEEE Conference on Computer Vision and Pattern Recognition, pp 779–788

Tian Y, Ye Q, Doermann D (2025) YOLOv12: Attention-centric real-time object detectors. Preprint at arXiv:2502.12524

Wang C, Sun Q, Dong X, Chen J (2024) Automotive adhesive defect detection based on improved YOLOv8. Signal, Image and Video Processing, pp 1–13

Shoman M, Ghoul T, Lanzaro G, Alsharif T, Gargoum S, Sayed T (2024) Enforcing traffic safety: a deep learning approach for detecting motorcyclists' helmet violations using YOLOv8 and deep convolutional generative adversarial network-generated images. Algorithms. https://doi.org/10.3390/a17050202

Patel GS, Desai AA, Kamble YY, Pujari GV, Chougule PA, Jujare VA (2023) Identification and separation of medicine through robot using YOLO and CNN algorithms for healthcare. In: 2023 International Conference on Artificial Intelligence for Innovations in Healthcare Industries (ICAIIHI), vol 1, pp 1–5. IEEE

Luo Y, Zhang Y, Sun X, Dai H, Chen X, et al (2021) Intelligent solutions in chest abnormality detection based on YOLOv5 and resnet50. J Healthcare Eng 2021

Salinas-Medina A, Neme A (2023) Enhancing hospital efficiency through web-deployed object detection: A YOLOv8-based approach for automating healthcare operations. In: 2023 Mexican International Conference on Computer Science (ENC), pp 1–6. IEEE

Pham D-L, Chang T-W et al (2023) A YOLO-based real-time packaging defect detection system. Proc Comput Sci 217:886–894

Klarák J, Andok R, Malík P, Kuric I, Ritomskỳ M, Klačková I, Tsai H-Y (2024) From anomaly detection to defect classification. Sensors 24(2):429

Arroyo MA, Ziad MTI, Kobayashi H, Yang J, Sethumadhavan S (2019) YOLO: frequently resetting cyber-physical systems for security. In: Autonomous Systems: Sensors, Processing, and Security for Vehicles and Infrastructure 2019, vol 11009, pp 166–183. SPIE

Bordoloi N, Talukdar AK, Sarma KK (2020) Suspicious activity detection from videos using YOLOv3. In: 2020 IEEE 17th India Council International Conference (INDICON), pp 1–5. IEEE

Badgujar CM, Poulose A, Gan H (2024) Agricultural object detection with you look only once (YOLO) algorithm: A bibliometric and systematic literature review. Preprint at arXiv:2401.10379

Li J, Qiao Y, Liu S, Zhang J, Yang Z, Wang M (2022) An improved YOLOv5-based vegetable disease detection method. Comput Electron Agric 202:107345

Fu L, Feng Y, Wu J, Liu Z, Gao F, Majeed Y, Al-Mallahi A, Zhang Q, Li R, Cui Y (2021) Fast and accurate detection of kiwifruit in orchard using improved YOLOv3-tiny model. Precision Agric 22:754–776

Zhong Y, Gao J, Lei Q, Zhou Y (2018) A vision-based counting and recognition system for flying insects in intelligent agriculture. Sensors 18(5):1489

Wang Y, Yang L, Chen H, Hussain A, Ma C, Al-gabri M (2022) Mushroom-YOLO: A deep learning algorithm for mushroom growth recognition based on improved YOLOv5 in agriculture 4.0. In: 2022 IEEE 20th International Conference on Industrial Informatics (INDIN), pp 239–244. IEEE

Jiang K, Xie T, Yan R, Wen X, Li D, Jiang H, Jiang N, Feng L, Duan X, Wang J (2022) An attention mechanism-improved YOLOv7 object detection algorithm for hemp duck count estimation. Agriculture 12(10):1659

Chen G, Hou Y, Cui T, Li H, Shangguan F, Cao L (2023) YOLOv8-CML: A lightweight target detection method for color-changing melon ripening in intelligent agriculture. ResearchSquare

Yu X, Yin D, Xu H, Pinto Espinosa F, Schmidhalter U, Nie C, Bai Y, Sankaran S, Ming B, Cui N, et al (2024) Maize tassel number and tasseling stage monitoring based on near-ground and UAV RGB images by improved YOLOv8. Precis Agric 1–39

Jia L, Wang T, Chen Y, Zang Y, Li X, Shi H, Gao L (2023) MobileNet-CA-YOLO: an improved YOLOv7 based on the MobileNetV3 and attention mechanism for rice pests and diseases detection. Agriculture 13(7):1285

Umar M, Altaf S, Ahmad S, Mahmoud H, Mohamed ASN, Ayub R (2024) Precision agriculture through deep learning: tomato plant multiple diseases recognition with CNN and improved YOLOv7. IEEE Access

Sapkota R, Ahmed D, Karkee M (2024) Comparing YOLOv8 and Mask R-CNN for instance segmentation in complex orchard environments. Artif Intell Agric

Bakirci M, Bayraktar I (2024) Transforming aircraft detection through Leo satellite imagery and YOLOv9 for improved aviation safety. In: 2024 26th International Conference on Digital Signal Processing and Its Applications (DSPA), pp. 1–6. IEEE

Prinzi F, Insalaco M, Orlando A, Gaglio S, Vitabile S (2024) A YOLO-based model for breast cancer detection in mammograms. Cogn Comput 16(1):107–120

Aly GH, Marey M, El-Sayed SA, Tolba MF (2021) YOLO based breast masses detection and classification in full-field digital mammograms. Comput Methods Programs Biomed 200:105823

Ünver HM, Ayan E (2019) Skin lesion segmentation in dermoscopic images with combination of YOLO and grabcut algorithm. Diagnostics 9(3):72

Tan L, Huangfu T, Wu L, Chen W (2021) Comparison of RetinaNet, SSD, and YOLO v3 for real-time pill identification. BMC Med Inform Decis Mak 21:1–11

Suksawatchon U, Srikamdee S, Suksawatchon J, Werapan W (2022) Shape recognition using unconstrained pill images based on deep convolution network. In: 2022 6th International Conference on Information Technology (InCIT), pp 309–313. IEEE

Pratibha K, Mishra M, Ramana G, Lourenço PB (2023) Deep learning-based YOLO network model for detecting surface cracks during structural health monitoring. In: International Conference on Structural Analysis of Historical Constructions, pp 179–187. Springer
Fahim F, Hasan MS (2024) Enhancing the reliability of power grids: A YOLO based approach for insulator defect detection. e-Prime-Advances in Electrical Engineering, Electronics and Energy 9:100663
Gorave A, Misra S, Padir O, Patil A, Ladole K (2020) Suspicious activity detection using live video analysis. In: Proceeding of International Conference on Computational Science and Applications: ICCSA 2019, pp 203–214. Springer
Kolpe R, Ghogare S, Jawale M, William P, Pawar A (2022) Identification of face mask and social distancing using YOLO algorithm based on machine learning approach. In: 2022 6th International Conference on Intelligent Computing and Control Systems (ICICCS), pp 1399–1403. IEEE
Bashir S, Qureshi R, Shah A, Fan X, Alam T (2023) YOLOv5-m: A deep neural network for medical object detection in real-time. In: 2023 IEEE Symposium on Industrial Electronics & Applications (ISIEA), pp 1–6. IEEE
Ajayi OG, Ashi J, Guda B (2023) Performance evaluation of YOLO v5 model for automatic crop and weed classification on UAV images. Smart Agric Technol 5:100231
Morbekar A, Parihar A, Jadhav R (2020) Crop disease detection using YOLO. In: 2020 International Conference for Emerging Technology (INCET), pp 1–5. IEEE
Li D, Ahmed F, Wu N, Sethi AI (2022) YOLO-JD: A deep learning network for jute diseases and pests detection from images. Plants 11(7):937
Cheeti S, Kumar GS, Priyanka JS, Firdous G, Ranjeeva PR (2021) Pest detection and classification using YOLO and CNN. Ann Roman Soc Cell Biol 15295–15300
Pham M-T, Courtrai L, Friguet C, Lefèvre S, Baussard A (2020) YOLO-Fine: One-stage detector of small objects under various backgrounds in remote sensing images. Remote Sens 12(15):2501
Cheng L, Li J, Duan P, Wang M (2021) A small attentional YOLO model for landslide detection from satellite remote sensing images. Landslides 18(8):2751–2765
Chen C, Zheng Z, Xu T, Guo S, Feng S, Yao W, Lan Y (2023) YOLO-based UAV technology: a review of the research and its applications. Drones 7(3):190
Luo X, Wu Y, Zhao L (2022) YOLOD: A target detection method for UAV aerial imagery. Remote Sens 14(14):3240
Li R, Yang J (2018) Improved YOLOv2 object detection model. In: 2018 6th International Conference on Multimedia Computing and Systems (ICMCS), pp 1–6. IEEE
Nakahara H, Yonekawa H, Fujii T, Sato S (2018) A lightweight YOLOv2: A binarized CNN with a parallel support vector regression for an FPGA. In: Proceedings of the 2018 ACM/SIGDA International Symposium on Field-programmable Gate Arrays, pp 31–40
Kim K-J, Kim P-K, Chung Y-S, Choi D-H (2018) Performance enhancement of YOLOv3 by adding prediction layers with spatial pyramid pooling for vehicle detection. In: 2018 15th IEEE International Conference on Advanced Video and Signal Based Surveillance (AVSS), pp 1–6. IEEE
Nepal U, Eslamiat H (2022) Comparing YOLOv3, YOLOv4 and YOLOv5 for autonomous landing spot detection in faulty UAVs. Sensors 22(2):464
Sozzi M, Cantalamessa S, Cogato A, Kayad A, Marinello F (2022) Automatic bunch detection in white grape varieties using YOLOv3, YOLOv4, and YOLOv5 deep learning algorithms. Agronomy 12(2):319
Mohod N, Agrawal P, Madaan V (2022) YOLOv4 vs YOLOv5: Object detection on surveillance videos. In: International Conference on Advanced Network Technologies and Intelligent Computing, pp 654–665. Springer
Ultralytics (2020) https://github.com/ultralytics. Accessed 31 Dec 2024
Li C, Li L, Jiang H, Weng K, Geng Y, Li L, Ke Z, Li Q, Cheng M, Nie W, et al (2022) YOLOv6: A single-stage object detection framework for industrial applications. Preprint at arXiv:2209.02976
Wang C-Y, Bochkovskiy A, Liao H-YM (2022) YOLOv7: Trainable bag-of-freebies sets new state-of-the-art for real-time object detectors. Preprint at arXiv:2207.02696. Accessed 05 Jun 2024
Wang C-Y, Bochkovskiy A, Liao H-YM (2023) YOLOv7: Trainable bag-of-freebies sets new state-of-the-art for real-time object detectors. In: Proceedings of the IEEE/CVF Conference on Computer Vision and Pattern Recognition, pp 7464–7475
Jocher G, Chaurasia A, Qiu J (2023) Ultralytics, YOLOv 8. https://docs.ultralytics.com/models/YOLOv8/. Accessed 31 Dec 2024
Wang C-Y, Yeh I-H, Liao H-YM (2024) YOLOv9: Learning what you want to learn using programmable gradient information. Preprint at arXiv:2402.13616
Ultralytics (2023a) YOLOv 9. https://docs.ultralytics.com/models/YOLOv9/. Accessed 31 Dec 2024
Ultralytics (2023b) YOLOv10: real-time end-to-end object detection. https://docs.ultralytics.com/models/YOLOv10/. Accessed 31 Dec 2024

Wang A, Chen H, Liu L, Chen K, Lin Z, Han J, Ding G (2024) YOLOv10: Real-time end-to-end object detection. Preprint at arXiv:2405.14458

Mao H, Yang X, Dally WJ (2019) A delay metric for video object detection: What average precision fails to tell. In: Proceedings of the IEEE/CVF International Conference on Computer Vision, pp 573–582

Chen B, Ghiasi G, Liu H, Lin TY. Kalenichenko D, Adam H, Le QV (2020) Mnasfpn: Learning latency-aware pyramid architecture for object detection on mobile devices. In: Proceedings of the IEEE/CVF Conference on Computer Vision and Pattern Recognition, pp 13607–13616

Pestana D, Miranda PR, Lopes JD, Duarte RP, Véstias MP, Neto HC, De Sousa JT (2021) A full featured configurable accelerator for object detection with YOLO. IEEE Access 9:75864–75877

Zhou P, Ni B, Geng C, Hu J, Xu Y (2018) Scale-transferrable object detection. In: Proceedings of the IEEE Conference on Computer Vision and Pattern Recognition, pp 528–537

Hall D, Dayoub F, Skinner J, Zhang H, Miller D, Corke P, Carneiro G, Angelova A, Sünderhauf N (2020) Probabilistic object detection: Definition and evaluation. In: Proceedings of the IEEE/CVF Winter Conference on Applications of Computer Vision, pp 1031–1040

Goutte C, Gaussier E (2005) A probabilistic interpretation of precision, recall and f-score, with implication for evaluation. In: European Conference on Information Retrieval, pp 345–359. Springer

Liang Z, Zhang Z, Zhang M, Zhao X, Pu S (2021) Rangeioudet: Range image based real-time 3d object detector optimized by intersection over union. In: Proceedings of the IEEE/CVF Conference on Computer Vision and Pattern Recognition, pp 7140–7149

Jiang J, Xu H, Zhang S, Fang Y (2019) Object detection algorithm based on multiheaded attention. Appl Sci 9(9):1829

Fu C-Y, Liu W, Ranga A, Tyagi A, Berg AC (2017) DSSD: Deconvolutional single shot detector. Preprint at arXiv:1701.06659

Zhang S, Wen L, Bian X, Lei Z, Li SZ (2018) Single-shot refinement neural network for object detection. In: Proceedings of the IEEE Conference on Computer Vision and Pattern Recognition, pp 4203–4212

Cui L, Ma R, Lv P, Jiang X, Gao Z, Zhou B, Xu M (2018) MDSSD: multi-scale deconvolutional single shot detector for small objects. Preprint at arXiv:1805.07009

Lin T, Zhao X, Shou Z (2017) Single shot temporal action detection. In: Proceedings of the 25th ACM International Conference on Multimedia, pp 988–996

Tang X, Du DK, He Z, Liu J (2018) Pyramidbox: a context-assisted single shot face detector. In: Proceedings of the European Conference on Computer Vision (ECCV), pp 797–813

Li Z, Yang L, Zhou F (2017) FSSD: feature fusion single shot multibox detector. Preprint at arXiv:1712.00960

Jiang P, Ergu D, Liu F, Cai Y, Ma B (2022) A review of YOLO algorithm developments. Proc Comput Sci 199:1066–1073

Terven J, Córdova-Esparza D-M, Romero-González J-A (2023) A comprehensive review of YOLO architectures in computer vision: from YOLOv1 to YOLOv8 and YOLO-NAS. Mach Learn Knowl Extract 5(4):1680–1716

Hussain M (2024) YOLOv1 to v8: Unveiling each variant-a comprehensive review of YOLO. IEEE Access 12:42816–42833

Hussain M (2023) YOLO-v1 to YOLO-v8, the rise of YOLO and its complementary nature toward digital manufacturing and industrial defect detection. Machines 11(7):677

Wang C-Y, Liao H-YM, et al (2024) YOLOv1 to YOLOv10: The fastest and most accurate real-time object detection systems. APSIPA Transact Signal Inform Process 13(1)

Jegham N, Koh CY, Abdelatti M, Hendawi A (2024) Evaluating the evolution of YOLO (you only look once) models: A comprehensive benchmark study of YOLO11 and its predecessors. Preprint at arXiv:2411.00201

Wang C-Y, Liao H-YM, Wu Y-H, Chen P-Y, Hsieh J-W, Yeh I-H (2020) CSPNET: A new backbone that can enhance learning capability of CNN. In: Proceedings of the IEEE/CVF Conference on Computer Vision and Pattern Recognition Workshops, pp 390–391

Wang C-Y, Liao H-YM, Yeh I-H (2022) Designing network design strategies through gradient path analysis. Preprint at arXiv:2211.04800

Sapkota R, Meng Z, Karkee M (2024) Synthetic meets authentic: leveraging LLM generated datasets for YOLO11 and YOLOv10-based apple detection through machine vision sensors. Smart Agric Technol 9:100614

Ultralytics (2024) YOLO11 NEW. https://docs.ultralytics.com/models/YOLO11. Accessed 31 Dec 2024

Liu S, Qi L, Qin H, Shi J, Jia J (2018) Path aggregation network for instance segmentation. In: Proceedings of the IEEE Conference on Computer Vision and Pattern Recognition, pp 8759–8768

Rothe R, Guillaumin M, Van Gool L (2015) Non-maximum suppression for object detection by passing messages between windows. In: Computer Vision–ACCV 2014: 12th Asian Conference on Computer Vision, Singapore, Singapore, November 1–5, 2014, Revised Selected Papers, Part I 12, pp 290–306. Springer

Li S, Li M, Li R, He C, Zhang L (2023) One-to-few label assignment for end-to-end dense detection. In: Proceedings of the IEEE/CVF Conference on Computer Vision and Pattern Recognition, pp 7350–7359

Tian Y, Deng N, Xu J, Wen Z (2024) A fine-grained dataset for sewage outfalls objective detection in natural environments. Sci Data 11(1):724

Bhagat S, Kokare M, Haswani V, Hambarde P, Kamble R (2021) Wheatnet-lite: A novel light weight network for wheat head detection. In: Proceedings of the IEEE/CVF International Conference on Computer Vision, pp 1332–1341

Hu Y, Tan W, Meng F, Liang Y (2023) A decoupled spatial-channel inverted bottleneck for image compression. In: 2023 IEEE International Conference on Image Processing (ICIP), pp 1740–1744. IEEE

Yang G, Wang J, Nie Z, Yang H, Yu S (2023) A lightweight YOLOv8 tomato detection algorithm combining feature enhancement and attention. Agronomy 13(7):1824

Lin T-Y, Maire M, Belongie S, Hays J, Perona P, Ramanan D, Dollár P, Zitnick CL (2014) Microsoft coco: Common objects in context. In: Computer Vision–ECCV 2014: 13th European Conference, Zurich, Switzerland, September 6-12, 2014, Proceedings, Part V 13, pp 740–755. Springer

Jocher G, et al (2022) YOLOv8: a comprehensive improvement of the YOLO object detection series. https://docs.ultralytics.com/YOLOv8/. Accessed 05 Jun 2024

Tishby N, Zaslavsky N (2015) Deep learning and the information bottleneck principle. In: 2015 IEEE Information Theory Workshop (ITW), pp. 1–5. IEEE

Zhang B, Li J, Bai Y, Jiang Q, Yan B, Wang Z (2023) An improved microaneurysm detection model based on SWINIR and YOLOv8. Bioengineering 10(12):1405

Chien C-T, Ju R-Y, Chou K-Y, Chiang J-S (2024) YOLOv9 for fracture detection in pediatric wrist trauma X-ray images. Preprint at arXiv:2403.11249

Ultralytics: Home—docs.ultralytics.com. https://docs.ultralytics.com/. Accessed 28 May 2024

Ultralytics: YOLOv8 object detection model: what is, how to use—roboflow.com. https://roboflow.com/model/YOLOv8. Accessed 28 May 2024

Ultralytics: ultralytics YOLOv8 solutions: quick walkthrough—ultralytics.medium.com. https://ultralytics.medium.com/ultralytics-YOLOv8-solutions-quick-walkthrough-b802fd6da5d7. Accessed 28 May 2024

Du S, Zhang B, Zhang P, Xiang P (2021) An improved bounding box regression loss function based on CIOU loss for multi-scale object detection. In: 2021 IEEE 2nd International Conference on Pattern Recognition and Machine Learning (PRML), pp 92–98. IEEE

Xu S, Wang X, Lv W, Chang Q, Cui C, Deng K, Wang G, Dang Q, Wei S, Du Y, et al (2022) PP-YOLOE: An evolved version of YOLO. Preprint at arXiv:2203.16250

Yue X, Qi K, Na X, Zhang Y, Liu Y, Liu C (2023) Improved YOLOv8-seg network for instance segmentation of healthy and diseased tomato plants in the growth stage. Agriculture 13(8):1643

Zhu X, Lyu S, Wang X, Zhao Q (2021) TPH-YOLOv5: Improved YOLOv5 based on transformer prediction head for object detection on drone-captured scenarios. In: Proceedings of the IEEE/CVF International Conference on Computer Vision, pp 2778–2788

Woo S, Park J, Lee J-Y, Kweon IS (2018) CBAM: Convolutional block attention module. In: Proceedings of the European Conference on Computer Vision (ECCV), pp 3–19

Bai Z, Pei X, Qiao Z, Wu G, Bai Y (2024) Improved YOLOv7 target detection algorithm based on UAV aerial photography. Drones 8(3):104

Sirisha U, Praveen SP, Srinivasu PN, Barsocchi P, Bhoi AK (2023) Statistical analysis of design aspects of various YOLO-based deep learning models for object detection. Int J Comput Intell Syst 16(1):126

Wang K, Liew JH, Zou Y, Zhou D, Feng J (2019) PANET: Few-shot image semantic segmentation with prototype alignment. In: Proceedings of the IEEE/CVF International Conference on Computer Vision, pp 9197–9206

Zhang Z, Lu X, Cao G, Yang Y, Jiao L, Liu F (2021) VIT-YOLO: Transformer-based YOLO for object detection. In: Proceedings of the IEEE/CVF International Conference on Computer Vision, pp 2799–2808

Ultralytics: Comprehensive Guide to Ultralytics YOLOv5—docs.ultralytics.com. https://docs.ultralytics.com/YOLOv5/. Accessed 28 May 2024

Ultralytics: GitHub-ultralytics/YOLOv5: YOLOv5 in PyTorch [CDATA[>]]> ONNX >[CDATA[>]] CoreML >[CDATA[>]] TFLite—github.com.https://github.com/ultralytics/YOLOv5. Accessed 28 May 2024

Bochkovskiy A, Wang C-Y, Liao H-YM (2020) YOLOv4: Optimal speed and accuracy of object detection. Preprint at arXiv:2004.10934

Mahasin M, Dewi IA (2022) Comparison of CSPDarkNet53, CSPResNeXt-50, and EfficientNet-B0 backbones on YOLO V4 as object detector. Int J Eng Sci Inform Technol 2(3):64–72

Redmon J, Farhadi A (2018) YOLOv3: An incremental improvement. Preprint at arXiv:1804.02767

Misra D (2019) Mish: A self regularized non-monotonic activation function. Preprint at arXiv:1908.08681

Yun S, Han D, Oh SJ, Chun S, Choe J, Yoo Y (2019) Cutmix: Regularization strategy to train strong classifiers with localizable features. In: Proceedings of the IEEE/CVF International Conference on Computer Vision, pp 6023–6032
Ghiasi G, Lin T-Y, Le QV (2018) Dropblock: a regularization method for convolutional networks. Adv Neural Inform Process Syst 31
Müller R, Kornblith S, Hinton GE (2019) When does label smoothing help? Adv Neural Inform Process Syst 32
Zhang Z, He T, Zhang H, Zhang Z, Xie J, Li M (2019) Bag of freebies for training object detection neural networks. Preprint at arXiv:1902.04103
Redmon J, Farhadi A (2017) YOLO9000: better, faster, stronger. In: Proceedings of the IEEE Conference on Computer Vision and Pattern Recognition, pp 7263–7271
Redmon J, Divvala S, Girshick R, Farhadi A (2016) You only look once: unified, real-time object detection. Preprint at arXiv:1506.02640. Accessed 05 Jun 2024
Ren P, Xiao Y, Chang X, Huang P-Y, Li Z, Chen X, Wang X (2021) A comprehensive survey of neural architecture search: challenges and solutions. ACM Comput Surv (CSUR) 54(4):1–34
Mithun M, Jawhar SJ (2024) Detection and classification on MRI images of brain tumor using YOLO NAS deep learning model. J Radiat Res Appl Sci 17(4):101113
Ge Z (2021) YOLOx: Exceeding YOLO series in 2021. Preprint at arXiv:2107.08430
Zhang Y, Zhang W, Yu J, He L, Chen J, He Y (2022) Complete and accurate holly fruits counting using YOLOx object detection. Comput Electron Agric 198:107062
Liu J, Sun W (2022) YOLOx-based ship target detection for shore-based monitoring. In: Proceedings of the 2022 5th International Conference on Signal Processing and Machine Learning, pp 234–241
Ashraf I, Hur S, Kim G, Park Y (2024) Analyzing performance of YOLOx for detecting vehicles in bad weather conditions. Sensors 24(2):522
Chang H-S, Wang C-Y, Wang RR, Chou G, Liao H-YM (2023) YOLOr-based multi-task learning. Preprint at arXiv:2309.16921
Andrei-Alexandru T, Cosmin C, Ioan S, Adrian-Alexandru T, Henrietta DE (2022) Novel ceramic plate defect detection using YOLO-r. In: 2022 14th International Conference on Electronics, Computers and Artificial Intelligence (ECAI), pp 1–6. IEEE
Sun H, Lu D, Li X, Tan J, Zhao J, Hou D (2024) Research on multi-apparent defects detection of concrete bridges based on YOLOr. Structures 65:106735
Xu X, Jiang Y, Chen W, Huang Y, Zhang Y, Sun X (2022) Damo-YOLO: a report on real-time object detection design. Preprint at arXiv:2211.15444
Wang C, He W, Nie Y, Guo J, Liu C, Wang Y, Han K (2024) Gold-YOLO: Efficient object detector via gather-and-distribute mechanism. Adv Neural Inform Process Syst 36
Vijayakumar A, Vairavasundaram S (2024) YOLO-based object detection models: a review and its applications. Multim Tools Appl 1–40
Alibabaei K, Gaspar PD, Lima TM, Campos RM, Girão I, Monteiro J, Lopes CM (2022) A review of the challenges of using deep learning algorithms to support decision-making in agricultural activities. Remote Sens 14(3):638
Wang C, Liu B, Liu L, Zhu Y, Hou J, Liu P, Li X (2021) A review of deep learning used in the hyperspectral image analysis for agriculture. Artif Intell Rev 54(7):5205–5253
Benjumea A, Teeti I, Cuzzolin F, Bradley A (2021) YOLO-z: Improving small object detection in YOLOv5 for autonomous vehicles. Preprint at arXiv:2112.11798
Sarda A, Dixit S, Bhan A (2021) Object detection for autonomous driving using YOLO [you only look once] algorithm. In: 2021 Third International Conference on Intelligent Communication Technologies and Virtual Mobile Networks (ICICV), pp 1370–1374. IEEE
Cai Y, Luan T, Gao H, Wang H, Chen L, Li Y, Sotelo MA, Li Z (2021) YOLOv4-5D: An effective and efficient object detector for autonomous driving. IEEE Trans Instrum Meas 70:1–13
Zhao J, Hao S, Dai C, Zhang H, Zhao L, Ji Z, Ganchev I (2022) Improved vision-based vehicle detection and classification by optimized YOLOv4. IEEE Access 10:8590–8603
Woo J, Baek J-H, Jo S-H, Kim SY, Jeong J-H (2022) A study on object detection performance of YOLOv4 for autonomous driving of tram. Sensors 22(22):9026
Ye C, Wang Y, Wang Y, Tie M (2022) Steering angle prediction YOLOv5-based end-to-end adaptive neural network control for autonomous vehicles. Proc Institut Mech Eng Part D 236(9):1991–2011
Jia X, Tong Y, Qiao H, Li M, Tong J, Liang B (2023) Fast and accurate object detector for autonomous driving based on improved YOLOv5. Sci Rep 13(1):9711
Chen Z, Wang X, Zhang W, Yao G, Li D, Zeng L (2023) Autonomous parking space detection for electric vehicles based on improved YOLOv5-OBB algorithm. World Electric Vehicle J 14(10):276
Liu X, Yan WQ (2022) Vehicle-related distance estimation using customized YOLOv7. In: International Conference on Image and Vision Computing New Zealand, pp 91–103. Springer

Mehla N, Ishita Talukdar R, Sharma DK (2023) Object detection in autonomous maritime vehicles: Comparison between YOLO v8 and efficientdet. In: International Conference on Data Science and Network Engineering, pp 125–141. Springer
Kumar D, Muhammad N (2023) Object detection in adverse weather for autonomous driving through data merging and YOLOv8. Sensors 23(20):8471
Oh G, Lim S (2023) One-stage brake light status detection based on YOLOv8. Sensors 23(17):7436
Afdhal A, Saddami K, Sugiarto S, Fuadi Z, Nasaruddin N (2023) Real-time object detection performance of YOLOv8 models for self-driving cars in a mixed traffic environment. In: 2023 2nd International Conference on Computer System, Information Technology, and Electrical Engineering (COSITE), pp. 260–265. IEEE
Wang H, Liu C, Cai Y, Chen L, Li Y (2024) YOLOv8-QSD: An improved small object detection algorithm for autonomous vehicles based on YOLOv8. IEEE Transactions on Instrumentation and Measurement
Bakirci M, Dmytrovych P, Bayraktar I, Anatoliyovych O (2024) Multi-class vehicle detection and classification with YOLO11 on uav-captured aerial imagery. In: 2024 IEEE 7th International Conference on Actual Problems of Unmanned Aerial Vehicles Development (APUAVD), pp 191–196. IEEE
Li Y, Leong W, Zhang H (2024) YOLOv10-based real-time pedestrian detection for autonomous vehicles. In: 2024 IEEE 8th International Conference on Signal and Image Processing Applications (ICSIPA), pp 1–6. IEEE
Arifando R, Eto S, Tibyani T, Wada C (2025) Improved YOLOv10 for visually impaired: balancing model accuracy and efficiency in the case of public transportation. Informatics 12:7
Wibowo A, Trilaksono BR, Hidayat EMI, Munir R (2023) Object detection in dense and mixed traffic for autonomous vehicles with modified YOLO. IEEE Access 11:134866–134877
Hung WCW, Zakaria MA, Ishak M, Heerwan P (2022) Object tracking for autonomous vehicle using YOLO v3. In: Enabling Industry 4.0 Through Advances in Mechatronics: Selected Articles from iM3F 2021, Malaysia, pp 265–273. Springer
Zaghari N, Fathy M, Jameii SM, Shahverdy M (2021) The improvement in obstacle detection in autonomous vehicles using YOLO non-maximum suppression fuzzy algorithm. J Supercomput 77(11):13421–13446
Ali Y, Haque MM, Mannering F (2023) A Bayesian generalised extreme value model to estimate real-time pedestrian crash risks at signalised intersections using artificial intelligence-based video analytics. Anal Methods Accident Res 38:100264
Hussain F, Ali Y, Li Y, Haque MM (2024) Revisiting the hybrid approach of anomaly detection and extreme value theory for estimating pedestrian crashes using traffic conflicts obtained from artificial intelligence-based video analytics. Accident Analysis & Prevention 199:107517
Ghaziamin P, Bajaj K, Bouguila N, Patterson Z (2024) A privacy-preserving edge computing solution for real-time passenger counting at bus stops using overhead fisheye camera. In: 2024 IEEE 18th International Conference on Semantic Computing (ICSC), pp 25–32. IEEE
Zhang S, Abdel-Aty M, Yuan J, Li P (2020) Prediction of pedestrian crossing intentions at intersections based on long short-term memory recurrent neural network. Transp Res Rec 2674(4):57–65
Yang HF, Ling Y, Kopca C, Ricord S, Wang Y (2022) Cooperative traffic signal assistance system for non-motorized users and disabilities empowered by computer vision and edge artificial intelligence. Trans Res Part C: Emerg Technol 145:103896
Jiao D, Fei T (2023) Pedestrian walking speed monitoring at street scale by an in-flight drone. Peer J Comput Sci 9:1226
Wang Y, Jia Y, Chen W, Wang T, Zhang A (2024) Examining safe spaces for pedestrians and e-bicyclists at urban crosswalks: an analysis based on drone-captured video. Accident Anal Prevent 194:107365
Zhou W, Liu Y, Zhao L, Xu S, Wang C (2023) Pedestrian crossing intention prediction from surveillance videos for over-the-horizon safety warning. IEEE Transactions on Intelligent Transportation Systems
Xiao X, Feng X (2023) Multi-object pedestrian tracking using improved YOLOv8 and oc-sort. Sensors, 23(20) https://doi.org/10.3390/s23208439
Li S, Wang S, Wang P (2023) A small object detection algorithm for traffic signs based on improved YOLOv7. Sensors. https://doi.org/10.3390/s23167145
Mahaur B, Mishra KK (2023) Small-object detection based on YOLOv5 in autonomous driving systems. Pattern Recogn Lett 168:115–122. https://doi.org/10.1016/j.patrec.2023.03.009
Zhang H, Qin L, Li J, Guo Y, Zhou Y, Zhang J, Xu Z (2020) Real-time detection method for small traffic signs based on YOLOv3. IEEE Access 8:64145–64156. https://doi.org/10.1109/ACCESS.2020.2984554
Li M, Zhang L, Li L, W S, (2022) YOLO-based traffic sign recognition algorithm. Comput Intell Neurosci. https://doi.org/10.1155/2022/2682921
Zhang J, Huang M, Jin X, Li X (2017) A real-time Chinese traffic sign detection algorithm based on modified YOLOv2. Algorithms https://doi.org/10.3390/a10040127
Bai W, Zhao J, Dai C, Zhang H, Zhao L, Ji Z, Ganchev I (2023) Two novel models for traffic sign detection based on YOLOv5s. Axioms. https://doi.org/10.3390/axioms12020160

Soylu E, Soylu T (2024) A performance comparison of YOLOv8 models for traffic sign detection in the RobotAXI-full scale autonomous vehicle competition. Multim Tools Appl 83(8):25005–25035

Zhang M (2024) Research on traffic sign detection based on improved YOLOv9. In: Kolivand, H., Moshayedi, A.J. (eds.) Fourth International Conference on Computer Graphics, Image, and Virtualization (ICCGIV 2024), vol 13288, p 132880. SPIE. https://doi.org/10.1117/12.3045722. International Society for Optics and Photonics

Karaca H, Atasoy NA (2025) Fine-grained classification of military aircraft using pre-trained deep learning models and YOLO11. Curr Trends Comput 2(2):150–171

Kaur R, Singh J (2022) Local regression based real-time traffic sign detection using YOLOv6. In: 2022 4th International Conference on Advances in Computing, Communication Control and Networking (ICAC3N), pp 522–526. IEEE

Dewi C, Chen R-C, Jiang X, Yu H (2022) Deep convolutional neural network for enhancing traffic sign recognition developed on YOLO v4. Multim Tools Appl 81(26):37821–37845

Pandey S, Chen K-F, Dam EB (2023) Comprehensive multimodal segmentation in medical imaging: Combining YOLOv8 with SAM and HQ-SAM models. In: Proceedings of the IEEE/CVF International Conference on Computer Vision, pp 2592–2598

Ju R-Y, Cai W (2023) Fracture detection in pediatric wrist trauma x-ray images using YOLOv8 algorithm. Sci Rep 13(1):20077

Inui A, Mifune Y, Nishimoto H, Mukohara S, Fukuda S, Kato T, Furukawa T, Tanaka S, Kusunose M, Takigami S et al (2023) Detection of elbow OCD in the ultrasound image by artificial intelligence using YOLOv8. Appl Sci 13(13):7623

Wu B, Pang C, Zeng X, Hu X (2022) ME-YOLO: Improved YOLOv5 for detecting medical personal protective equipment. Appl Sci 12(23):11978

Zhao X, Wang Q, Zhang M, Wei Z, Ku R, Zhang Z, Yu Y, Zhang B, Liu Y, Wang C (2024) CSFF-YOLOv5: Improved YOLOv5 based on channel split and feature fusion in femoral neck fracture detection. Internet Things 26:101190

Goel L, Patel P (2024) Improving YOLOv6 using advanced PSO optimizer for weight selection in lung cancer detection and classification. Multim Tools Appl 1–34

Norkobil Saydirasulovich S, Abdusalomov A, Jamil MK, Nasimov R, Kozhamzharova D, Cho Y-I (2023) A YOLOv6-based improved fire detection approach for smart city environments. Sensors 23(6):3161

Zou J, Arshad MR (2024) Detection of whole body bone fractures based on improved YOLOv7. Biomed Signal Process Control 91:105995

Razaghi M, Komleh HE, Dehghani F, Shahidi Z (2024) Innovative diagnosis of dental diseases using YOLO v8 deep learning model. In: 2024 13th Iranian/3rd International Machine Vision and Image Processing Conference (MVIP), pp 1–5. IEEE

Pham T-L, Le V-H (2024) Ovarian tumors detection and classification from ultrasound images based on YOLOv8. J Adv Inform Technol 15(2)

Krishnamurthy V, Balasubramanian S, Kanmani RS, Srividhya S, Deepika J, Nimeshika GN (2023) Endoscopic surgical operation and object detection using custom architecture models. In: International Conference on Human-Centric Smart Computing, pp 637–654. Springer

Palanivel N, Deivanai S, Sindhuja B, et al (2023) The art of YOLOv8 algorithm in cancer diagnosis using medical imaging. In: 2023 International Conference on System, Computation, Automation and Networking (ICSCAN), pp 1–6. IEEE

Karaköse M, Yetış H, Çeçen M (2024) A new approach for effective medical deepfake detection in medical images. IEEE Access

Bhojane R, Chourasia S, Laddha SV, Ochawar RS (2023) Liver lesion detection from mr t1 in-phase and out-phase fused images and CT images using YOLOv8. In: International Conference on Data Science and Applications, pp 121–135. Springer

Akdoğan S, Öziç MÜ, Tassoker M (2025) Development of an ai-supported clinical tool for assessing mandibular third molar tooth extraction difficulty using panoramic radiographs and YOLO11 sub-models. Diagnostics 15(4):462

Ali BS, Nasir H, Khan A, Ashraf M, Akbar SM (2024) A machine learning-based model for the detection of skin cancer using YOLOv10. In: 2024 IEEE 8th International Conference on Signal and Image Processing Applications (ICSIPA), pp 1–6. IEEE

Julia R, Prince S, Bini D (2024) Medical image analysis of masses in mammography using deep learning model for early diagnosis of cancer tissues. In: Computational Intelligence and Modelling Techniques for Disease Detection in Mammogram Images, pp 75–89. Elsevier

Salahin SS, Ullaa MS, Ahmed S, Mohammed N, Farook TH, Dudley J (2023) One-stage methods of computer vision object detection to classify carious lesions from smartphone imaging. Oral 3(2):176–190

Doniyorjon M, Madinakhon R, Shakhnoza M, Cho Y-I (2022) An improved method of polyp detection using custom YOLOv4-tiny. Appl Sci 12(21):10856

Ding B, Zhang Z, Liang Y, Wang W, Hao S, Meng Z, Guan L, Hu Y, Guo B, Zhao R, et al (2021) Detection of dental caries in oral photographs taken by mobile phones based on the YOLOv3 algorithm. Ann Transl Med 9(21)

Wang L, Yang S, Yang S, Zhao C, Tian G, Gao Y, Chen Y, Lu Y (2019) Automatic thyroid nodule recognition and diagnosis in ultrasound imaging with the YOLOv2 neural network. World J Surg Oncol 17:1–9

Kwaśniewska A, Rumiński J, Czuszyński K, Szankin M (2018) Real-time facial features detection from low resolution thermal images with deep classification models. J Med Imaging Health Inform 8(5):979–987

Majeed F, Khan FZ, Nazir M, Iqbal Z, Alhaisoni M, Tariq U, Khan MA, Kadry S (2022) Investigating the efficiency of deep learning based security system in a real-time environment using YOLOv5. Sustain Energy Technol Assess 53:102603

Aboah A, Shoman M, Mandal V, Davami S, Adu-Gyamfi Y, Sharma A (2021) A vision-based system for traffic anomaly detection using deep learning and decision trees. In: 2021 IEEE/CVF Conference on Computer Vision and Pattern Recognition Workshops (CVPRW), pp 4202–4207. https://doi.org/10.1109/CVPRW53098.2021.00475

AFFES N, KTARI J, BEN AMOR N, FRIKHA T, HAMAM H (2023) Comparison of YOLOV5, YOLOV6, YOLOV7 and YOLOV8 for intelligent video surveillance. J Inform Assur Secur 18(5)

Cao F, Ma S (2023) Enhanced campus security target detection using a refined YOLOv7 approach. Traitement du Signal 40(5)

Chatterjee N, Singh AV, Agarwal R (2024) You only look once (YOLOv8) based intrusion detection system for physical security and surveillance. In: 2024 11th International Conference on Reliability, Infocom Technologies and Optimization (Trends and Future Directions)(ICRITO), pp 1–5. IEEE

Sandhya, Kashyap A (2024) Real-time object-removal tampering localization in surveillance videos by employing YOLO-v8. J For Sci

Tran DQ, Aboah A, Jeon Y, Shoman M, Park M, Park S (2024) Low-light image enhancement framework for improved object detection in fisheye lens datasets. In: Proceedings of the IEEE/CVF Conference on Computer Vision and Pattern Recognition (CVPR) Workshops, pp 7056–7065

Bakirci M, Bayraktar I (2024) Boosting aircraft monitoring and security through ground surveillance optimization with YOLOv9. In: 2024 12th International Symposium on Digital Forensics and Security (ISDFS), pp 1–6. IEEE

Bakirci M, Bayraktar I (2024) YOLOv9-enabled vehicle detection for urban security and forensics applications. In: 2024 12th International Symposium on Digital Forensics and Security (ISDFS), pp 1–6. IEEE

Chakraborty S, Zahir S, Orchi NT, Hafiz MFB, Shamsuddoha A, Dipto SM (2024) Violence detection: A multi-model approach towards automated video surveillance and public safety. In: 2024 International Conference on Advances in Computing, Communication, Electrical, and Smart Systems (iCACCESS), pp 1–6. IEEE

Chen G, Du W, Xu T, Wang S, Qi X, Wang Y (2024) Investigating enhanced YOLOv8 model applications for large-scale security risk management and drone-based low-altitude law enforcement. Highlights Sci Eng Technol 98:390–396

Pashayev F, Babayeva L, Isgandarova Z, Kalejahi BK (2023) Face recognition in smart cameras by YOLO8. Khazar J Sci Technol (KJSAT) 67

Kaç SB, Eken S, Balta DD, Balta M, İskefiyeli M, Özçelik İ (2024) Image-based security techniques for water critical infrastructure surveillance. Appl Soft Comput 111730

Gao Q, Deng H, Zhang G (2024) A contraband detection scheme in X-ray security images based on improved YOLOv8s network model. Sensors 24(4):1158

Antony JC, Chowdary CLS, Murali E, Mayan A, et al (2024) Advancing crowd management through innovative surveillance using YOLOv8 and bytetrack. In: 2024 International Conference on Wireless Communications Signal Processing and Networking (WiSPNET), pp 1–6. IEEE

Zhang D (2024) A YOLO-based approach for fire and smoke detection in IoT surveillance systems. Int J Adv Comput Sci Appl 15(1)

Khin PP, Htaik NM (2024) Gun detection: A comparative study of retinanet, efficientdet and YOLOv8 on custom dataset. In: 2024 IEEE Conference on Computer Applications (ICCA), pp 1–7. IEEE

Nkuzo L, Sibiya M, Markus ED (2023) A comprehensive analysis of real-time car safety belt detection using the YOLOv7 algorithm. Algorithms 16(9):400

Chang R, Zhang B, Zhu Q, Zhao S, Yan K, Yang Y, et al (2023) FFA-YOLOv7: Improved YOLOv7 based on feature fusion and attention mechanism for wearing violation detection in substation construction safety. J Electrical Comput Eng 2023

Bakirci M, Bayraktar I (2024) Assessment of YOLO11 for ship detection in sar imagery under open ocean and coastal challenges. In: 2024 21st International Conference on Electrical Engineering, Computing Science and Automatic Control (CCE), pp 1–6. IEEE

Žigulić N, Glučina M, Frank D, Lorencin I, Šverko Z, Matika D (2024) Deploying YOLOv10 for affordable real-time handgun detection. In: 2024 IEEE 22nd Jubilee International Symposium on Intelligent Systems and Informatics (SISY), pp 000283–000288. IEEE

Han L, Ma C, Liu Y, Jia J, Sun J (2023) SC-YOLOv8: a security check model for the inspection of prohibited items in X-ray images. Electronics 12(20):4208

Yuan J, Zhang N, Xie Y, Gao X (2022) Detection of prohibited items based upon x-ray images and improved YOLOv7. In: Journal of Physics: Conference Series, vol. 2390. Guangzhou, China, p. 012114. https://doi.org/10.1088/1742-6596/2390/1/012114. 3rd International Conference on Advanced Materials and Intelligent Manufacturing (ICAMIM 2022). https://iopscience.iop.org/article/10.1088/1742-6596/2390/1/012114

Awang S, Rokei MQR, Sulaiman J (2023) Suspicious activity trigger system using YOLOv6 convolutional neural network. In: 2023 International Conference on Artificial Intelligence in Information and Communication (ICAIIC), pp 527–532. IEEE

Xiao Y, Chang A, Wang Y, Huang Y, Yu J, Huo L (2022) Real-time object detection for substation security early-warning with deep neural network based on YOLO-v5. In: 2022 IEEE IAS Global Conference on Emerging Technologies (GlobConET), pp 45–50. IEEE

Wang G, Ding H, Duan M, Pu Y, Yang Z, Li H (2023) Fighting against terrorism: a real-time CCTV autonomous weapons detection based on improved YOLO v4. Digit Signal Process 132:103790

Kashika P, Venkatapur RB (2022) Automatic tracking of objects using improvised YOLOv3 algorithm and alarm human activities in case of anomalies. Int J Inf Technol 14(6):2885–2891

Mohandoss T, Rangaraj J (2024) Multi-object detection using enhanced YOLOv2 and Lunet algorithms in surveillance videos. e-Prime-Adv Electrical Eng Electron Energy 8:100535

Smiley CJ (2015) From silence to propagation: understanding the relationship between “stop snitchin” and “YOLO”. Deviant Behav 36(1):1–16

Pendse R, Rajput H, Saraf S, Sarwate A, Jadhav J et al (2023) Defect detection in manufacturing using YOLOv7. IJRAR-Int J Res Anal Rev (IJRAR) 10(2):179–185

Yi F, Zhang H, Yang J, He L, Mohamed ASA, Gao S (2024) YOLOv7-SIAMFF: Industrial defect detection algorithm based on improved YOLOv7. Comput Electr Eng 114:109090

Wang H, Xu X, Liu Y, Lu D, Liang B, Tang Y (2023) Real-time defect detection for metal components: a fusion of enhanced Canny–Devernay and YOLOv6 algorithms. Appl Sci 13(12):6898

Ludwika AS, Rifai AP (2024) Deep learning for detection of proper utilization and adequacy of personal protective equipment in manufacturing teaching laboratories. Safety 10(1):26

Beak S, Han Y-H, Moon Y, Lee J, Jeong J (2023) YOLOv7-based anomaly detection using intensity and ng types in labeling in cosmetic manufacturing processes. Processes 11(8):2266

Zhao H, Wang X, Sun J, Wang Y, Chen Z, Wang J, Xu X (2024) Artificial intelligence powered real-time quality monitoring for additive manufacturing in construction. Constr Build Mater 429:135894

Liu Z, Ye K (2023) YOLO-IMF: an improved YOLOv8 algorithm for surface defect detection in industrial manufacturing field. In: International Conference on Metaverse, pp 15–28. Springer

Wen Y, Wang L (2024) YOLO-SD: simulated feature fusion for few-shot industrial defect detection based on YOLOv8 and stable diffusion. Int J Mach Learn Cybernet 1–13

Karna N, Putra MAP, Rachmawati SM, Abisado M, Sampedro GA (2023) Towards accurate fused deposition modeling 3d printer fault detection using improved YOLOv8 with hyperparameter optimization. IEEE Access

Li W, Solihin MI, Nugroho HA (2024) Rca: YOLOv8-based surface defects detection on the inner wall of cylindrical high-precision parts. Arab J Sci Eng 1–19

Hu Y, Wang J, Wang X, Sun Y, Yu H, Zhang J (2024) Real-time evaluation of the blending uniformity of industrially produced gravelly soil based on Cond-YOLOv8-seg. J Ind Inf Integr 39:100603

Yang S, Zhang Z, Wang B, Wu J (2024) Dcs-YOLOv8: An improved steel surface defect detection algorithm based on YOLOv8. In: Proceedings of the 2024 7th International Conference on Image and Graphics Processing, pp 39–46

Wang X, Gao H, Jia Z, Li Z (2023) BL-YOLOv8: An improved road defect detection model based on YOLOv8. Sensors 23(20):8361

Luo B, Kou Z, Han C, Wu J (2023) A “hardware-friendly” foreign object identification method for belt conveyors based on improved YOLOv8. Appl Sci 13(20):11464

Wu Q, Kuang X, Tang X, Guo D, Luo Z (2023) Industrial equipment object detection based on improved YOLOv7. In: International Conference on Computer, Artificial Intelligence, and Control Engineering (CAICE 2023), vol 12645, pp 600–608. SPIE

Kim O, Han Y, Jeong J (2022) Real-time inspection system based on Moire pattern and YOLOv7 for coated high-reflective injection molding product. WSEAS Trans Comput Res 10:120–125

Chen J, Bai S, Wan G, Li Y (2023) Research on YOLOv7-based defect detection method for automotive running lights. Syst Sci Control Eng 11(1):2185916

Hussain M, Al-Aqrabi H, Munawar M, Hill R, Alsboui T (2022) Domain feature mapping with YOLOv7 for automated edge-based pallet racking inspections. Sensors 22(18):6927
Zhu B, Xiao G, Zhang Y, Gao H (2023) Multi-classification recognition and quantitative characterization of surface defects in belt grinding based on YOLOv7. Measurement 216:112937
Banduka N, Tomić K, Živadinović J, Mladineo M (2024) Automated dual-side leather defect detection and classification using YOLOv11: a case study in the finished leather industry. Processes 12(12):2892
Liao L, Song C, Wu S, Fu J (2025) A novel YOLOv10-based algorithm for accurate steel surface defect detection. Sensors 25(3):769
Huang G, Huang Y, Li H, Guan Z, Li X, Zhang G, Li W, Zheng X (2024) An improved YOLOv9 and its applications for detecting flexible circuit boards connectors. Int J Comput Intell Syst 17(1):261
Gupta C, Gill NS, Gulia P, Chatterjee JM (2023) A novel finetuned YOLOv6 transfer learning model for real-time object detection. J Real-Time Image Proc 20(3):42
Zendehdel N, Chen H, Leu MC (2023) Real-time tool detection in smart manufacturing using you-only-look-once (YOLO) v5. Manuf Lett 35:1052–1059
Jiang D, Wang H, Lu Y (2024) An efficient automobile assembly state monitoring system based on channel-pruned YOLOv4 algorithm. Int J Comput Integr Manuf 37(3):372–382
Yan J, Wang Z (2022) YOLO v3+ vgg16-based automatic operations monitoring and analysis in a manufacturing workshop under industry 4.0. J Manuf Syst 63:134–142
Arima K, Nagata F, Shimizu T, Otsuka A, Kato H, Watanabe K, Habib MK (2023) Improvements of detection accuracy and its confidence of defective areas by YOLOv2 using a data set augmentation method. Artif Life Robot 28(3):625–631
Badgujar CM, Armstrong PR, Gerken AR, Pordesimo LO, Campbell JF (2023) Real-time stored product insect detection and identification using deep learning: System integration and extensibility to mobile platforms. J Stored Products Res 104:102196. https://doi.org/10.1016/j.jspr.2023.102196. Accessed 05 Jun 2024
Shoman M, Aboah A, Morehead A, Duan Y, Daud A, Adu-Gyamfi Y (2022) A region-based deep learning approach to automated retail checkout. In: Proceedings of the IEEE/CVF Conference on Computer Vision and Pattern Recognition (CVPR) Workshops, pp 3210–3215
Bist RB, Subedi S, Yang X, Chai L (2023) A novel YOLOv6 object detector for monitoring piling behavior of cage-free laying hens. AgriEngineering 5(2):905–923
Kumar P, Kumar N (2023) Drone-based apple detection: Finding the depth of apples using YOLOv7 architecture with multi-head attention mechanism. Smart Agric Technol 5:100311
Zhang L, Ding G, Li C, Li D (2023) DCF-YOLOv8: An improved algorithm for aggregating low-level features to detect agricultural pests and diseases. Agronomy 13(8):2012
Sharma A, Kumar V, Longchamps L (2024) Comparative performance of YOLOv8, YOLOv9, YOLOv10, YOLOv11 and Faster R-CNN models for detection of multiple weed species. Smart Agric Technol. https://doi.org/10.1016/j.atech.2024.100648
Junior LCM, Ulson JAC (2021) Real time weed detection using computer vision and deep learning. In: 2021 14th IEEE International Conference on Industry Applications (INDUSCON), pp 1131–1137. IEEE
Khalid M, Sarfraz MS, Iqbal U, Aftab MU, Niedbała G, Rauf HT (2023) Real-time plant health detection using deep convolutional neural networks. Agriculture 13(2):510
Gallo I, Rehman AU, Dehkordi RH, Landro N, La Grassa R, Boschetti M (2023) Deep object detection of crop weeds: performance of YOLOv7 on a real case dataset from UAV images. Remote Sens 15(2):539
Vaidya S, Kavthekar S, Joshi A (2023) Leveraging YOLOv7 for plant disease detection. In: 2023 4th International Conference on Innovative Trends in Information Technology (ICITIIT), pp 1–6. IEEE
Zayani HM, Ammar I, Ghodhbani R, Maqbool A, Saidani T, Slimane JB, Kachoukh A, Kouki M, Kallel M, Alsuwaylimi AA et al (2024) Deep learning for tomato disease detection with YOLOv8. Eng Technol Appl Sci Res 14(2):13584–13591
Ma B, Hua Z, Wen Y, Deng H, Zhao Y, Pu L, Song H (2024) Using an improved lightweight YOLOv8 model for real-time detection of multi-stage apple fruit in complex orchard environments. Artif Intell Agric
Junos MH, Mohd Khairuddin AS, Thannirmalai S, Dahari M (2021) An optimized YOLO-based object detection model for crop harvesting system. IET Image Proc 15(9):2112–2125
Zhao H, Tang Z, Li Z, Dong Y, Si Y, Lu M, Panoutsos G (2024) Real-time object detection and robotic manipulation for agriculture using a YOLO-based learning approach. Preprint at arXiv:2401.15785
Chen W, Zhang J, Guo B, Wei Q, Zhu Z (2021) An apple detection method based on des-YOLO v4 algorithm for harvesting robots in complex environment. Math Probl Eng 2021:1–12
Nergiz M, (2023) Enhancing strawberry harvesting efficiency through YOLO-v7 object detection assessment. Turk J Sci Technol 18(2):519–533
Wang C, Han Q, Li C, Li J, Kong D, Wang F, Zou X (2024) Assisting the planning of harvesting plans for large strawberry fields through image-processing method based on deep learning. Agriculture 14(4):560

Chen W, Lu S, Liu B, Chen M, Li G, Qian T (2022) CitrusYOLO: a algorithm for citrus detection under orchard environment based on YOLOv4. Multimed Tools Appl 81(22):31363–31389
Mirhaji H, Soleymani M, Asakereh A, Mehdizadeh SA (2021) Fruit detection and load estimation of an orange orchard using the YOLO models through simple approaches in different imaging and illumination conditions. Comput Electron Agric 191:106533
Sapkota R, Ahmed D, Churuvija M, Karkee M (2024) Immature green apple detection and sizing in commercial orchards using YOLOv8 and shape fitting techniques. IEEE Access 12:43436–43452
Wu D, Lv S, Jiang M, Song H (2020) Using channel pruning-based YOLO v4 deep learning algorithm for the real-time and accurate detection of apple flowers in natural environments. Comput Electron Agric 178:105742
Wang J, Gao Z, Zhang Y, Zhou J, Wu J, Li P (2021) Real-time detection and location of potted flowers based on a zed camera and a YOLO v4-tiny deep learning algorithm. Horticulturae 8(1):21
Khanal SR, Sapkota R, Ahmed D, Bhattarai U, Karkee M (2023) Machine vision system for early-stage apple flowers and flower clusters detection for precision thinning and pollination. IFAC-PapersOnLine 56(2):8914–8919
Xiao F, Wang H, Xu Y, Zhang R (2023) Fruit detection and recognition based on deep learning for automatic harvesting: an overview and review. Agronomy 13(6):1625
Yijing W, Yi Y, Xue-fen W, Jian C, Xinyun L (2021) Fig fruit recognition method based on YOLO v4 deep learning. In: 2021 18th International Conference on Electrical Engineering/Electronics, Computer, Telecommunications and Information Technology (ECTI-CON), pp 303–306. IEEE
Zhang Y, Li L, Chun C, Wen Y, Xu G (2024) Multi-scale feature adaptive fusion model for real-time detection in complex citrus orchard environments. Comput Electron Agric 219:108836
Zhou J, Zhang Y, Wang J (2023) A dragon fruit picking detection method based on YOLOv7 and PSP-ellipse. Sensors 23(8):3803
Xiuyan G, Zhang Y (2023) Detection of fruit using YOLOv8-based single stage detectors. Int J Adv Comput Sci Appl 14(12)
He Z, Karkee M, Upadhayay P (2021) Detection of strawberries with varying maturity levels for robotic harvesting using YOLOv4. In: 2021 ASABE Annual International Virtual Meeting, p. 1. American Society of Agricultural and Biological Engineers
Boudaa B, Abada K, Aichouche WA, Belakermi AN (2024) Advancing plant diseases detection with pre-trained YOLO models. In: 2024 6th International Conference on Pattern Analysis and Intelligent Systems (PAIS), pp 1–6. IEEE
Andreyanov N, Shleymovich M, Sytnik A (2022) Driver assistance system for agricultural machinery for obstacles detection based on deep neural networks. In: 2022 International Conference on Industrial Engineering, Applications and Manufacturing (ICIEAM), pp 880–885. IEEE
Jung T-H, Cates B, Choi I-K, Lee S-H, Choi J-M (2020) Multi-camera-based person recognition system for autonomous tractors. Designs 4(4):54
Xu S, Rai R (2024) Vision-based autonomous navigation stack for tractors operating in peach orchards. Comput Electron Agric 217:108558
Sapkota R, Karkee M (2025) Improved YOLOv12 with LLM-generated synthetic data for enhanced apple detection and benchmarking against YOLOv11 and YOLOv10. Preprint at arXiv:2503.00057
Sapkota R, Meng Z, Churuvija M, Du X, Ma Z, Karkee M (2024) Comprehensive performance evaluation of YOLO11, YOLOv10, YOLOv9 and YOLOv8 on detecting and counting fruitlet in complex orchard environments. Preprint at arXiv:2407.12040
Meng Z, Du X, Sapkota R, Ma Z, Cheng H (2025) YOLOv10-pose and YOLOv9-pose: Real-time strawberry stalk pose detection models. Comput Ind 165:104231
Sapkota R, Karkee M (2024) Comparing YOLOv11 and YOLOv8 for instance segmentation of occluded and non-occluded immature green fruits in complex orchard environment. Preprint at arXiv:2410.19869
Vo H-T, Mui KC, Thien NN, Tien PP (2024) Automating tomato ripeness classification and counting with YOLOv9. Int J Adv Comput Sci Appl 15(4)
Zhao J, Qu J (2019) A detection method for tomato fruit common physiological diseases based on YOLOv2. In: 2019 10th International Conference on Information Technology in Medicine and Education (ITME), pp 559–563. IEEE
Diwan T, Anirudh G, Tembhurne JV (2023) Object detection using YOLO: challenges, architectural successors, datasets and applications. Multimed Tools Appl 82(6):9243–9275
Ji S-J, Ling Q-H, Han F (2023) An improved algorithm for small object detection based on YOLO v4 and multi-scale contextual information. Comput Electr Eng 105:108490
Fang W, Wang L, Ren P (2019) Tinier-YOLO: a real-time object detection method for constrained environments. IEEE Access 8:1935–1944
Ye R, Gao Q, Qian Y, Sun J, Li T (2024) Improved YOLOv8 and Sahi model for the collaborative detection of small targets at the micro scale: a case study of pest detection in tea. Agronomy 14(5):1034

Olorunshola OE, Irhebhude ME, Evwiekpaefe AE (2023) A comparative study of YOLOv5 and YOLOv7 object detection algorithms. J Comput Soc Inform 2(1):1–12
Li N, Wang M, Yang G, Li B, Yuan B, Xu S (2023) Dens-YOLOv6: A small object detection model for garbage detection on water surface. Multimed Tools Appl 1–21
Jung H-K, Choi G-S (2022) Improved YOLOv5: efficient object detection using drone images under various conditions. Appl Sci 12(14):7255
Wang A, Peng T, Cao H, Xu Y, Wei X, Cui B (2022) TIA-YOLOv5: an improved YOLOv5 network for real-time detection of crop and weed in the field. Front Plant Sci 13:1091655
Wu T-H, Wang T-W, Liu Y-Q (2021) Real-time vehicle and distance detection based on improved YOLO v5 network. In: 2021 3rd World Symposium on Artificial Intelligence (WSAI), pp 24–28. IEEE
Ali MAM, Aly T, Raslan AT, Gheith M, Amin EA (2024) Advancing crowd object detection: a review of YOLO, CNN and vits hybrid approach. J Intell Learn Syst Appl 16(3):175–221
Pfeifer R, Iida F (2004) Embodied artificial intelligence: Trends and challenges. Lecture Notes Comput Sci 1–26
Sanket NJ (2021) Active vision based embodied-AI design for nano-UAV autonomy. PhD thesis, University of Maryland, College Park
Wang T, Zheng P, Li S, Wang L (2024) Multimodal human-robot interaction for human-centric smart manufacturing: a survey. Adv Intell Syst 6(3):2300359
Lakshmipathy A, Vardhineedi M, Sekharamahanthi VRP, Patel DD, Saini S, Mohammed S (2024) Medicaption: Integrating YOLO-driven computer vision and NLP for advanced pharmaceutical package recognition and annotation. Authorea Preprints
Li C, Zhang R, Wong J, Gokmen C, Srivastava S, Martín-Martín R, Wang C, Levine G, Ai W, Martinez B, et al (2024) Behavior-1k: A human-centered, embodied ai benchmark with 1000 everyday activities and realistic simulation. Preprint at arXiv:2403.09227
Pande AK, Brantley P, Tanveer MH, Voicu RC (2024) From AI to AGI-the evolution of real-time systems with GPT integration. In: SoutheastCon 2024, pp 699–707. IEEE
Qu Y, Wei C, Du P, Che W, Zhang C, Ouyang W, Bian Y, Xu F, Hu B, Du K, Wu H, Liu J, Liu Q (2024) Integration of cognitive tasks into artificial general intelligence test for large models. iScience 27
Rouhi A, Patiño D, Han DK (2025) Enhancing object detection by leveraging large language models for contextual knowledge. Lecture Notes in Computer Science (including subseries Lecture Notes in Artificial Intelligence and Lecture Notes in Bioinformatics) 15317 LNCS, 299–314, https://doi.org/10.1007/978-3-031-78447-7_20
Sapkota R, Paudel A, Karkee M (2024) Zero-shot automatic annotation and instance segmentation using LLM-generated datasets: Eliminating field imaging and manual annotation for deep learning model development. Preprint at arXiv:2411.11285
Xu R, Ji K, Yuan Z, Wang C, Xia Y (2024) Exploring the evolution trend of China's digital carbon footprint: a simulation based on system dynamics approach. Sustainability (Switzerland). https://doi.org/10.3390/su16104230
Dhar P (2020) The carbon impact of artificial intelligence. Nat Mach Intell. https://doi.org/10.1038/s42256-020-0219-9

**Publisher's Note** Springer Nature remains neutral with regard to jurisdictional claims in published maps and institutional affiliations.

## Authors and Affiliations

**Ranjan Sapkota[1] · Marco Flores-Calero[2] · Rizwan Qureshi[3] · Chetan Badgujar[4] · Upesh Nepal[5] · Alwin Poulose[6] · Peter Zeno[7] · Uday Bhanu Prakash Vaddevolu[8] · Sheheryar Khan[9] · Maged Shoman[10] · Hong Yan[11,12] · Manoj Karkee[1,13]**

✉ Ranjan Sapkota
rs2672@cornell.edu

✉ Manoj Karkee
mk2684@cornell.edu

1 Biological & Environmental Engineering, Cornell University, Ithaca, New York, USA

[2] Department of Electrical, Electronics and Telecommunications, Universidad de las Fuerzas Armadas ESPE, Av. General Rumiñahui s/n, Sangolquí 171-5-231B, Ecuador

[3] Center for Research in Computer Vision, University of Central Florida, Orlando, FL, USA

[4] Biosystems Engineering and Soil Sciences, The University of Tennessee, Knoxville, TN 37996, USA

[5] Cooper Machine Company, Inc., Wadley, GA 30477, USA

[6] School of Data Science, Indian Institute of Science Education and Research Thiruvananthapuram (IISER TVM), Thiruvananthapuram, Kerala 695551, India

[7] ZenoRobotics, LLC, Billings, MT 59106, USA

[8] Biological and Agricultural Engineering, Texas A&M University, College Station, TX 77840, USA

[9] School of Professional Education and Executive Development, The Hong Kong Polytechnic University, Hong Kong 999077, SAR China, China

[10] University of Tennessee, Knoxville, TN, USA

[11] Department of Electrical Engineering, City University of Hong Kong, Kowloon, Hong Kong 999077, SAR China, China

[12] Center for Intelligent Multidimensional Data Analysis, CIMDA, Hong Kong Science Park, Hong Kong, SAR China, China

[13] Biological Systems Engineering, Washington State University, Prosser Campus, Pullman, WA, USA